\documentclass[pdflatex,sn-apa]{sn-jnl}

\usepackage{graphicx}%
\usepackage{subcaption}%
\usepackage{multirow}%
\usepackage{amsmath,amssymb,amsfonts}%
\usepackage{amsthm}%
\usepackage{mathrsfs}%
\usepackage[title]{appendix}%
\usepackage[dvipsnames]{xcolor}%
\usepackage{textcomp}%
\usepackage{manyfoot}%
\usepackage{booktabs}%
\usepackage{algorithm}%
\usepackage{algorithmicx}%
\usepackage{algpseudocode}%
\usepackage{listings}%
\usepackage{anyfontsize}%
\usepackage{float}
\usepackage{tikz}%
\usetikzlibrary{ext.paths.ortho}

\theoremstyle{thmstyleone}%
\theoremstyle{thmstyletwo}%

\theoremstyle{thmstylethree}%

\begin{document}

\title[CLUES-WEASEL]{CLUES-WEASEL: No additional clues required to choose your time series clustering algorithm}

\author[]{\fnm{Johann} \sur{Faouzi}}\email{johann.faouzi@ensai.fr}

\affil[]{Univ Rennes, Ensai, CNRS, CREST - UMR 9194, F-35000 Rennes, France}

\abstract{
    Time series data is very common in many real-world applications and in numerous domains, with increasing interest for automated information extraction using machine learning.
    One of these subfields is time series clustering, which consists in identifying clusters among a set of time series in an unsupervised fashion.
    Most time series clustering algorithms suffer from the same balancing act: they trade clustering performance for faster runtimes or vice versa.
    We present a novel time series clustering algorithm that we call CLUES-WEASEL, which stands for \emph{\textbf{CL}ustering with the \textbf{U}nsupervis\textbf{E}d \textbf{S}econd version of \textbf{W}ord \textbf{E}xtr\textbf{A}ction for time \textbf{SE}ries c\textbf{L}assification}.
    CLUES-WEASEL extracts features using the unsupervised version of the transformation step of WEASEL 2.0, which is a time series classification algorithm, then reduces these features using principal component analysis, and finally performs clustering with the $k$-means algorithm using these reduced extracted features.
    Through extensive experiments, we prove that CLUES-WEASEL is significantly better than any other existing time series clustering algorithm while being (much) faster than any state-of-the-art one.
    We also show that the architecture of CLUES-WEASEL can work well with other time series feature extraction algorithms.
    Our findings highlight the relevance of CLUES-WEASEL for time series clustering.
}


\maketitle

\section{Introduction}\label{sec1}

We consider time series as an ordered sequence of real-valued observations (often called univariate time series) sampled at equal frequency.
Such data are very common in many real-world applications given their nature.
With the increasing availability of sensors as well the development of \emph{Internet of things} devices, more time series data and applications become available.
Numerous varied fields are affected, from econometrics and finance \citep{perron1989great} to climate science with weather forecasting \citep{agrawal2012application} to geology with earthquake prediction \citep{amei2012time}.

Clustering is a subfield of machine learning, and more specifically unsupervised learning, aiming at grouping similar samples together, while dissimilar instances belong to different groups \citep{rokach2005clustering}.
Formally, the set of samples is represented as a union of disjoint subsets of samples.
Some clustering algorithms always assign a group to every sample, while some samples might not be assigned to any group if they are deemed to be outliers by other clustering algorithms.
Popular standard clustering algorithms include $k$-means \citep{macqueenMethodsClassificationAnalysis1967}, spectral clustering \citep{vonluxburgTutorialSpectralClustering2007}, Gaussian mixtures \citep{jainAlgorithmsClusteringData1988}, hierarchical clustering \citep{jainAlgorithmsClusteringData1988}, and density-based spatial clustering of applications with noise \citep{esterDensitybasedAlgorithmDiscovering1996}.
Clustering is an essential approach in data analysis for extracting relevant information from unlabeled data, with successful applications in many fields such as transportation and logistics, manufacturing, energy, and healthcare \citep{oyewole2023data}.

Time series clustering consists in performing clustering on time series data and has been employed in various fields, such as healthcare and speech recognition, and for various applications, such as anomaly detection and indexing \citep{paparrizosBridgingGapDecade2024}.
Many algorithms dedicated to time series clustering have been developed.
Some of them consist in adapting partition-based clustering algorithms with distances, assignment methods, and averaging functions more specific to time series data \citep{holderBarycentreAveragingMoveSplitMerge2023, holderRockKASBABlazingly2024,petitjeanGlobalAveragingMethod2011a, ismail-fawazShapeDBAGeneratingEffective2023}.
Others consist in extracting features from time series and applying a standard clustering algorithms \citep{tianoFeatTSFeaturebasedTime2021, zakariaClusteringTimeSeries2012, zhangUnsupervisedFeatureLearning2016, rutaSAXNavigatorTime2019,jorgeTimeSeriesClustering2024, liRandomnetClusteringTime2024}.

We present a novel time series clustering algorithm that we call CLUES-WEASEL, which stands for \emph{\textbf{CL}ustering with the \textbf{U}nsupervis\textbf{E}d \textbf{S}econd version of \textbf{W}ord \textbf{E}xtr\textbf{A}ction for time \textbf{SE}ries c\textbf{L}assification}.
CLUES-WEASEL uses the unsupervised version of the transformation step of WEASEL 2.0 \citep{schaferWEASEL20Random2023}, which is an algorithm developed for time series classification, to extract features, then performs dimensionality reduction using principal component analysis (PCA) \citep{gewersPrincipalComponentAnalysis2021} to reduce the feature space, and finally applies the $k$-means algorithm \citep{macqueenMethodsClassificationAnalysis1967,lloydLeastSquaresQuantization1982a} on these reduced extracted features.
We evaluate CLUES-WEASEL on the UCR time series classification archive \citep{dauUCRTimeSeries2019e} and compare its clustering performance to state-of-the-art time series clustering algorithms.
We also investigate the use of other unsupervised transformation algorithms developed for time series classification.
Our results show that CLUES-WEASEL delivers state-of-the-art time series clustering performance by a large margin while also being faster than state-of-the-art algorithms.
We prove the relevance of other unsupervised transformation algorithms for time series classification, and that the unsupervised version of the transformation of WEASEL 2.0 yields the best results.
We also investigate the maximum number of features extracted by the unsupervised version of the transformation step of WEASEL 2.0.

The remaining of this paper is structured as follows.
Section~\ref{sec2} provides relevant information about existing time series analysis algorithms, notably for clustering and classification.
In Section~\ref{sec3}, we describe the architecture of CLUES-WEASEL, the reasoning behind its architecture and which elements were optimized.
Section~\ref{sec4} presents the algorithms that we compare CLUES-WEASEL to, as well as the data sets and metrics used to perform these comparisons.
We provide the results of all the experiments in Section~\ref{sec5} and perform short exploratory analyses in Section~\ref{sec6} before concluding in Section~\ref{sec7}.

\section{Related work}\label{sec2}

In this section, we provide information about time series clustering and classification algorithms.
We include time series classification because it is relevant for our work and because some time series clustering algorithms are heavily inspired by time series classification ones.
We categorize the algorithms based on their approach and present the more relevant ones to our study, as providing a systematic review is out of scope of this study.
For recent reviews, we refer the readers to \citep{middlehurstBakeReduxReview2024} for time series classification and \citep{paparrizosBridgingGapDecade2024} for time series clustering.
We do not include time series clustering approaches based on neural networks in this section as it is out of scope of this study, and refer the readers to \citep{paparrizosBridgingGapDecade2024, lafabregueEndtoendDeepRepresentation2022}.

\subsection{Distance-based approaches}

Computing similarity scores between samples is a common approach in machine learning.
For instance, such methods include the $k$-nearest neighbors \citep{fixDiscriminatoryAnalysisNonparametric1989, coverNearestNeighborPattern1967} and support vector machines \citep{cortesSupportVectorNetworks1995a} algorithms in supervised learning, as well as the $k$-means \citep{macqueenMethodsClassificationAnalysis1967, lloydLeastSquaresQuantization1982a}, hierarchical clustering \citep{murtaghAlgorithmsHierarchicalClustering2012}, t-distributed Stochastic Neighbor Embedding (t-SNE) \citep{maatenVisualizingDataUsing2008}, and Uniform Manifold Approximation and Projection (UMAP) \citep{mcinnesUMAPUniformManifold2020} algorithms in unsupervised learning, to name a few.
Additionally, for clustering algorithms that compute centroids, an averaging method is necessary to compute the centroid of each cluster.
For instance, for the $k$-means algorithm with the Euclidean distance, the centroid of a cluster is the average of all the samples assigned to this cluster.

For time series analysis, specific metrics have been developed.
Dynamic Time Warping (DTW) \citep{sakoeDynamicProgrammingAlgorithm1978, berndtUsingDynamicTime1994} is an elastic distance that uses dynamic programming to find the optimal alignment between two time series by computing the minimum path through a cost matrix consisting of the pairwise pointwise squared differences.
Starting from the starting indices of both time series, each index can be incremented by at most one at every step, and the final indices must be the ending indices of both time series.
Several variants of DTW have been developed, some of them adding a region constraint on the possible set of paths \citep{sakoeDynamicProgrammingAlgorithm1978, itakuraMinimumPredictionResidual1975} and some others adding weights penalizing alignments with high phase differences \citep{jeongWeightedDynamicTime2011, herrmannAmercingIntuitiveEffective2023}.
One of them, called Shape-DTW \citep{zhaoShapeDTWShapeDynamic2018}, uses shape descriptors to encode structural information, computes the optimal warping path on this new representation, and finally applies this optimal path on the original time series.
The Move-Split-Merge (MSM) distance \citep{stefanMoveSplitMergeMetricTime2013} is inspired by edit distances.
It also relies on a cost matrix, but the latter consists of the pairwise pointwise absolute differences.
Moreover, a diagonal move on the cost matrix (that is both indices being incremented by one) is penalized contrary to DTW.

\subsubsection{Time series classification}

The DTW distance and its variants, as well as the MSM distance, have been used in conjunction with the one-nearest-neighbor algorithm for time series classification \citep{stefanMoveSplitMergeMetricTime2013}.
It should be noted that the MSM distance satisfies the triangle inequality, contrary to the DTW distance, enabling the use of more efficient nearest-neighbor search algorithms \citep{bentleyMultidimensionalBinarySearch1975a, omohundro1989five} than the brute-force approach.
A one-nearest-neighbor classification algorithm with the DTW distance is still used as a baseline classifier in current time series classification benchmarks.

\subsubsection{Time series clustering}

Numerous research works have been carried out on times series clustering with distance-based approaches.
Some of them have a structure very similar to the standard $k$-means algorithm, that is:
\begin{enumerate}
    \item \textbf{Initialization step}: The centroid of each cluster is initialized.
    \item \textbf{} Until a stopping criterion is satisfied, the following two steps are repeated:
    \begin{enumerate}
        \item \textbf{Assignment step}: Each time series is assigned to a cluster using an assignment method based on a distance.
        \item \textbf{Update step}: The centroid of each cluster is computed using an averaging function.
    \end{enumerate}
\end{enumerate}
There are four key hyperparameters to these approaches: the initialization method, the distance, the assignment method, and the averaging function.
For instance, the $k$-means algorithm uses either the Forgy method (the initial clusters are samples randomly chosen) \citep{forgy1965cluster} or the $k$-means++ method \citep{arthurKmeansAdvantagesCareful2007a}, the Euclidean distance, the Lloyd's assignment \citep{lloydLeastSquaresQuantization1982a} (that is the cluster whose centroid is the closest to the sample for the chosen distance), and the mean function.
For time series clustering, specific methods have been developed for these four hyperparameters.

The DTW barycenter averaging (DBA) \citep{petitjeanGlobalAveragingMethod2011a} is an averaging technique aiming at minimizing the sum of squared DTW distances from the average time series to the set of time series.
Shape-DBA \citep{ismail-fawazShapeDBAGeneratingEffective2023} uses the Shape-DTW distance and employs an averaging technique aiming at minimizing the sum of squared Shape-DTW distances from the average time series to the set of time series.
The MSM distance was also investigated with the $k$-means algorithm and was found to be the best performing elastic distance compared to nine other elastic distances \citep{holderReviewEvaluationElastic2024}.
The $k$-means accelerated stochastic subgradient barycenter average (KASBA) \citep{holderRockKASBABlazingly2024} uses the MSM distance, an initialization procedure based on this distance, and a barycenter stochastic gradient descent method as the averaging function.

The MSM distance also been investigated with other partition-based methods, namely $k$-medoids algorithms, and it was found that the Partition Around Medoids (PAM) \citep{leonard1990partitioning} was the best performing method in conjunction with this distance \citep{holderClusteringTimeSeries2023}.

\subsection{Feature-based approaches}

Feature-based approaches consist in computing statistics from the whole time series.

\subsubsection{Time series classification}

For time series classification, a standard classification algorithm is applied on top of these computed statistics.

The canonical time series characteristics (Catch22) \citep{lubbaCatch22CAnonicalTimeseries2019} are 22 features that have been determined to be discriminative on the UCR data sets \citep{dauUCRTimeSeries2019e}, and a decision tree was used to performed classification.

The Time Series Feature Extraction based on Scalable Hypothesis Tests (TSFresh) algorithm \citep{christTimeSeriesFeatuRe2018a} is a set of close to 800 features extracted from time series.
This set of features can be pruned using statistical tests.
Random forest \citep{breimanRandomForests2001a} and AdaBoost \citep{freundExperimentsNewBoosting1996} were the investigated classification algorithms applied on top of these extracted features.

\subsubsection{Time series clustering}

To the best of our knowledge, these feature-based algorithms have been little used for time series clustering.

The Feature-based Time Series Clustering (FeatTS) algorithm \citep{tianoFeatTSFeaturebasedTime2021} extracts features from time series using TSFresh and performs clustering using $k$-medoids \citep{jainAlgorithmsClusteringData1988} on these features.

The other feature-based time series clustering algorithms that we found are generally specific to some time series types or data sets \citep{choksiFeatureBasedClustering2020, liAngClustAngleFeatureBased2023, rasanenFeatureBasedClusteringElectricity2009} and have not been evaluated on the UCR time series classification archive \citep{dauUCRTimeSeries2019e}.

\subsection{Interval-based approaches}

Interval-based methods generate intervals, extract the subseries corresponding to these intervals, and compute statistics from these subseries.

\subsubsection{Time series classification}

Several interval-based approaches have been developed for time series classification.

The Time Series Forest (TSF) algorithm \citep{dengTimeSeriesForest2013} computes three statistics for each subseries (the mean, the standard deviation and the slope) and then applies a random forest \citep{breimanRandomForests2001a} on top of this transformation.

The Canonical Interval Forest (CIF) algorithm \citep{middlehurstCanonicalIntervalForest2020} is an extension of TSF with more features being extracted.
The Diverse Representation Canonical Interval Forest (DrCIF) algorithm \citep{middlehurstHIVECOTE20New2021a} extends CIF with two additional time series representations: periodograms and first-order differences.

The QUANT algorithm \citep{dempsterQuantMinimalistInterval2024} uses fixed dyadic intervals as well as quantiles instead of all other features, but includes four time series representations: raw time series, first- and second-order differences, and Fourier coefficients.
The classification algorithm used in conjunction is extremely randomized trees \citep{geurtsExtremelyRandomizedTrees2006a}.

\subsubsection{Time series clustering}

To the best of our knowledge, no time series clustering algorithm based on interval-based time series classification algorithm has been published as of the time of writing.

\subsection{Shapelet-based approaches}

Shapelets are subseries extracted from the training time series and used to differentiate time series.
To do so, the most commonly used metric is the minimum Euclidean distance between the shapelet and all the subseries, of the same length as the shapelet, extracted from the time series.

\subsubsection{Time series classification}

For time series classification, shapelets were first investigated in \citep{yeTimeSeriesShapelets2011} and the transformation was followed by a decision tree to perform classification.

The Shapelet Transform Classifier (STC) \citep{hillsClassificationTimeSeries2014a} investigates all the possible shapelets from the training set before selecting the most discriminative ones, followed by an ensemble of classifiers.
Several refinements have been made since the release of its original version to improve its performance and scalability, notably performing a random search for the shapelets and using a single classification algorithm on top of the transformation \citep{bostromBinaryShapeletTransform2017, bostromEvaluatingImprovementsShapelet2016a}.

The Random Dilated Shapelet Transform (RDST) algorithm \citep{guillaumeRandomDilatedShapelet2022} adds two novel elements to existing shapelet-based approaches.
The first one is the use of dilation, meaning that the entries of a shapelet are not necessarily consecutive entries of a training time series: every other $d$ entries are used, with $d$ being the value of the dilation.
The second one is the addition of two new features computed for each shapelet: the position of the minimum distance and the number of occurrences of the shapelet.

\subsubsection{Time series clustering}

Shapelet-based approaches have also been investigated for time series clustering.
The U-Shapelets algorithm \citep{zakariaClusteringTimeSeries2012} proposed a novel method for unsupervised shapelet searching by maximizing the separation gap between clusters.
The Unsupervised Shapelet Learning Model (USLM) algorithm \citep{zhangUnsupervisedFeatureLearning2016} involves a more efficient shapelet learning strategy by jointly optimizing a cost function including both the shapelet transform and the distances.

\subsection{Dictionary-based approaches}

\begin{figure}
    \includegraphics[width=\textwidth]{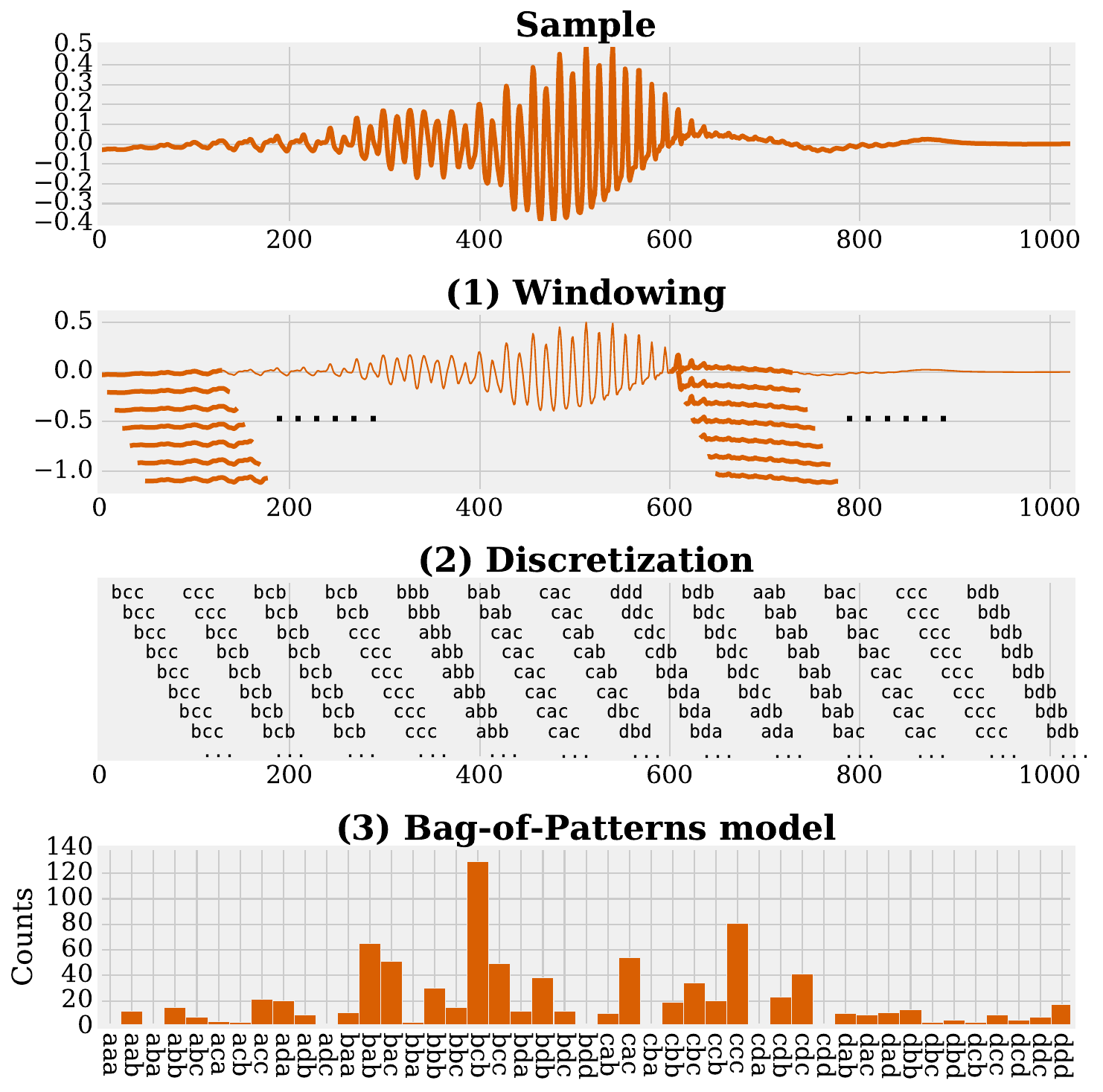}
    \caption{
        Transformation of a time series into word counts.
        From a time series (top), subseries are extracted using overlapping windows (second to top) and are transformed into words (second to bottom), and the word counts are computed (bottom).
        This type of transformation is the core of dictionary-based approaches.
        The image is from \citep{schaferFastAccurateTime2017a} and the authors have permitted its reuse.
    }
    \label{fig1}
\end{figure}

Dictionary-based methods, similarly to shapelet-based approaches, also extract subseries from time series.
However, no distance between the subseries and the time series is computed.
Instead, each subseries is turned into a short sequence of discrete symbols, which is usually called a word.
The frequencies of the all the words extracted from a time series are then computed to obtain the new representation of this time series.
Figure~\ref{fig1} illustrates this pipeline.
For time series classification, a standard machine learning classification algorithm is applied on top of this transformation.

There exist two main symbolic representations of time series.
The first one, in the time domain, is the Symbolic Aggregate approXimation (SAX) \citep{linExperiencingSAXNovel2007a}, which performs dimensionality reduction first in the time domain (using the mean from non-overlapping windows) and then in the value domain (using discretization based on quantiles).
The second one, in the frequency domain, is the Symbolic Fourier Approximation (SFA) \citep{schaferSFASymbolicFourier2012}, which computes the discrete Fourier transform of the time series, selects a subset of the Fourier coefficients, and discretizes them (using quantiles).

\subsubsection{Time series classification}

The SAX representation has been used for time series classification.
Indeed, the authors proposing SAX also proposed a distance between two SAX representations, investigated its use with a nearest-neighbor classification algorithm, and showed that it was competitive with the Euclidean distance while being much faster.

The SAX representation has also been used in conjunction with a vector space model \citep{seninSAXVSMInterpretableTime2013}.
Given the nature of the transformation, the authors proposed to use techniques from natural language processing, notably the \emph{term frequency -- inverse document frequency}.
Here, a term is word and a document is a time series.
For each class, the frequency of each word is first computed.
Then, these frequencies are weighted by the inverse of their frequencies among the classes.
This weighing is done to highlight words that are specific to one or few classes in contrary to words that are present in many classes.
Thus, a vector is obtained for each class.
For a new time series, it is first transformed into a vector of word frequencies, and the predicted class is the class associated with the class vector giving the highest cosine similarity with this vector of word frequencies.

The SFA representation was first introduced for indexing high dimensional data sets, but has also been used in several algorithms developed for time series classification.

The Bag-of-SFA-Symbols (BOSS) algorithm \citep{schaferBOSSConcernedTime2015} extracts subseries from a time series using overlapping windows, then transforms each subseries into a word using SFA, and finally the word frequencies are computed.
A new distance for this representation was also developed and used in conjunction with a nearest-neighbor classifier to perform classification.
An ensemble of BOSS classification algorithms was used in practice, with different values for several hyperparameters.
A modified version of BOSS in vector space \citep{schaferBagOfSFASymbolsVectorSpace2015}, inspired by SAX-VSM, was proposed as a faster but less accurate alternative to BOSS.

\begin{figure}
    \includegraphics[width=\textwidth]{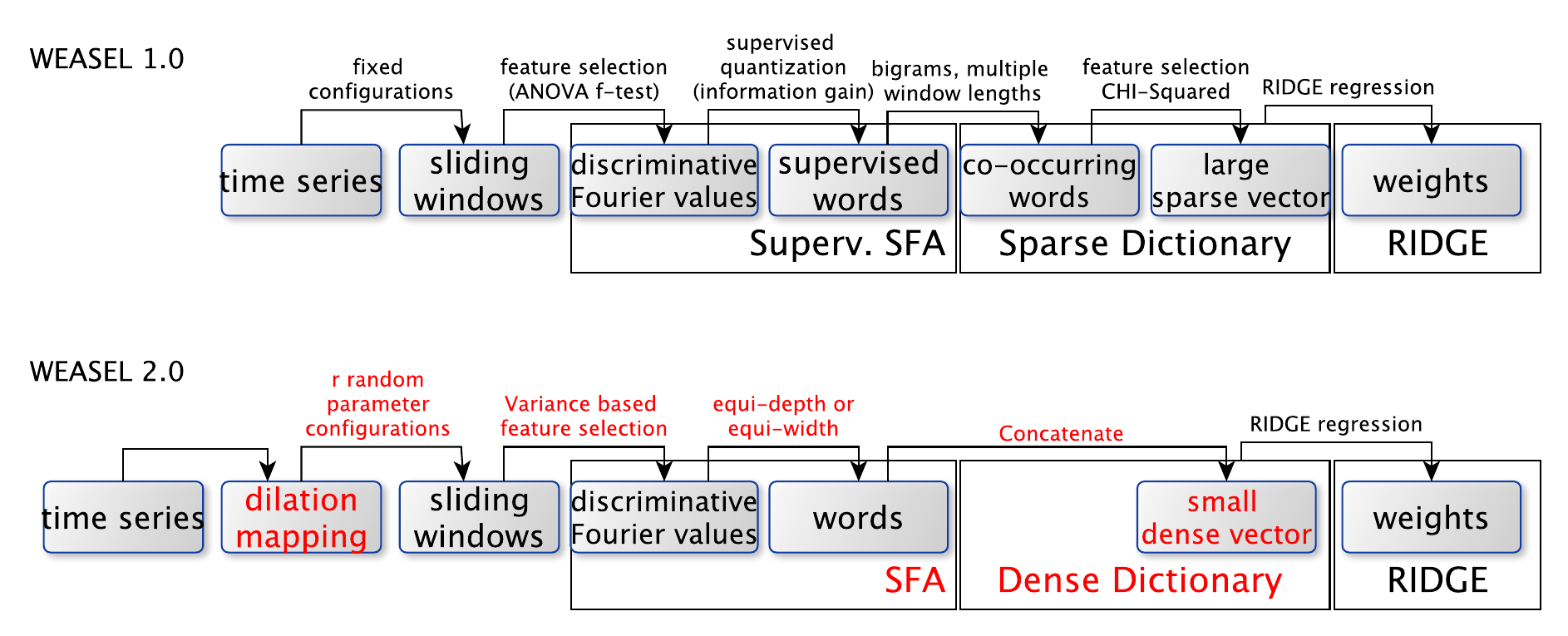}
    \caption{
        Differences between the first and second versions of WEASEL, highlighted in red.
        Compared to the first version, WEASEL 2.0 adds a dilation mapping before the sliding window operator, uses an unsupervised version of SFA, chooses parameters to control the size and memory consumption of the dictionary, and adds randomization to increase variance.
        The classification algorithm used in WEASEL 1.0 is actually logistic regression and not Ridge.
        The image is from \citep{schaferWEASEL20Random2023} and the authors have permitted its reuse.
    }
    \label{fig2}
\end{figure}

The Word Extraction for Time Series Classification (WEASEL) algorithm \citep{schaferFastAccurateTime2017a} involves several new changes compared to BOSS.
Notably, the selection of the Fourier coefficients is supervised (using ANOVA tests), the quantization of SFA is also supervised (using information gain), bigrams and multiple window lengths are considered, a subset of words is selected in a supervised fashion (using chi-squared tests), and the classification step is replaced with logistic regression.
WEASEL was later refined in a new version called WEASEL 2.0 \citep{schaferWEASEL20Random2023}, adding notably a novel dilation mapping, using less supervised selection methods in SFA, and replacing logistic regression with a Ridge classification algorithm.
Figure~\ref{fig2} highlights these differences.

\subsubsection{Time series clustering}

To the best of our knowledge, dictionary-based approaches have not been much used for time series clustering.

The authors who proposed SAX also investigated its use for time series clustering, applying Lloyd's algorithm on top of the SAX representation.
\cite{keoghParameterfreeDataMining2004} proposed a compression-based dissimilarity measure to compare two strings and applied it to time series clustering, with time series being transformed into strings using SAX.
\cite{rutaSAXNavigatorTime2019} used agglomerative hierarchical clustering using the SAX representation and its corresponding distance to perform time series clustering.
\cite{ijcai2019p406} used the SAX representation to transform time series and proposed an ensemble of trees where each split is determined by the presence or absence of a pattern in the transformed time series.

To the best of our knowledge, no time series clustering algorithm using SFA or another algorithm based on SFA has been published as of the time of writing.

\subsection{Convolution-based approaches}

Convolution is a mathematical operation that has applications in numerous fields such as signal and image processing.
In machine learning, a convolution is characterized by a kernel with its weights and its possible bias, as well as its padding and its dilation.
In deep learning, convolutional layers, that are the backbones of convolutional neural networks, are characterized by their kernels.
Several convolution-based algorithms have been developed for time series classification, which have inspired new algorithms for time series clustering.

\subsubsection{Time series classification}

Instead of learning the kernels given a specific classification task, it has been proposed to use random kernels for time series classification.
This strategy was introduced with the Random Convolutional Kernel Transform (ROCKET) algorithm \citep{dempsterROCKETExceptionallyFast2020}.
ROCKET generates numerous random kernels (with random length, weights, bias, padding and dilation), applies the convolution for each of them, and extracts two aggregate features for each of them: the maximum value and the proportion of positive values.
Finally, a Ridge classifier is trained on these extracted features.

ROCKET has been extended in two versions.
The first one is MiniROCKET \citep{dempsterMiniRocketVeryFast2021a} which involves much less randomness in the generation of the kernels than ROCKET.
Moreover, only the proportion of positive values is extracted from the convolutional outputs.
MiniROCKET is faster than ROCKET with very similar performance.
The second extension is MultiROCKET \citep{tanMultiRocketMultiplePooling2022}, which adopts the improvements of MiniROCKET but also includes two notable new changes.
First, in addition to the proportion of positive values, three new features are extracted from the convolutional output: the mean of positive values, the mean of indices of positive values, and the longest stretch of positive values.
Second, half of the convolutions are applied to the first-order difference of the time series.

A model combining both dictionary- and convolution-based approaches, called HYbrid Dictionary-ROCKET Architecture (Hydra) \citep{dempsterHydraCompetingConvolutional2023}, was later proposed.
The kernels are aggregated into groups, the best matching kernel among each group is recorded, and their frequencies are computed.
The combination of both MultiROCKET and Hydra is a very performing time series classification algorithm while still being fast \citep{dempsterHydraCompetingConvolutional2023}.

\subsubsection{Time series clustering}

The unsupervised convolution-based transformations developed for time series classification have inspired new algorithms for time series clustering.

R-Clustering \citep{jorgeTimeSeriesClustering2024} uses a modified version of MiniROCKET to extract features, followed by principal component analysis to reduce the dimensionality of the feature space, and finally a $k$-means algorithm to perform clustering on these reduced features.

RandomNet \citep{liRandomnetClusteringTime2024} is a deep neural network composed with random weights and biases.
It is an ensemble method with several independent convolutional and recurrent blocks to extract features, followed by $k$-means algorithms to perform one clustering for each block.
Finally, a subset of the clusterings is selected by removing clusterings with too small or large clusters, and the final clustering is computed using bipartite graph partitioning \citep{fernSolvingClusterEnsemble2004} from this subset of clusterings.
Even though we mentioned that we do not cover artificial neural networks in this section, we included RandomNet because its parameters are random and not learned, similarly to the other convolution-based methods aforementioned.

\section{Methodology}\label{sec3}

In this section, we detail the proposed clustering algorithm and specify which hyperparameters were optimized and the values for hyperparameters that were not optimized.

\subsection{Time series clustering algorithm}

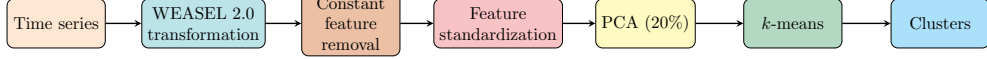
\begin{figure}
    \resizebox{\linewidth}{!}{
    \begin{tikzpicture}[node distance=2cm]

        \tikzstyle{io} = [
            rectangle,
            centered,
            rounded corners,
            minimum width=2cm,
            minimum height=1cm,
            text centered,
            align=center,
            draw=black,
        ]
        \tikzstyle{arrow} = [thick,->,>=stealth]

        \node (input) [io, fill=Apricot!30] at (0, 0) {Time series};
        \node (weasel) [io, fill=Aquamarine!30] at (3, 0) {WEASEL 2.0\\ transformation};
        \node (constant) [io, fill=Bittersweet!30] at (6, 0) {Constant\\ feature\\ removal};
        \node (standardization) [io, fill=WildStrawberry!30] at (9, 0) {Feature\\ standardization};
        \node (pca) [io, fill=Yellow!30] at (12, 0) {PCA (20\%)};
        \node (kmeans) [io, fill=ForestGreen!30] at (15, -0) {$k$-means};
        \node (clusters) [io, fill=Cyan!30] at (18, 0) {Clusters};

        \draw [arrow] (input) -- (weasel);
        \draw [arrow] (weasel) -- (constant);
        \draw [arrow] (constant) -- (standardization);
        \draw [arrow] (standardization) -- (pca);
        \draw [arrow] (pca) -- (kmeans);
        \draw [arrow] (kmeans) -- (clusters);

    \end{tikzpicture}}
    \caption{
        Architecture of CLUES-WEASEL.
        First, features (word counts) are extracted using the transformation step of WEASEL 2.0 \citep{schaferWEASEL20Random2023}.
        We remove the potential constant features as they are obviously not discriminative.
        We then standardize the features to have zero mean and unit variance before applying principal component analysis (PCA) for dimensionality reduction.
        The number of components selected is the smallest one such that the selected components explain 20\% of the total explained variance.
        Finally, the $k$-means algorithm is applied on these reduced extracted features to predict the clusters.
    }
    \label{fig3}
\end{figure}

We present a novel time series clustering algorithm that we call \emph{CLUES-WEASEL}, which stands for \emph{\textbf{CL}ustering with the \textbf{U}nsupervis\textbf{E}d \textbf{S}econd version of \textbf{W}ord \textbf{E}xtr\textbf{A}ction for time \textbf{SE}ries c\textbf{L}assification}.
CLUES-WEASEL uses the unsupervised version of the transformation of WEASEL 2.0, then performs dimensionality reduction using PCA, and finally applies the $k$-means algorithm on these reduced extracted features.
Figure~\ref{fig3} highlights the pipeline after hyperparameter optimization.
In the remaining of this subsection, we provide more information about each step of CLUES-WEASEL.

\subsubsection{Feature extraction}

As mentioned in Section \ref{sec2}, a lot of algorithms have been developed for time series classification and clustering.
They can be grouped in different categories and, for time series classification, many of them consist in extracting features and then applying a standard classification algorithm.
Moreover, most of the recently published algorithms use the same classification algorithm, namely the classification version of Ridge \citep{hoerlRidgeRegressionApplications1970, hoerlRidgeRegressionBiased1970} because of its efficiency to perform leave-one-out cross-validation on the regularization term and to handle multiclass classification tasks.
Since we are interested in time series clustering, we are only interested in the feature extraction step.
Moreover, since clustering is unsupervised, the feature extraction step needs to be unsupervised (or at lest to have an unsupervised version).
We choose the feature extraction step of WEASEL 2.0 \citep{schaferWEASEL20Random2023} because of the predictive performance of WEASEL 2.0 and the simplicity of using an unsupervised version of its feature extraction step.

WEASEL 2.0 has a parameter called \texttt{max\_feature\_count}, which is the upper limit of the number of kept features, to limit the memory footprint.
The default value for this parameter is 30,000, and three methods are available to perform this potential selection:
\begin{itemize}
    \item By default, the feature selection is performed using chi-squared tests, and features with the highest $p$-values are dropped. This method is supervised and we cannot use it.
    \item The feature selection can also be totally random. This method is unsupervised.
    \item The final option is to keep all the features, that is no feature selection is performed, and the \texttt{max\_feature\_count} parameter is ignored.
\end{itemize}
We decide to use the default value for the \texttt{max\_feature\_count} parameter, that is 30,000, and to select features at random if more than 30,000 features are extracted.
We justify this choice in the next section as it is related to the dimensionality reduction technique that is used.
Our pipeline also removes the potential constant features as they are not discriminative.

\subsubsection{Dimensionality reduction}

As aforementioned, the first step of many of time series classification algorithms consists in extracting features.
For some of them, the default number of features can be very large.
For instance, MultiROCKET extracts nearly 50,000 features, RDST extracts 30,000 features, and WEASEL 2.0 extracts at most 30,000 features.
The task of figuring out which of these features are useful for the given classification task is left to the classification algorithm, which usually includes regularization or bootstrap aggregating to tackle this large number of features.
For time series clustering, we cannot use any supervised method to find the relevant features and their weights.
Moreover, we have to take the curse of dimensionality into account: distances in high dimensional space are not always relevant, as the distribution of pairwise distances becomes more and more uniform as the dimensionality increases.
For these reasons, we decide to apply dimensionality reduction.
Among the different options, we decide to use principal component analysis for two reasons.
The first one is that it is a linear dimensionality reduction technique, and so is the classification algorithm of WEASEL 2.0, namely the classification version of Ridge.
The second one is the possibility to investigate how much total variance the selected components explain.
Before applying PCA, we standardize the features to have zero mean and unit variance not to give more weights to some features than others because we do not have any a priori information.

Principal component analysis requires computing the singular value decomposition of the input data.
The complexity of this decomposition is equal to the number of samples times the square of the number of features.
With a quadratic complexity in terms of the number of features, we would like the number of features not to be too high.
With the default value of at most 30,000 extracted features selected at random, CLUES-WEASEL is still fast and the required memory is reasonable for modern hardware.
We are also aware of incremental versions of PCA that do not process all the samples at once but by batches \citep{levySequentialKarhunenLoeveBasis1998, rossIncrementalLearningRobust2008}, reducing the complexity in terms of the number of samples, but we do not further investigate this approach.

\subsubsection{Clustering on reduced extracted features}

Now that the feature space is smaller, we can apply a standard clustering algorithm.
We choose $k$-means \citep{macqueenMethodsClassificationAnalysis1967} for several reasons.
First, it is one of the most popular and standard clustering algorithms.
Second, it is an algorithm that can easily and naturally be applied to new unseen data: the predicted cluster is simply the cluster whose centroid is the closest to this new observation.
Third, it has already been successfully applied to time series clustering both in this vanilla version with extracted features \citep{jorgeTimeSeriesClustering2024} and in modified versions with time series specific metrics \citep{paparrizosKShapeEfficientAccurate2016, yangPatternsTemporalVariation2011,petitjeanGlobalAveragingMethod2011a, holderBarycentreAveragingMoveSplitMerge2023, ismail-fawazShapeDBAGeneratingEffective2023, cuturiSoftDTWDifferentiableLoss2017a}.

\subsection{Hyperparameter values and optimization}\label{sec3sub2}

Among the five steps of the CLUES-WEASEL pipeline, three have hyperparameters that could be optimized: the transformation step of WEASEL 2.0, principal component analysis, and $k$-means.
Performing an exhaustive grid search over many hyperparameters would be computationally intensive.
Among those three steps, we assume that the most critical one is principal component analysis.
Indeed, WEASEL 2.0 has proven to be a strong time series classification algorithm with the default values for its hyperparameters, so we assume that the default values for the hyperparameters of its transformation step are likely effective.
The only change that we make is to replace the supervised feature selection using chi-squared tests with unsupervised feature selection at random, as aforementioned.
Likewise, $k$-means is a popular standard clustering algorithm without many hyperparameters.
Centroids are initialized with the $k$-means++ initialization \citep{arthurKmeansAdvantagesCareful2007a}, which is usually effective in generating good initial centroids, so we use a single initialization to reduce the complexity of CLUES-WEASEL.

The step that we focus on for hyperparameter optimization is thus the dimensionality reduction step using PCA.
This technique consists in finding a representation of the data through principal components.
The principal components are a sequence of unit vectors such that the $i$-th vector is the best approximation of the data (i.e., maximizing the explained variance) while being orthogonal to the first $i - 1$ vectors.

The main hyperparameter of PCA is the number of components selected.
The upper bound for the number of selected components is equal to minimum between the number of samples and the number of features.
For data sets with very different sizes, this can lead to very different upper bounds.
Moreover, it seems inappropriate to use a raw value for the number of selected components when some data sets can be orders of magnitude larger than others.
However, one of the upsides of PCA is that it provides the variance explained by each component.
The explained variances can also be computed as ratios to have the relative variance instead of the raw variance.
This information is used in R-Clustering \citep{jorgeTimeSeriesClustering2024}: the components that explain less than 1\% of the total variance are removed.

Instead of reasoning in terms of variance explained ratio by each component independently, we reason in terms of \emph{total} explained variance ratio.
Since the components are orthogonal, the ratio of variance explained by all the selected components is the sum of the ratios of variance explained by each selected component.
For instance, if many components explain less than 1\% of the total variance are removed, a lot of information might be lost, while the first ones would still be selected with our approach, until the total variance explained ratio is greater than a given threshold.
We investigate the following total explained variance ratios: 100\%, 99\%, 95\%, 90\%, 80\%, 70\%, 60\%, 50\%, 40\%, 30\%, 20\%, 10\%, and 5\%.
With our method, the theoretical minimum number of components that are selected is one.
We increase this minimum to two, as performing clustering in a one-dimensional space might be too difficult.

\section{Experimental setup}\label{sec4}

\subsection{Data sets and algorithms}

We use the UCR time series archive \citep{dauUCRTimeSeries2019e} to assess the performance of CLUES-WEASEL.
More precisely, we consider the 128 univariate time series classification data sets from this archive.
Out of the 128 data sets, we exclude the ones with variable length time series (since some algorithms do not work on variable length time series) and the ones with missing values (since all the algorithms cannot natively deal with missing values, and we do not want the missing value imputation strategy to have an impact on the results).
We end up including 112 data sets, which are the data sets used in most benchmarks.
Unless otherwise stated, we use the default training and test splits of the data sets, train the models on the training sets and evaluate their performance on the test sets.
Since CLUES-WEASEL includes randomness, all the reported results are the average of 10 runs with different seeds for the random number generators, similarly to \citep{holderRockKASBABlazingly2024}.
We also independently standardize the time series to have zero mean and unit variance before using them as input to CLUES-WEASEL, similarly to \citep{holderRockKASBABlazingly2024}, even though the debate on this topic is still open \citep{javedBenchmarkStudyTime2020, rakthanmanonAddressingBigData2013}.

We compare CLUES-WEASEL to several other time series clustering algorithms.
Since our setup is very similar to the one in \citep{holderRockKASBABlazingly2024}, we use the results reported in their study and select the best algorithms in different categories.
For our main comparisons, we include the following algorithms: KASBA \citep{holderRockKASBABlazingly2024}, MSM \citep{stefanMoveSplitMergeMetricTime2013, holderReviewEvaluationElastic2024}, PAM-MSM \citep{holderClusteringTimeSeries2023}, DBA \citep{petitjeanGlobalAveragingMethod2011a}, Shape-DBA \citep{ismail-fawazShapeDBAGeneratingEffective2023}, and R-Clustering \citep{jorgeTimeSeriesClustering2024}.
We also perform less extensive comparisons to other algorithms, such as RandomNet \citep{liRandomnetClusteringTime2024} and deep learning algorithms \citep{lafabregueEndtoendDeepRepresentation2022}, since the setups are different.

\subsection{Metrics, hyperparameter optimization and comparisons}

To evaluate the algorithms, we use four clustering metrics (when available).
The first one is clustering accuracy, which is equal to the proportion of correct predictions where each cluster has been assigned to its best matching class.
The second one is the adjusted Rand index, which is a version of the Rand index adjusted for chance.
The third and fourth ones are the adjusted and normalized mutual information, which are versions of the mutual information adjusted for chance and scaled between 0 and 1 respectively.

To perform hyperparameter optimization, we use a development subset of 41 data sets among the 112 data sets, which is the same subset that was used in the development of ROCKET \citep{dempsterROCKETExceptionallyFast2020} and R-Clustering \citep{jorgeTimeSeriesClustering2024}.
In order to obtain the final architecture of CLUES-WEASEL, we compute the clustering accuracy scores for every configuration and for each of the 41 development data sets, compute the mean rank of each combination over the 41 development data sets, and choose the highest ranked combination.
We evaluate the algorithms on the remaining 71 data sets, which we call the evaluation subset, to avoid any unfair advantage given to CLUES-WEASEL.

To compare the algorithms on several data sets, we apply several commonly used tools.
The first one is a modified version of the critical difference diagram \citep{demsarStatisticalComparisonsClassifiers2006a} with the post-hoc Nemenyi test being replaced with a pairwise comparison using the Wilcoxon signed-rank tests and cliques being formed using the Holm correction as recommended in \citep{garciaExtensionStatisticalComparisons2008, benavoliShouldWeReally2016}.
As in \citep{holderRockKASBABlazingly2024}, we use a significance level of 0.05 for all the hypothesis tests.
In the critical difference diagrams, we compare the average ranks (for a chosen metric) of several algorithms, and cliques, represented by solid lines, indicate no significant difference between these algorithms.
We also use the multiple comparison matrix \citep{ismail-fawazApproachMultipleComparison2023} to highlight pairwise comparisons between several algorithms as well as scatter plots for pairwise comparisons between two algorithms.

We also perform small exploratory analyses on the reduced extracted features used as input to the $k$-means algorithm.
To visualize them in a two-dimensional space, we use the Uniform Manifold Approximation and Projection (UMAP) algorithm \citep{mcinnesUMAPUniformManifold2020} as the dimensionality reduction technique.

\subsection{Experiments, implementation and reproducibility}

All the experiments were run on a MacBook Air (M1, 2020) with 16 GB RAM using Python 3.13.11.
We used the following Python packages to implement CLUES-WEASEL:
\begin{itemize}
    \item \emph{aeon} \citep{middlehurstAeonPythonToolkit2024} is a popular Python package dedicated to time series analysis using machine learning. We used the implementation of the transformation step of WEASEL 2.0.
    \item \emph{NumPy} \citep{harrisArrayProgrammingNumPy2020} is a standard Python package for manipulating $n$-dimensional arrays. We used it to compute the cumulative explained variance ratios of PCA.
    \item \emph{scikit-learn} \citep{pedregosaScikitlearnMachineLearning2011} is a popular Python package dedicated to machine learning, providing implementations of many algorithms as well as utility tools. We used it to remove the constant features, to standardize the features, as well as their implementation of PCA and $k$-means.
\end{itemize}

Additionally, to perform all the experiments, including saving the results and generating the figures, we also used the following Python packages:
\begin{itemize}
    \item \emph{Matplotlib} \citep{hunterMatplotlib2DGraphics2007} is a general Python package to create visualizations.
    \item \emph{multi-comp-matrix} \citep{ismail-fawazApproachMultipleComparison2023} provides the implementation of the multiple comparison matrix.
    \item \emph{pandas} \citep{mckinney-proc-scipy-2010} is a data analysis package that makes working on tabular data very convenient.
    \item \emph{seaborn} \citep{waskomSeabornStatisticalData2021} is a data visualization package more specific to data science than \emph{Matplotlib}.
    \item \emph{umap-learn} \citep{mcinnesUMAPUniformManifold2018} is the official implementation of UMAP.
\end{itemize}

The source code is publicly available on a GitHub repository.\footnote{\url{https://github.com/johannfaouzi/CLUES-WEASEL}}
We provide detailed instructions so that our results can be easily reproduced by anyone.

\section{Results}\label{sec5}

We present our results in several sections.
Section~\ref{sec5subsec1} highlights the results of hyperparameter optimization on the 41 development data sets, from which the final architecture of CLUES-WEASEL (described in Figure~\ref{fig3}) is derived.
In Section~\ref{sec5subsec2}, we provide the results of the comparisons between CLUES-WEASEL and existing time series clustering algorithms.
Section~\ref{sec5subsec3} provides the results of the additional comparisons to other time series clustering algorithms.
In Section~\ref{sec5subsec4}, we compare CLUES-WEASEL to other algorithms in terms of runtimes.
Section~\ref{sec5subsec5} highlights the results of our ablation experiments.

\subsection{Hyperparameter optimization}\label{sec5subsec1}

As mentioned in Section~\ref{sec3sub2}, we perform hyperparameter optimization on the number of components extracted by PCA, which is not absolute but based on the total variance explained ratio.
We investigate the following values of total explained variance ratio: 100\%, 99\%, 95\%, 90\%, 80\%, 70\%, 60\%, 50\%, 40\%, 30\%, 20\%, 10\%, and 5\%.

Table~\ref{tab1} presents the results for all the total explained variance ratios across the 41 development data sets in terms of clustering accuracy.
We compute the mean rank and the winning count for each total explained variance ratio.
These results show non-monotonic relationships between the total explained variance ratios and the mean ranks as well as the winning counts.
There is a clear top 3, both in terms of mean ranks and winning counts: 20\% of total explained variance ratio obtains a mean rank of 5.280 and a winning count of 6, while 30\% of total explained variance ratio obtains a mean rank of 5.293 and a winning count of 7, and 40\% of total explained variance ratio obtains a mean rank of 5.305 and a winning count of 5.
We choose the 20\% of total explained variance ratio as it obtains the lowest mean rank.
This choice leads to the final architecture of CLUES-WEASEL, which is highlighted in Figure~\ref{fig3}.

\begin{table}[tbp]
    \caption{Results in terms of clustering accuracy for all the total variance explained ratios investigated across the 41 development data sets.}
    \label{tab1}
    \begin{tabular}{lcc}
        \toprule
        Total variance explained & Mean rank & Winning count \\
        \midrule
        20\% & \textbf{5.280} & 6 \\
        30\% & 5.293 & \textbf{7} \\
        40\% & 5.305 & 5 \\
        50\% & 6.171 & 1 \\
        60\% & 6.537 & 1 \\
        100\% & 6.768 & 0 \\
        10\% & 6.890 & 2 \\
        70\% & 7.122 & 3 \\
        5\% & 7.354 & 1 \\
        80\% & 7.720 & 2 \\
        99\% & 8.232 & 1 \\
        90\% & 9.098 & 2 \\
        95\% & 9.232 & 2 \\
        \bottomrule
    \end{tabular}
\end{table}

\subsection{Comparisons to other time series clustering algorithms with the same setup}\label{sec5subsec2}

In this section, we use the results reported in the KASBA \citep{holderRockKASBABlazingly2024}, which are obtained from the official GitHub repository.\footnote{\url{https://github.com/time-series-machine-learning/tsml-eval/tree/main/tsml_eval/publications/clustering/kasba}}

\subsubsection{With cross-validation}

\begin{figure}
    \begin{subfigure}{0.49\textwidth}
        \includegraphics[width=\textwidth]{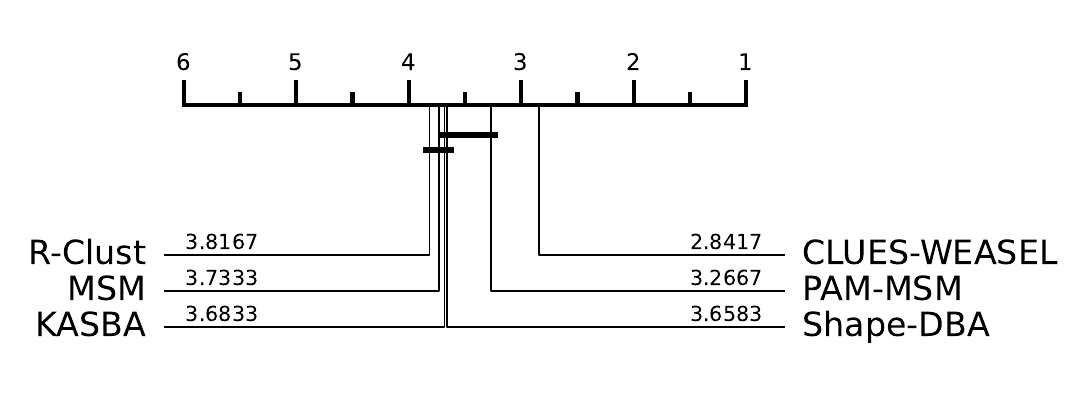}
        \caption{Clustering accuracy}
    \end{subfigure}%
    \hfill
    \begin{subfigure}{0.49\textwidth}
        \includegraphics[width=\textwidth]{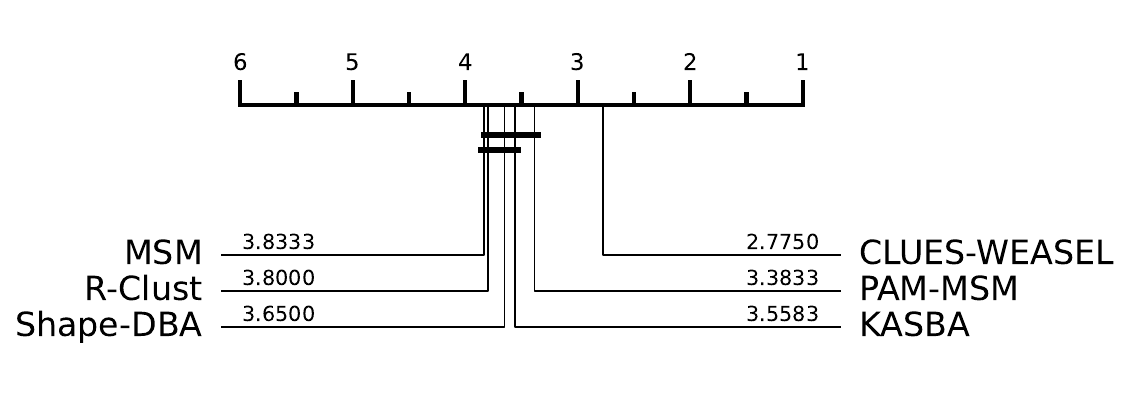}
        \caption{Adjusted Rand index}
    \end{subfigure}%

    \begin{subfigure}{0.49\textwidth}
        \includegraphics[width=\textwidth]{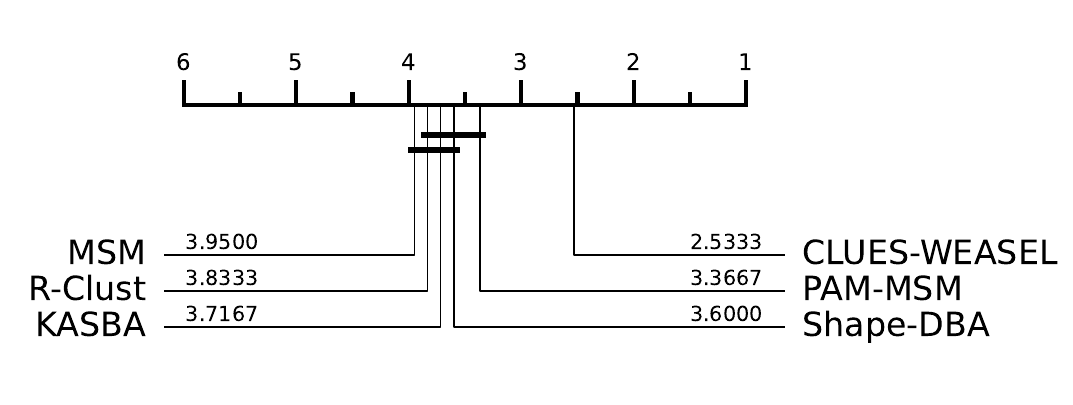}
        \caption{Adjusted mutual information}
    \end{subfigure}%
    \hfill
    \begin{subfigure}{0.49\textwidth}
        \includegraphics[width=\textwidth]{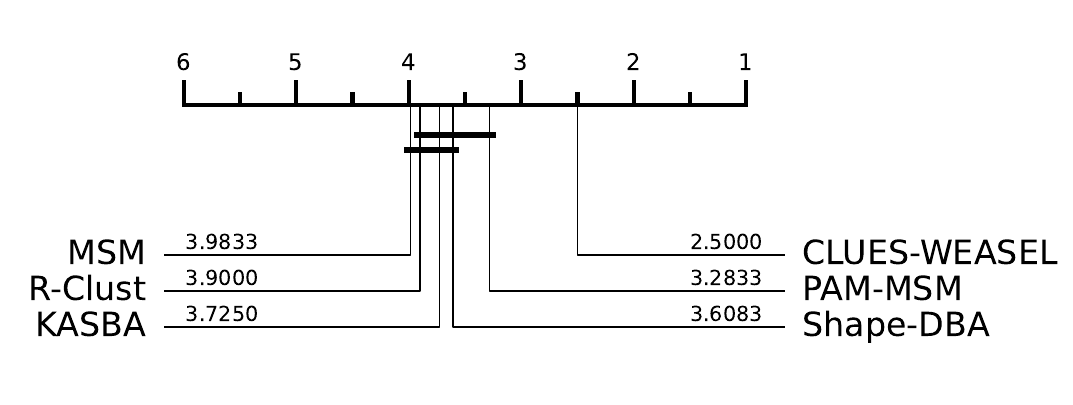}
        \caption{Normalized mutual information}
    \end{subfigure}
    \caption{
        CLUES-WEASEL against five state-of-the-art time series clustering algorithms with cross-validation.
        All the algorithms are trained on the default training sets and evaluated on the default test sets.
        The mean ranks are computed over 60 of the 71 evaluation data sets as the results are missing for 11 data sets for some algorithms because of a too high runtime.
    }
    \label{fig4}
\end{figure}

\begin{figure}
    \begin{subfigure}{0.49\textwidth}
        \includegraphics[width=\textwidth]{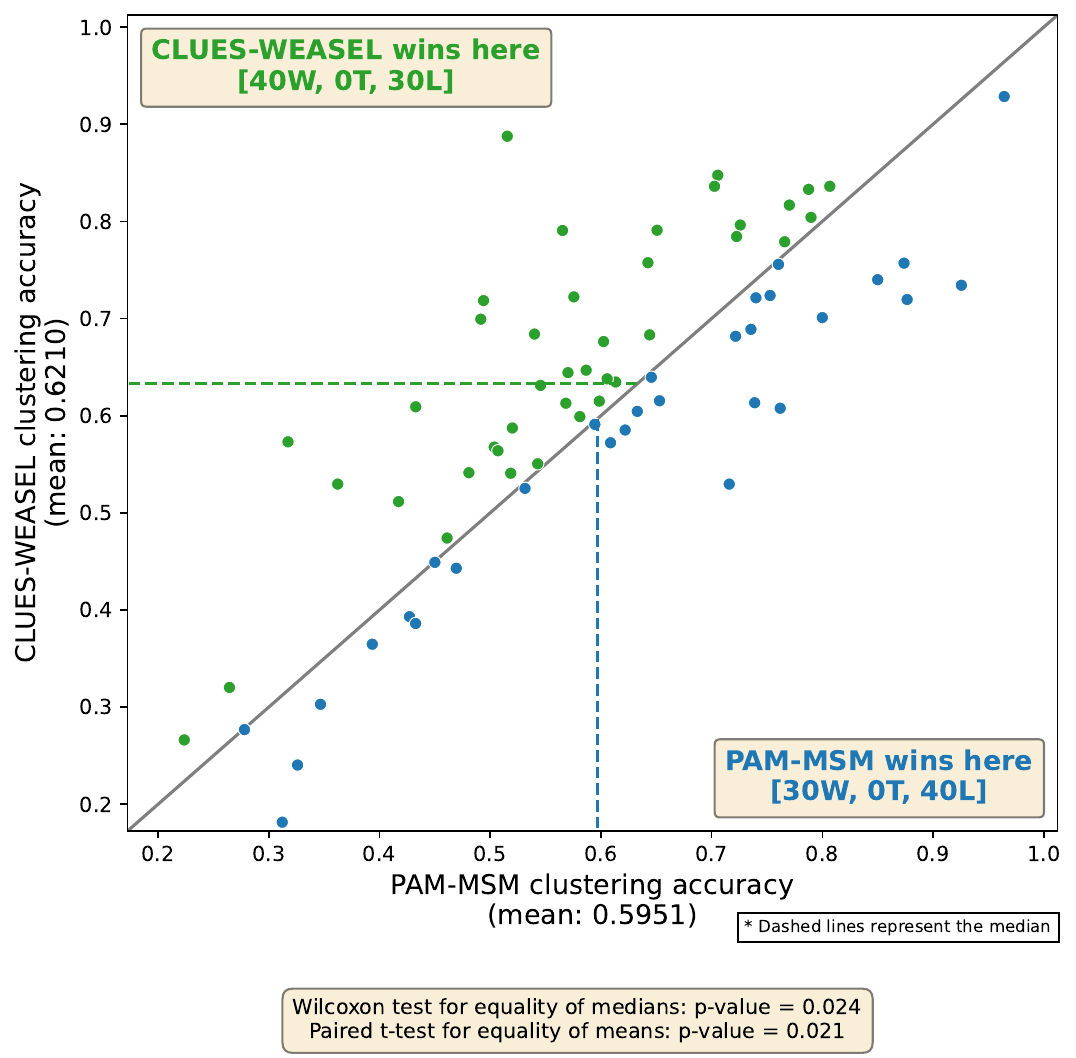}
        \caption{PAM-MSM -- Clustering accuracy}
    \end{subfigure}%
    \hfill
    \begin{subfigure}{0.49\textwidth}
        \includegraphics[width=\textwidth]{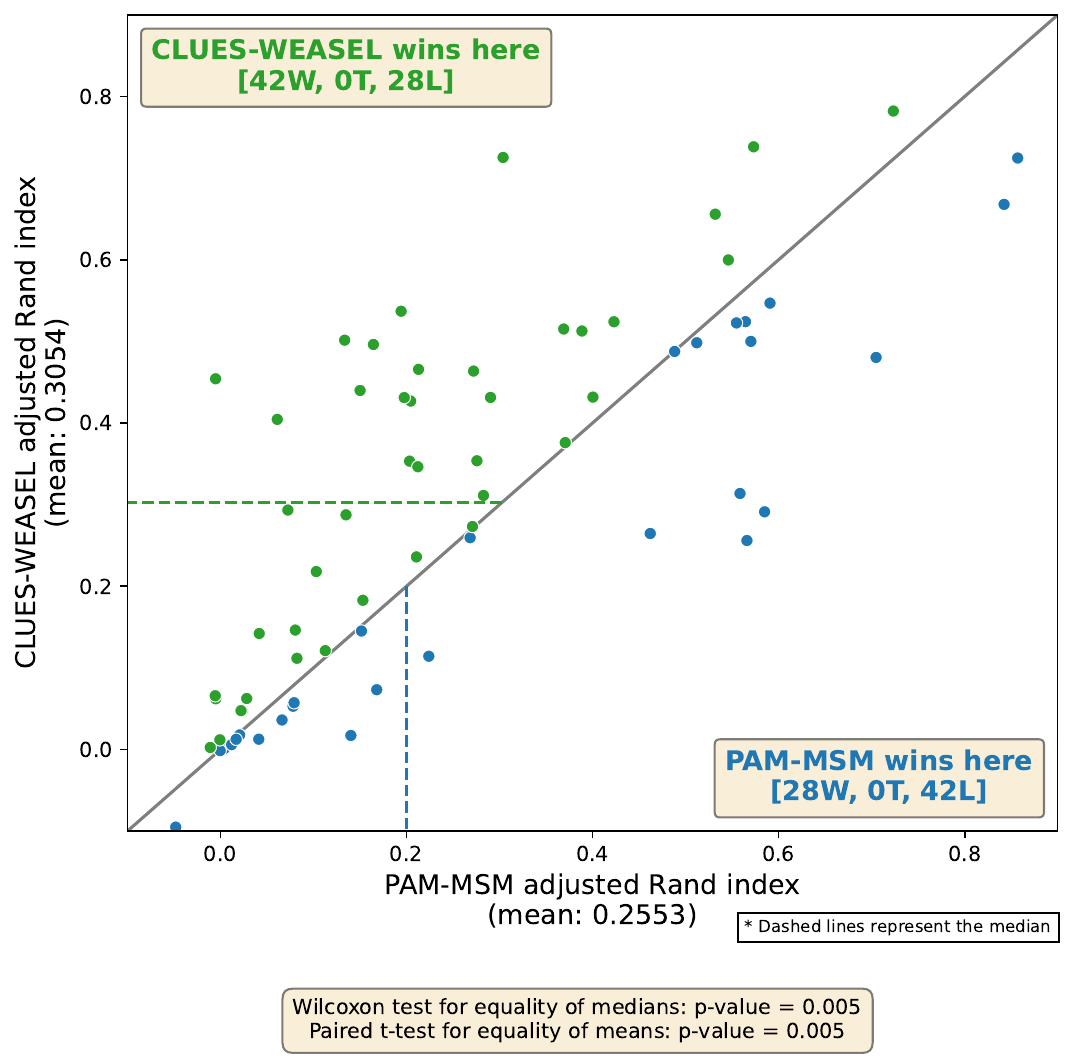}
        \caption{PAM-MSM -- Adjusted Rand index}
    \end{subfigure}%

    \begin{subfigure}{0.49\textwidth}
        \includegraphics[width=\textwidth]{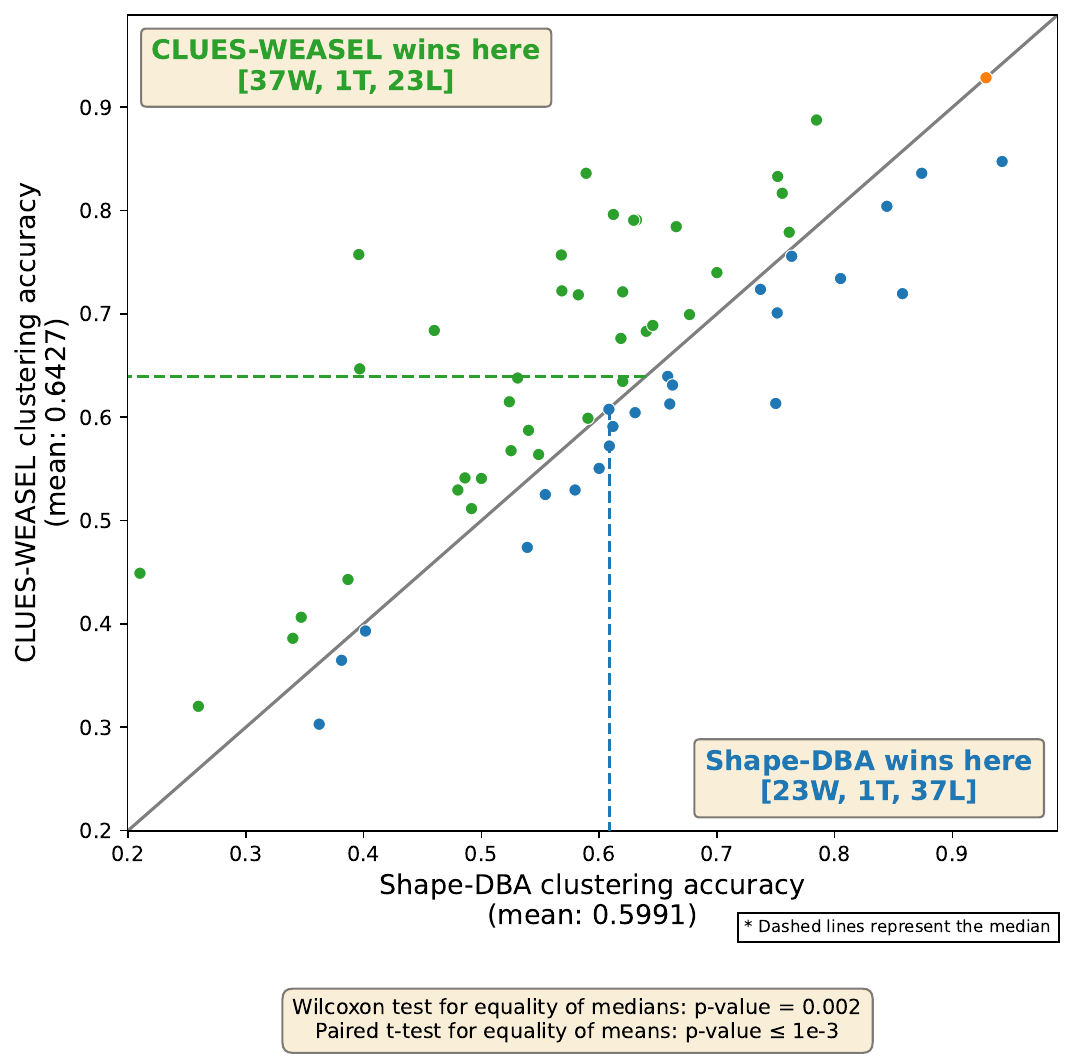}
        \caption{Shape-DBA -- Clustering accuracy}
    \end{subfigure}%
    \hfill
    \begin{subfigure}{0.49\textwidth}
        \includegraphics[width=\textwidth]{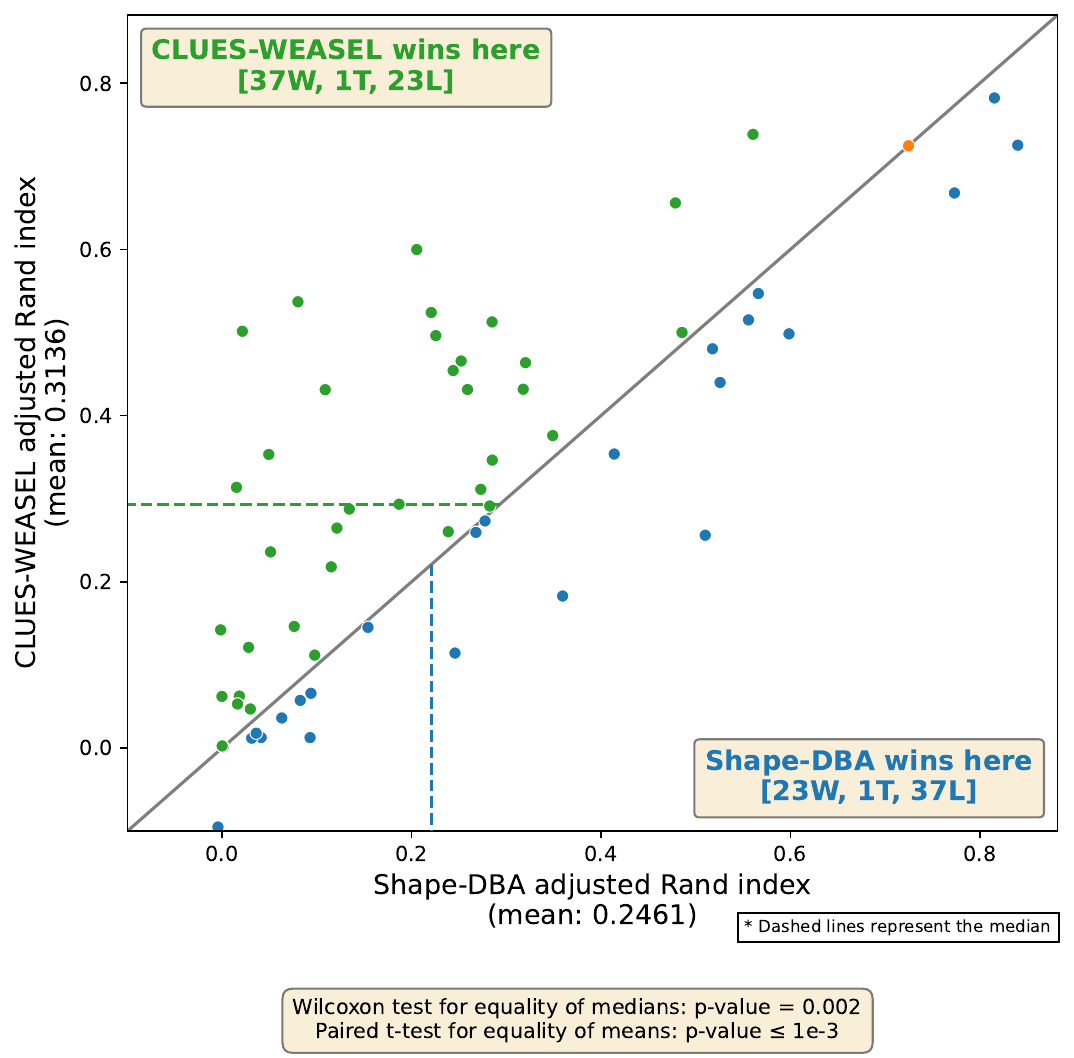}
        \caption{Shape-DBA -- Adjusted Rand index}
    \end{subfigure}%

    \begin{subfigure}{0.49\textwidth}
        \includegraphics[width=\textwidth]{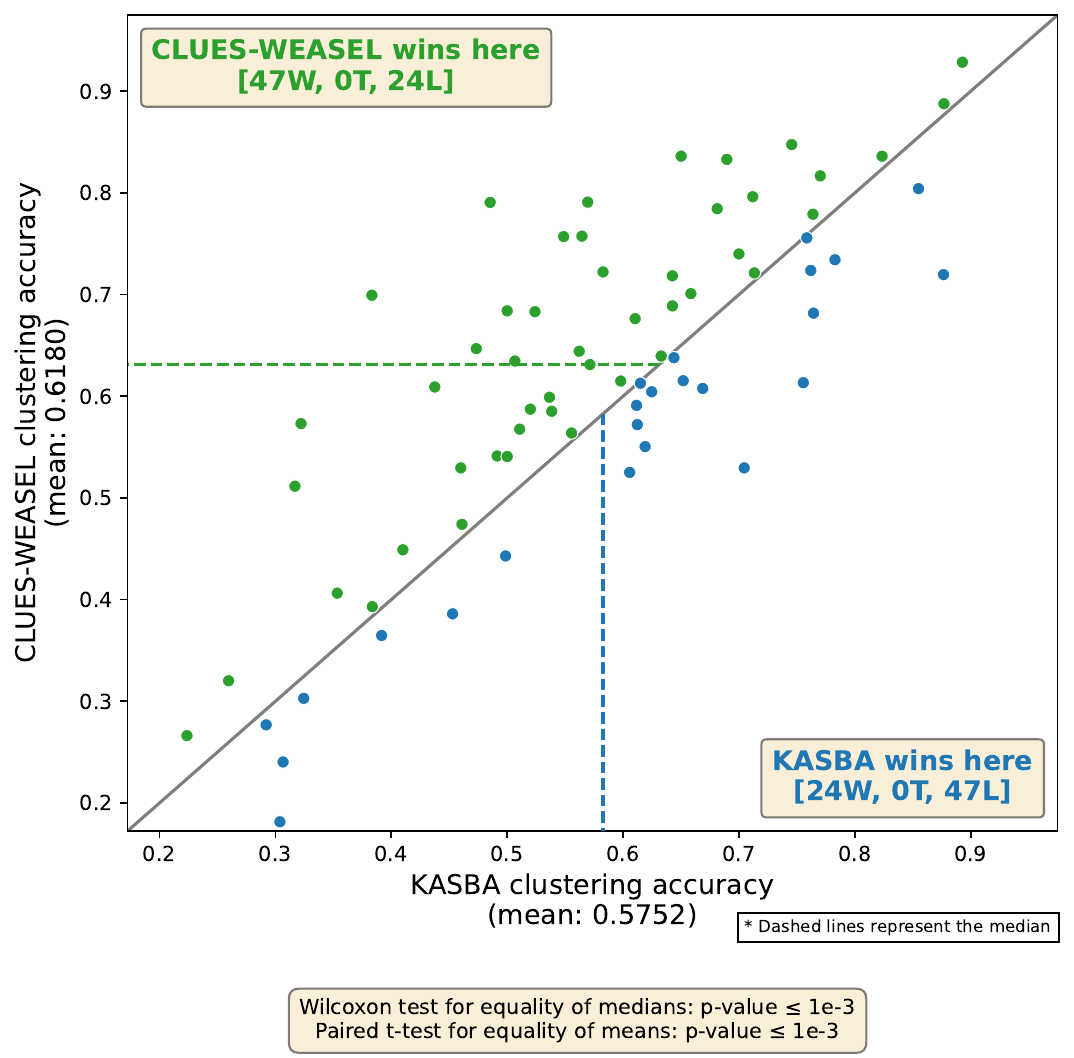}
        \caption{KASBA -- Clustering accuracy}
    \end{subfigure}%
    \hfill
    \begin{subfigure}{0.49\textwidth}
        \includegraphics[width=\textwidth]{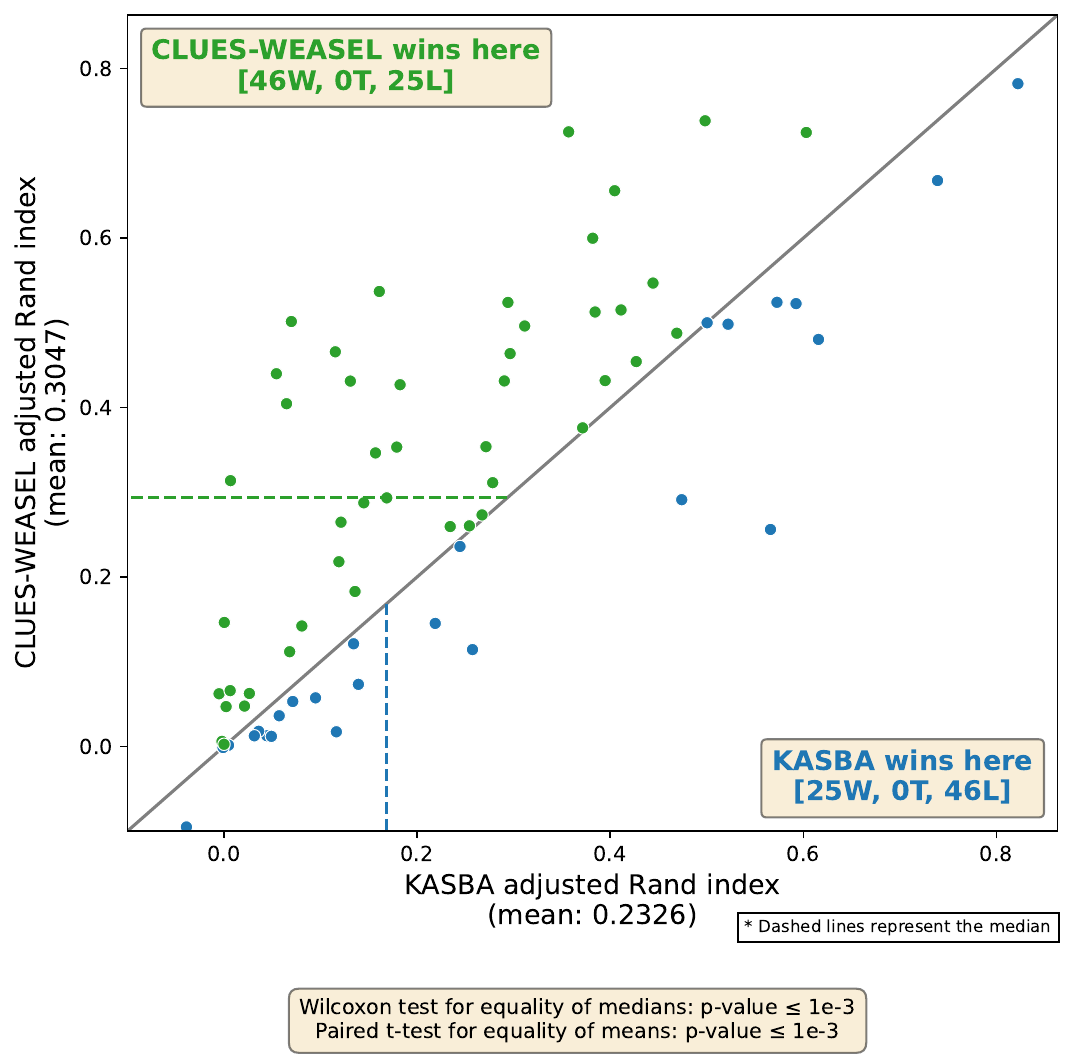}
        \caption{KASBA -- Adjusted Rand index}
    \end{subfigure}%

    \caption{
        CLUES-WEASEL against three state-of-the-art time series clustering algorithms in terms of clustering accuracy and adjusted Rand index.
        All the algorithms are trained on the default training sets and evaluated on the default test sets.
        Out of the 71 evaluation data sets, the results are available for 70 data sets for PAM-MSM, 61 data sets for Shape-DBA, and 71 data sets for KASBA.
    }
    \label{fig5}
\end{figure}

We compare CLUES-WEASEL to five state-of-the-art time series clustering algorithms with cross-validation: KASBA \citep{holderRockKASBABlazingly2024}, MSM \citep{stefanMoveSplitMergeMetricTime2013, holderReviewEvaluationElastic2024}, PAM-MSM \citep{holderClusteringTimeSeries2023}, DBA \citep{petitjeanGlobalAveragingMethod2011a}, Shape-DBA \citep{ismail-fawazShapeDBAGeneratingEffective2023}, and R-Clustering \citep{jorgeTimeSeriesClustering2024}.
All the algorithms are trained on the default train sets and evaluated on the default test sets.
Figure~\ref{fig4} shows the critical difference diagrams comparing the mean ranks of the algorithms for four clustering metrics on the evaluation data sets.
For the four clustering metrics, CLUES-WEASEL has a significantly lower mean rank than all the other algorithms.
PAM-MSM is the second-best algorithm, followed by Shape-DBA or KASBA (depending on the metric).
For more specific comparisons with these three algorithms, we use scatter plots with the predictive performance in terms of clustering accuracy and adjusted Rand index (see Figure~\ref{fig5}).
CLUES-WEASEL has more wins and a higher median score than the three other algorithms for both metrics.
The results are similar when the algorithms are evaluated on all the data sets (see Figure~\ref{figa1}).

\subsubsection{Without cross-validation}

\begin{figure}
    \begin{subfigure}{0.49\textwidth}
        \includegraphics[width=\textwidth]{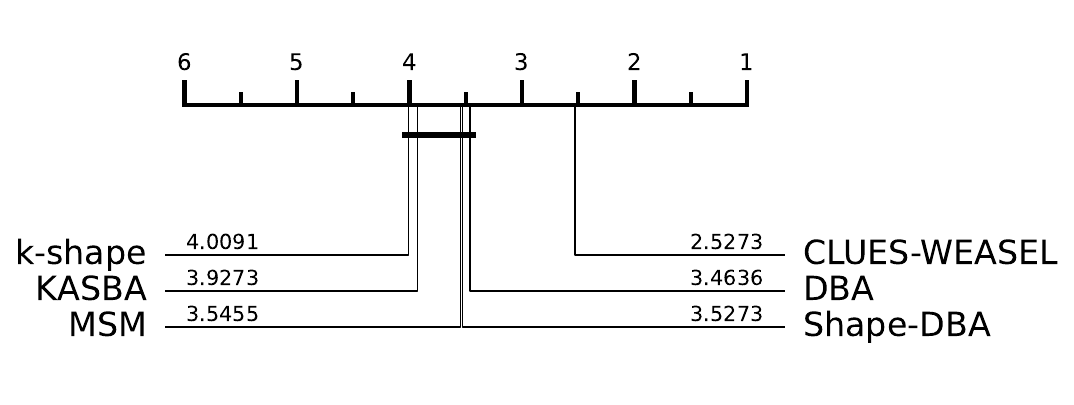}
        \caption{Clustering accuracy}
    \end{subfigure}%
    \hfill
    \begin{subfigure}{0.49\textwidth}
        \includegraphics[width=\textwidth]{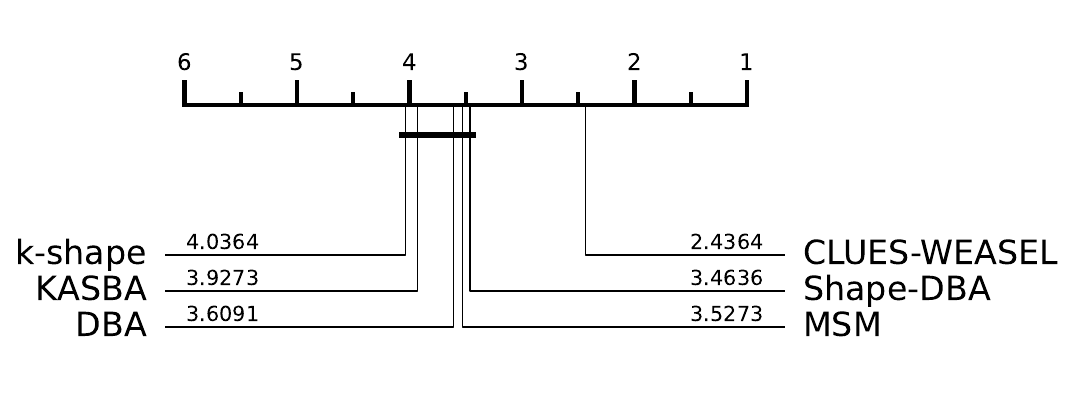}
        \caption{Adjusted Rand index}
    \end{subfigure}%

    \begin{subfigure}{0.49\textwidth}
        \includegraphics[width=\textwidth]{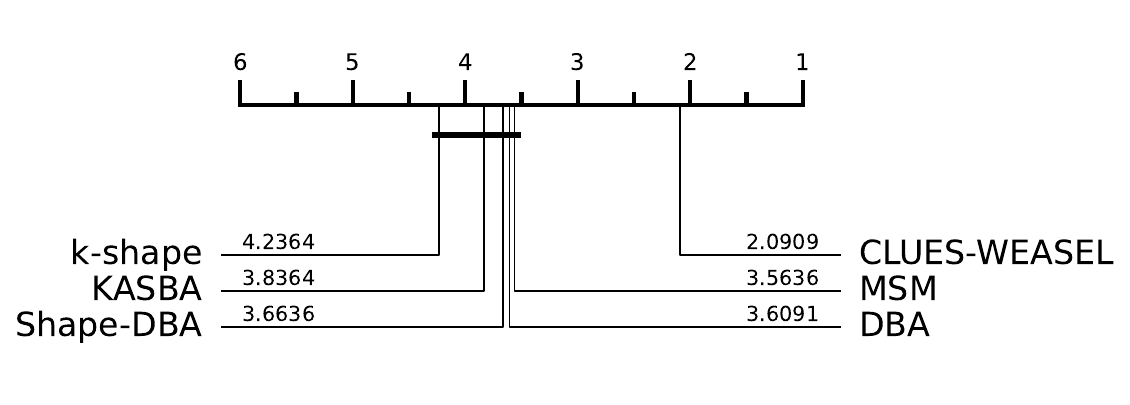}
        \caption{Adjusted mutual information}
    \end{subfigure}%
    \hfill
    \begin{subfigure}{0.49\textwidth}
        \includegraphics[width=\textwidth]{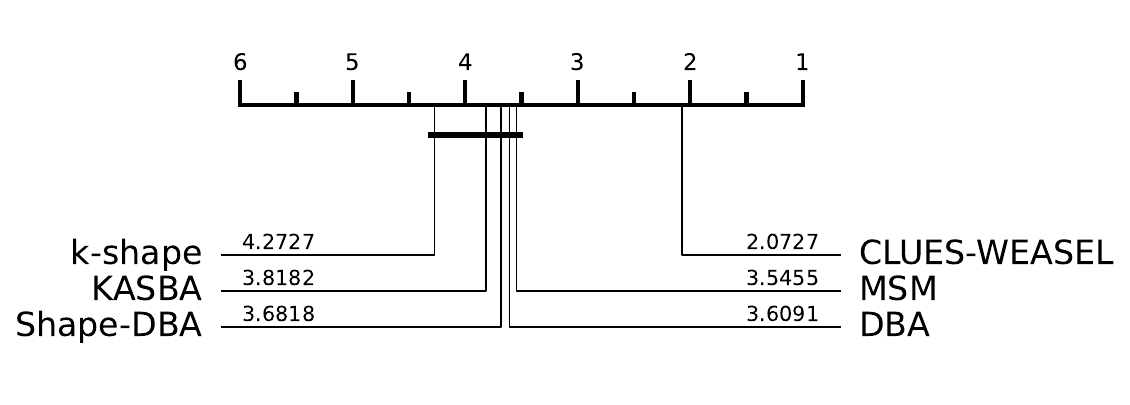}
        \caption{Normalized mutual information}
    \end{subfigure}
    \caption{
        CLUES-WEASEL against five state-of-the-art time series clustering algorithms without cross-validation.
        All the algorithms are trained and evaluated on the merged training and test sets.
        The mean ranks are computed over 55 of the 71 evaluation data sets as the results are missing for 16 data sets for some algorithms because of a too high runtime.
    }
    \label{fig6}
\end{figure}

We compare CLUES-WEASEL to five state-of-the-art time series clustering algorithms without cross-validation: KASBA \citep{holderRockKASBABlazingly2024}, MSM \citep{stefanMoveSplitMergeMetricTime2013, holderReviewEvaluationElastic2024}, DBA \citep{petitjeanGlobalAveragingMethod2011a}, Shape-DBA \citep{ismail-fawazShapeDBAGeneratingEffective2023}, $k$-shape \citep{paparrizosKShapeEfficientAccurate2016}.
All the algorithms are trained and evaluated on the merged training and test sets.
Figure~\ref{fig6} shows the critical difference diagrams comparing the mean ranks of the algorithms for four clustering metrics on the evaluation data sets.
For the four clustering metrics, CLUES-WEASEL has a significantly lower mean rank than all the other algorithms.
Shape-DBA is the second-best algorithm, followed by KASBA, MSM or DBA (depending on the metric).
The results are similar when the algorithms are evaluated on all the data sets (see Figure~\ref{figa2}).

\subsection{Comparisons to other time series clustering algorithms with different setups}\label{sec5subsec3}

\subsubsection{RandomNet}

We compare CLUES-WEASEL to RandomNet \citep{jorgeTimeSeriesClustering2024}, which is a deep neural network with random parameters.
The authors considered the 128 univariate data sets (including the ones with varying length and missing values), used a development subset of 20 data sets, and evaluated the algorithm with the adjusted Rand index.
They trained and evaluated RandomNet on the merged training and test sets (without cross-validation).
They ran RandomNet 10 times and reported the average adjusted Rand index scores.
We use the results available on the official GitHub repository.\footnote{\url{https://github.com/Jackxiini/RandomNet}}

\begin{figure}
    \begin{subfigure}{0.49\textwidth}
        \includegraphics[width=\textwidth]{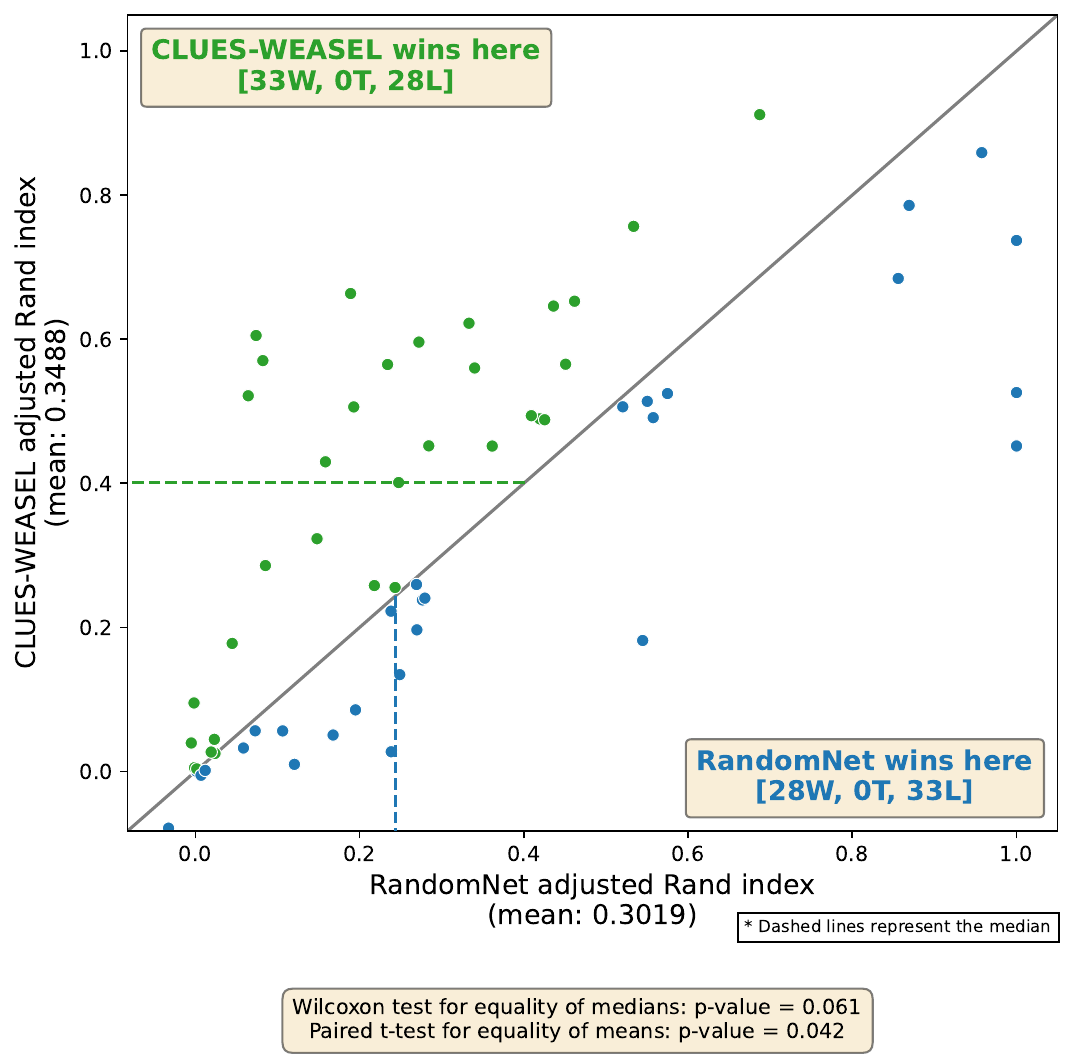}
        \caption{Evaluation data sets}
    \end{subfigure}%
    \hfill
    \begin{subfigure}{0.49\textwidth}
        \includegraphics[width=\textwidth]{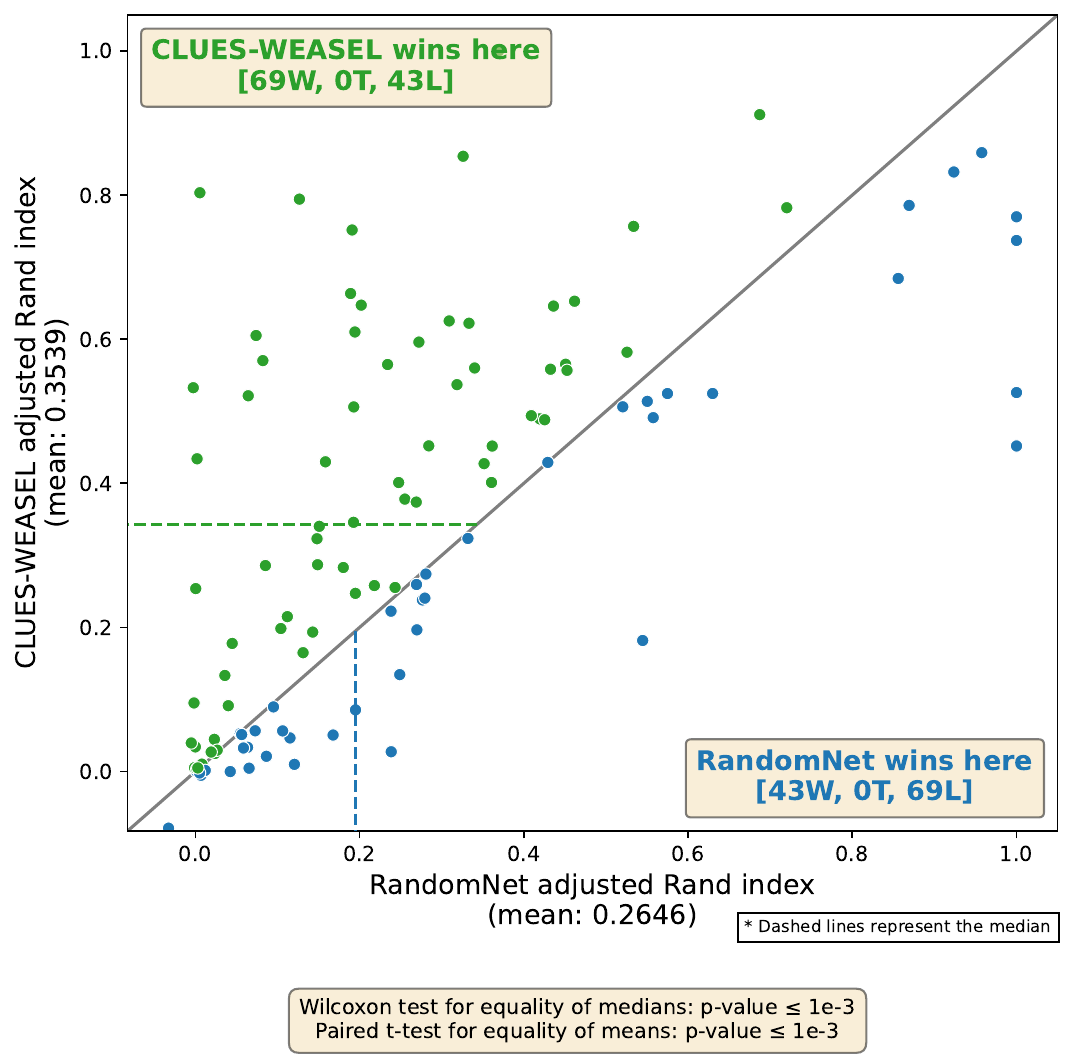}
        \caption{All the data sets}
    \end{subfigure}
    \caption{
        CLUES-WEASEL against RandomNet without cross-validation in terms of adjusted Rand index.
        All the algorithms are trained and evaluated on the merged training and test sets.
        The algorithms are compared on the intersection of the evaluation data sets (left) and on all the data sets (right).
    }
    \label{fig7}
\end{figure}

We compare CLUES-WEASEL and RandomNet on the intersection of the evaluation data sets (61 data sets) and on all the data sets (112 data sets).
Figure~\ref{fig7} shows the corresponding scatter plots.
CLUES-WEASEL is significantly better than RandomNet in both cases.

\subsubsection{Deep learning}

We also compare CLUES-WEASEL to trained deep neural networks using the results of \cite{lafabregueEndtoendDeepRepresentation2022}.
In this study, the authors trained 296 neural networks with different architectures, pretext losses (losses used to train the neural network in a self-supervised fashion), and possibly clustering losses (losses to learn representations that are also suitable for clustering), to learn latent representations.
The clustering task was then performed using the $k$-means algorithm on these learned latent representations.
The authors trained the algorithms on the training sets and evaluated their performance on the test sets.
However, we identified an issue in the source code.\footnote{\url{https://github.com/blafabregue/TimeSeriesDeepClustering/blob/208fac0343d281f2c5997609916004875aae86fd/networks/trainer.py}}
The authors trained the $k$-means algorithm over on the test sets instead of using the centroids learned on the training sets, which is inappropriate and might lead to overestimating the clustering performance of the algorithms.

The authors trained each algorithm five times and reported the mean scores on the test sets.
The results are missing from some combinations of algorithms and data sets due to convergence issues.
The authors used the normalized mutual information as their main metric, but also reported the clustering accuracy scores.
We use the results available on the official GitHub repository.\footnote{\url{https://github.com/blafabregue/TimeSeriesDeepClustering}}

In order to select the best algorithms, we use the following procedure.
We consider both the normalized mutual information (the main metric in their study) and the clustering accuracy (the main metric in our study).
We consider algorithms whose results are available for all the data sets and algorithms whose results are available for at least 92 of the 112 data sets (that is we exclude algorithms with too many missing results as we deem them unreliable).
Finally, we consider the top 3 algorithms in terms of mean scores and mean ranks (thus at least three algorithms and at most six, depending on the intersection of both top 3 rankings).

\begin{figure}
    \begin{subfigure}{0.49\textwidth}
        \includegraphics[width=\textwidth]{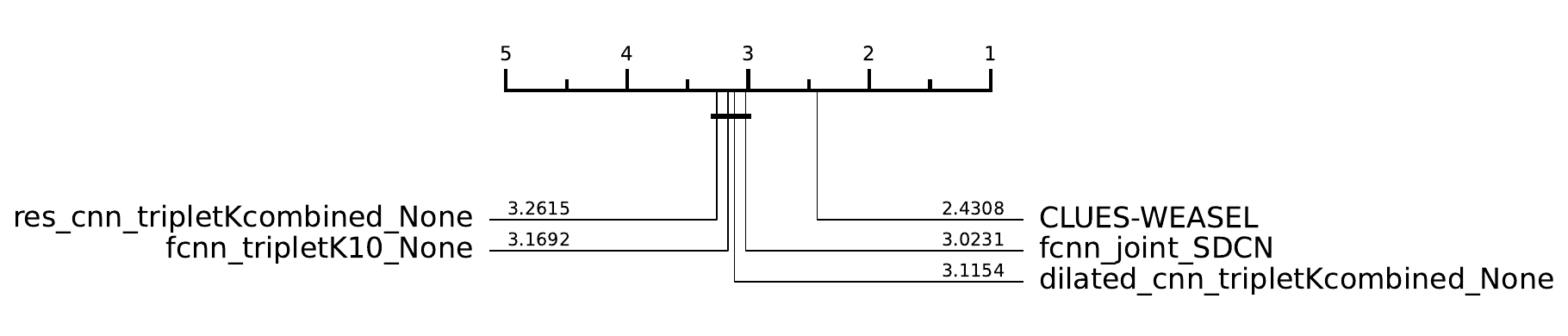}
        \caption{Clustering accuracy - With missing results}
    \end{subfigure}%
    \hfill
    \begin{subfigure}{0.49\textwidth}
        \includegraphics[width=\textwidth]{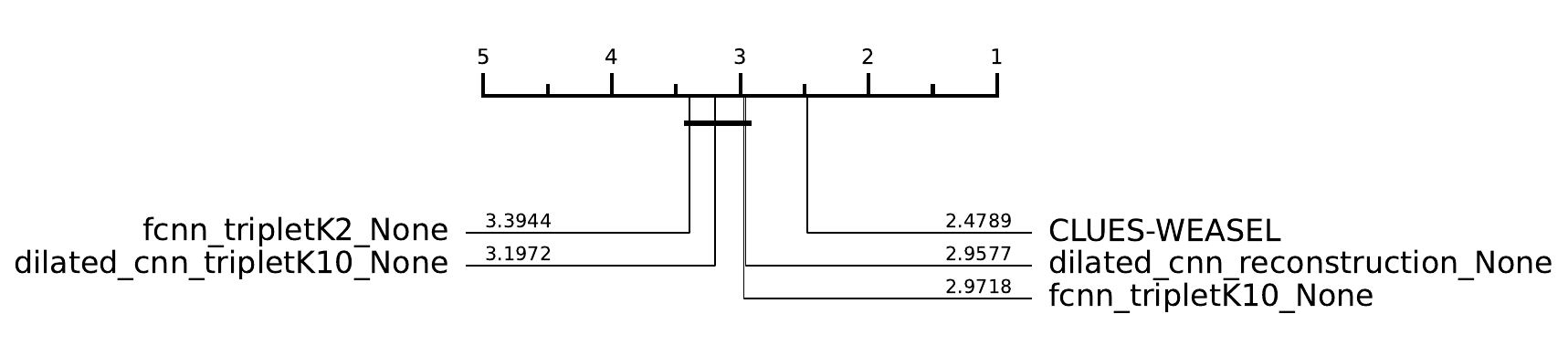}
        \caption{Clustering accuracy - Without missing results}
    \end{subfigure}%

    \begin{subfigure}{0.49\textwidth}
        \includegraphics[width=\textwidth]{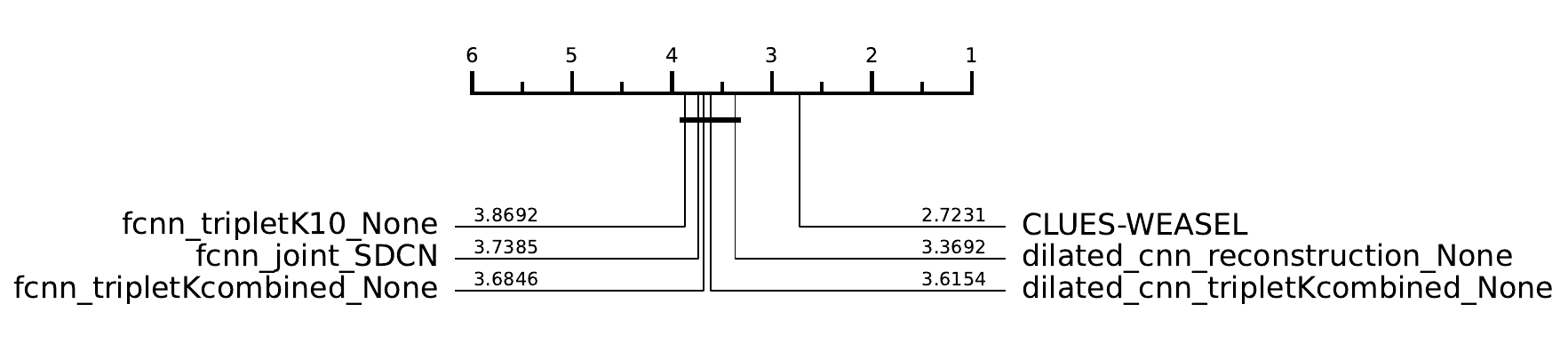}
        \caption{Normalized mutual information - With missing results}
    \end{subfigure}%
    \hfill
    \begin{subfigure}{0.49\textwidth}
        \includegraphics[width=\textwidth]{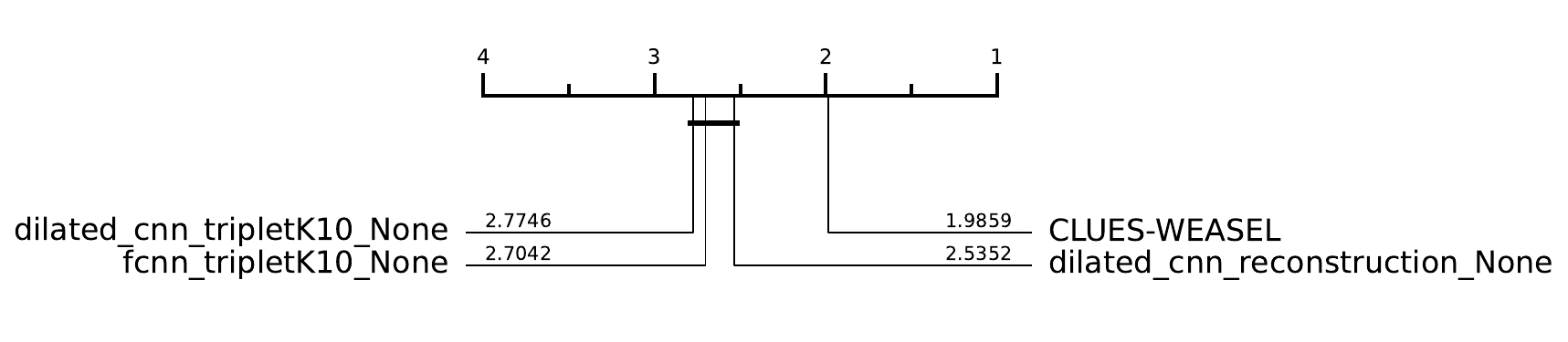}
        \caption{Normalized mutual information - Without missing results}
    \end{subfigure}
    \caption{
        CLUES-WEASEL against state-of-the-art deep learning time series clustering algorithms with cross-validation.
        For each configuration, the best deep learning algorithms have been selected as the union of the top 3 rankings in terms of mean scores and mean ranks.
        All the algorithms are trained on the default training sets and evaluated on the default test sets.
        The mean ranks are computed over the available evaluation data sets.
    }
    \label{fig8}
\end{figure}

Figure~\ref{fig8} shows the critical difference diagrams for the four configurations on the evaluation data sets.
In all the configurations, CLUES-WEASEL is significantly better than all the trained neural networks.
The same conclusion holds when evaluating the algorithms on all the 112 data sets (including the 41 development data sets) (see Figure~\ref{figa3}).

\subsection{Runtimes}\label{sec5subsec4}

Due to limited computational resources, we could not perform exhaustive comparisons in terms of runtimes for existing time series clustering algorithms.
There is currently no faster algorithm than KASBA that is performing better than KASBA \citep{holderRockKASBABlazingly2024}.
Moreover, KASBA is orders of magnitude faster than other algorithms with similar clustering performance such as PAM-MSM and Shape-DBA.
We thus compare CLUES-WEASEL to KASBA in terms of runtimes.

We ran CLUES-WEASEL and KASBA ten times with ten different seeds for the random number generators.
Table~\ref{tab2} presents the results.
CLUES-WEASEL is approximately 2.6 times faster than KASBA while being significantly better KASBA and other time series clustering algorithms that take much longer than KASBA.
The runtimes of CLUES-WEASEL are also very similar across different seeds, with the slowest run taking only 0.68\% more time than the fastest run, compared to 11.94\% for KASBA.

\begin{table}[tbp]
    \caption{
        Runtimes in hours of CLUES-WEASEL and KASBA.
        The runtimes are the sum of the training runtimes on the default training sets and the inference runtimes on the default test sets over the 112 data sets.
        The descriptive statistics are computed over ten sequential runs with ten different seeds for the random number generators.
    }
    \label{tab2}
    \begin{tabular}{lcc}
        \toprule
        & CLUES-WEASEL & KASBA \\
        \midrule
        Minimum & 1.326 & 3.250 \\
        Median & 1.331 & 3.476 \\
        Mean & 1.331 & 3.467 \\
        Maximum & 1.335 & 3.638 \\
        \bottomrule
    \end{tabular}
\end{table}

\subsection{Ablation experiments}\label{sec5subsec5}

We perform two types of ablation experiments.
First, we investigate the use of other unsupervised transformations from algorithms developed for time series classification.
We do not include distance-based approaches as they are not compatible with the architecture of CLUES-WEASEL.
Second, we investigate the maximum number of features extracted by the transformation step of WEASEL 2.0.

\subsubsection{Use of other unsupervised transformations}

For each type of approaches, we select the transformation step of one time series classification algorithm based on the following criteria: ease of using an unsupervised version of the transformation step, and predictive performance on the UCR time series classification archive.
Based on a recent review of time series classification algorithms \citep{middlehurstBakeReduxReview2024}, we choose the transformation steps of the following algorithms: TSFresh \citep{christTimeSeriesFeatuRe2018a} for feature-based approaches, QUANT \citep{dempsterQuantMinimalistInterval2024} for interval-based approaches, RDST \citep{guillaumeRandomDilatedShapelet2022} for shapelet-based approaches, and MultiROCKET \citep{tanMultiRocketMultiplePooling2022} for convolution-based approaches.
We also include Hydra \citep{dempsterHydraCompetingConvolutional2023}, which is both a convolution- and dictionary-based approach.

The architecture for each of these algorithms is the same as in Figure~\ref{fig3}, except that we replace the transformation step of WEASEL 2.0 with the corresponding transformation.
We perform the same type of hyperparameter optimization, that is we investigate the optimal total explained variance ratio for PCA, with the same values.
Table~\ref{tab3} presents the mean ranks (in terms of clustering accuracy) over the 41 development data sets for each transformation.
The optimal values are 100\% for TSFresh, 90\% for MultiROCKET, and 70\% for Hydra, QUANT and RDST.

\begin{table}[tbp]
    \caption{
        Results in terms of clustering accuracy for all the total variance explained ratios investigated across the 41 development data sets for each transformation.
        The lowest mean rank for each transformation is highlighted in bold.
    }
    \label{tab3}
    \begin{tabular}{lccccc}
        \toprule
        Total variance explained & Hydra & MultiROCKET & QUANT & RDST & TSFresh \\
        \midrule
        100\% & 6.341 & 5.768 & 6.793 & 6.268 & \textbf{5.622} \\
        99\% & 6.195 & 5.878 & 6.805 & 6.488 & 6.854 \\
        95\% & 5.854 & 5.610 & 6.524 & 6.012 & 6.561 \\
        90\% & 6.463 & \textbf{5.549} & 6.732 & 6.183 & 8.073 \\
        80\% & 6.220 & 5.768 & 5.866 & 6.341 & 6.463 \\
        70\% & \textbf{5.720} & 6.500 & \textbf{5.476} & \textbf{5.195} & 6.122 \\
        60\% & 6.561 & 6.622 & 6.195 & 6.012 & 6.878 \\
        50\% & 7.171 & 8.451 & 6.902 & 6.256 & 7.317 \\
        40\% & 8.256 & 8.073 & 8.146 & 7.951 & 6.232 \\
        30\% & 8.146 & 8.195 & 7.854 & 8.500 & 7.427 \\
        20\% & 8.024 & 8.195 & 7.902 & 8.622 & 7.817 \\
        10\% & 8.024 & 8.195 & 7.902 & 8.585 & 7.817 \\
        5\% & 8.024 & 8.195 & 7.902 & 8.585 & 7.817 \\
        \bottomrule
    \end{tabular}
\end{table}

\begin{figure}
    \begin{subfigure}{0.49\textwidth}
        \includegraphics[width=\textwidth]{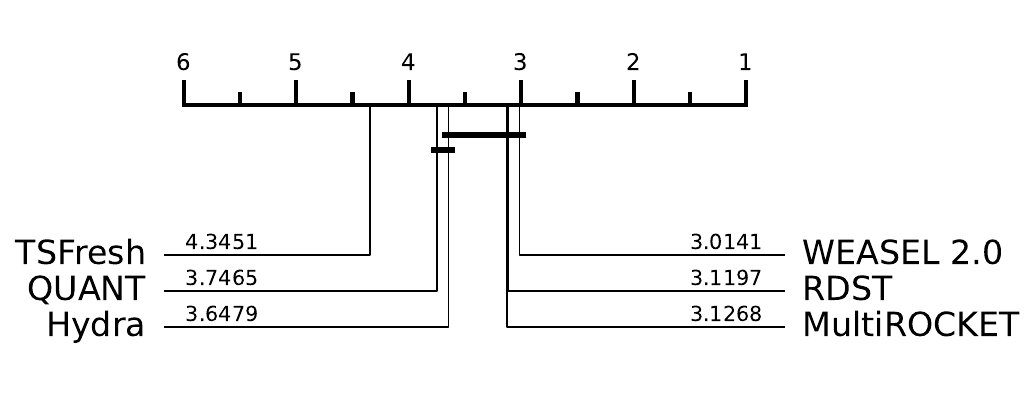}
        \caption{Clustering accuracy}
    \end{subfigure}%
    \hfill
    \begin{subfigure}{0.49\textwidth}
        \includegraphics[width=\textwidth]{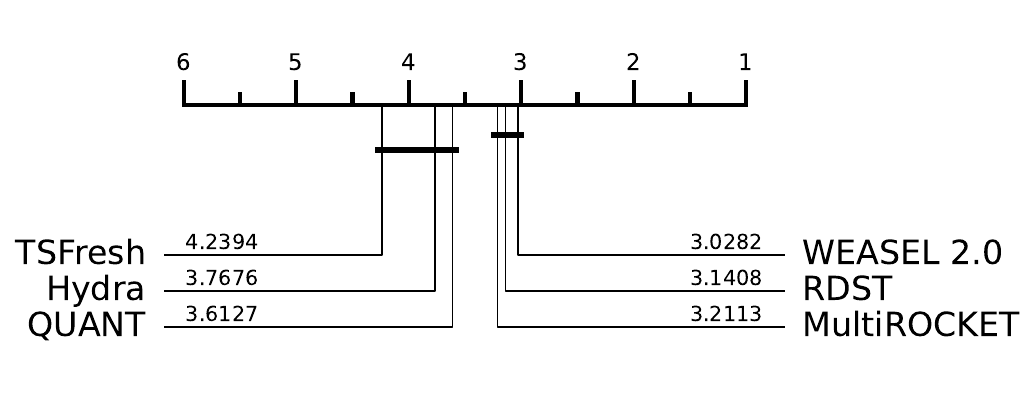}
        \caption{Adjusted Rand index}
    \end{subfigure}%

    \begin{subfigure}{0.49\textwidth}
        \includegraphics[width=\textwidth]{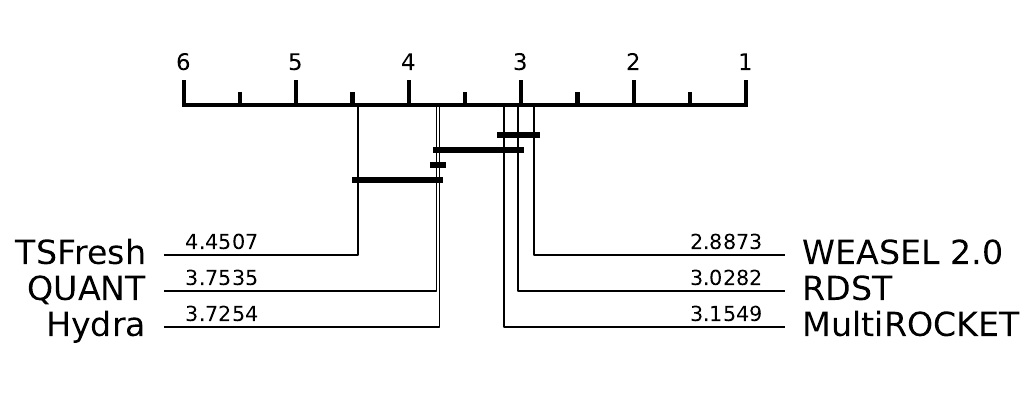}
        \caption{Adjusted mutual information}
    \end{subfigure}%
    \hfill
    \begin{subfigure}{0.49\textwidth}
        \includegraphics[width=\textwidth]{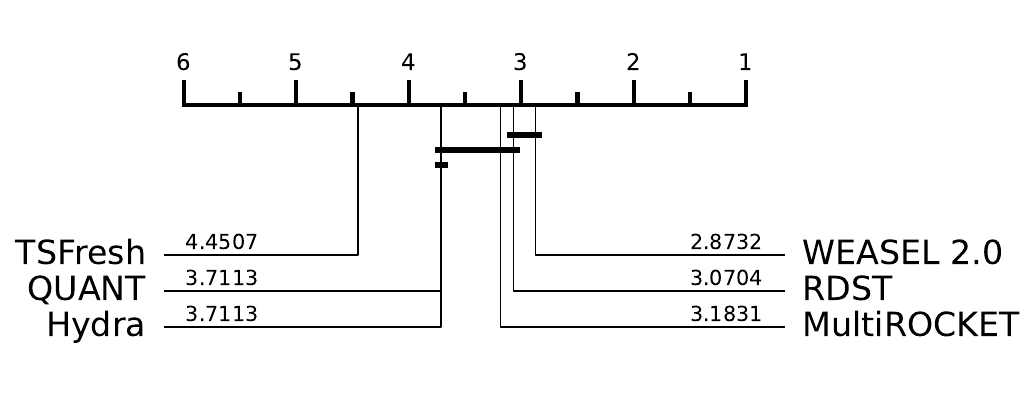}
        \caption{Normalized mutual information}
    \end{subfigure}
    \caption{
        WEASEL 2.0 against five transformation steps of state-of-the-art time series classification algorithms.
        The architectures of these time series clustering algorithms is the same as the one depicted in Figure~\ref{fig3}, except that the second step is replaced with the corresponding transformation and that the total variance explained ratio has been optimized for each transformation.
        All the algorithms are trained on the default training sets and evaluated on the default test sets.
        The mean ranks are computed over the 71 evaluation data sets.
    }
    \label{fig9}
\end{figure}

\begin{figure}
    \includegraphics[width=\textwidth]{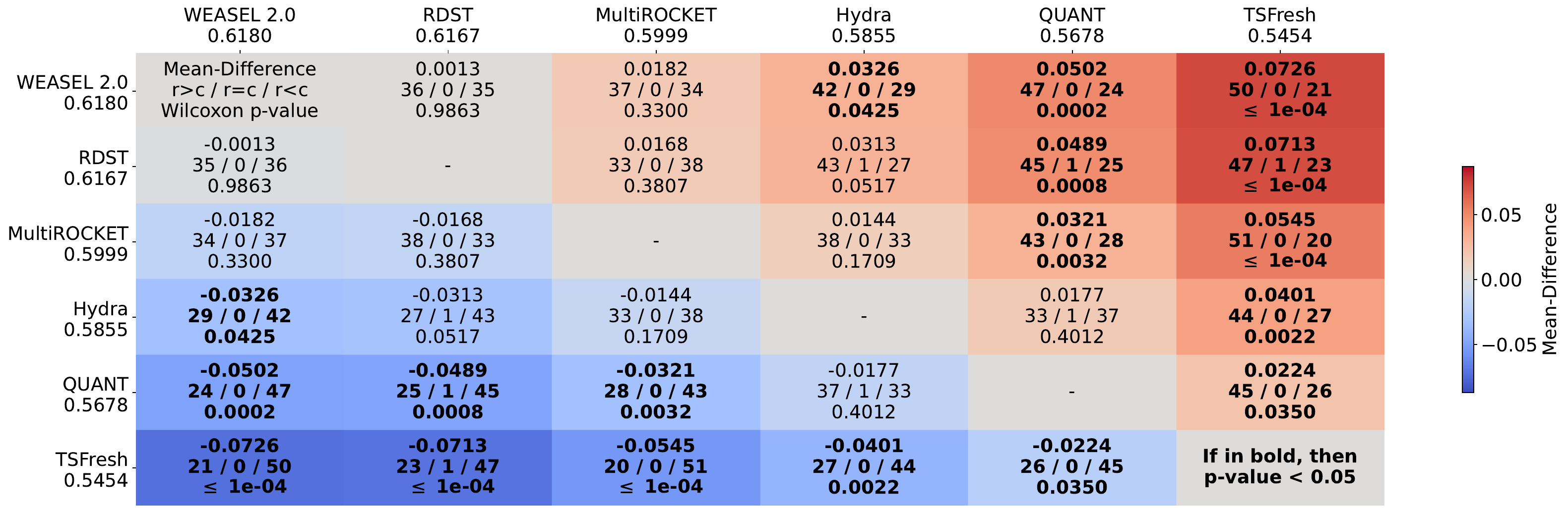}
    \caption{
        Multiple comparison matrix for 6 unsupervised transformation algorithms in terms of clustering accuracy on the 71 evaluation data sets.
        All the algorithms are trained on the default training sets and evaluated on the default test sets.
        In each cell, the mean difference (top), the numbers of wins/ties/losses (middle), and the $p$-value (unadjusted for multiple comparisons) for a one-sided Wilcoxon sign rank test are highlighted.
        Bold values indicate significant differences (at the confidence level of 0.05).
    }
    \label{fig10}
\end{figure}

Using these optimal values, we evaluate the algorithms on the 71 evaluation data sets.
Figure~\ref{fig9} shows the critical difference diagrams comparing the mean ranks of the algorithms for four clustering metrics.
The transformation step of WEASEL 2.0 has the lowest mean rank for all the metrics, followed by RDST, MultiROCKET, Hydra, QUANT, and TSFresh.
WEASEL 2.0 is only significantly better than Hydra, QUANT and TSFresh.
The multiple comparison matrix (see Figure~\ref{fig10}) confirms these results.
The same conclusions hold when evaluating the algorithms on the 112 data sets (including the 41 development data sets) (see Figures~\ref{figa4} and \ref{figa5}).

\begin{figure}
    \begin{subfigure}{0.49\textwidth}
        \includegraphics[width=\textwidth]{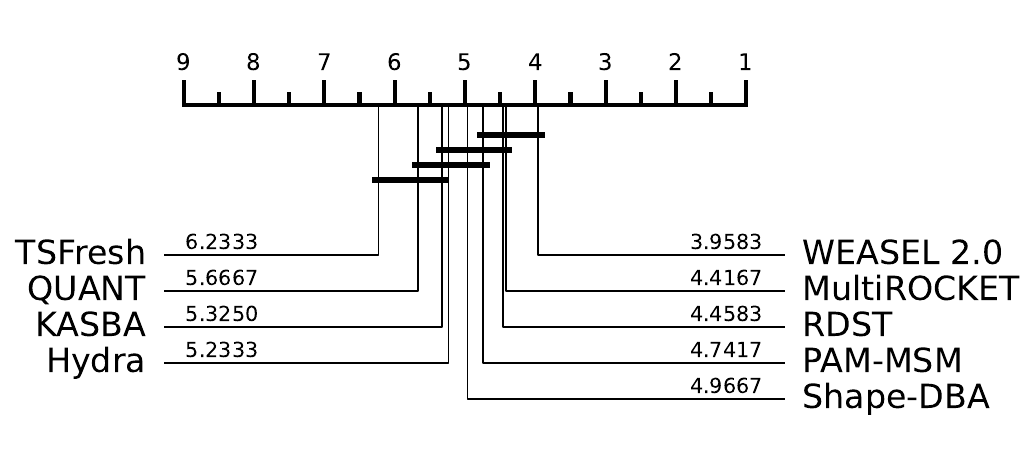}
        \caption{Clustering accuracy}
    \end{subfigure}%
    \hfill
    \begin{subfigure}{0.49\textwidth}
        \includegraphics[width=\textwidth]{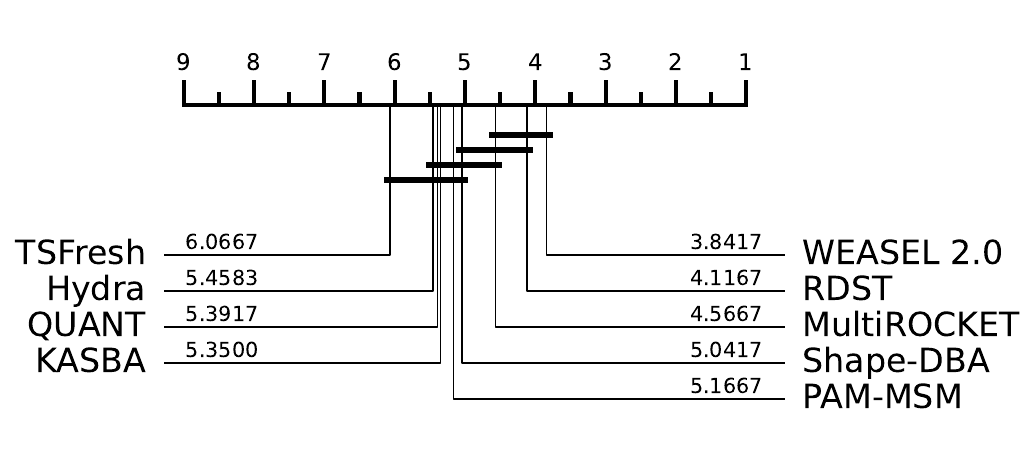}
        \caption{Adjusted Rand index}
    \end{subfigure}%

    \begin{subfigure}{0.49\textwidth}
        \includegraphics[width=\textwidth]{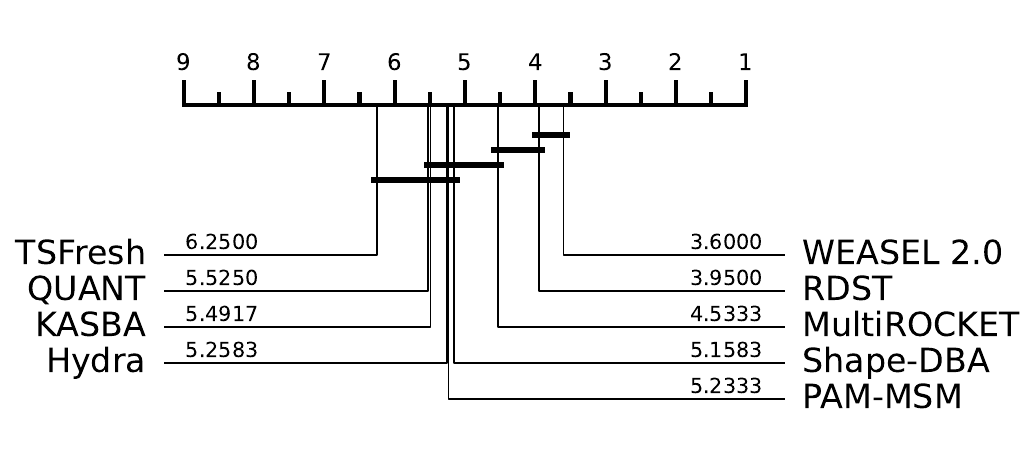}
        \caption{Adjusted mutual information}
    \end{subfigure}%
    \hfill
    \begin{subfigure}{0.49\textwidth}
        \includegraphics[width=\textwidth]{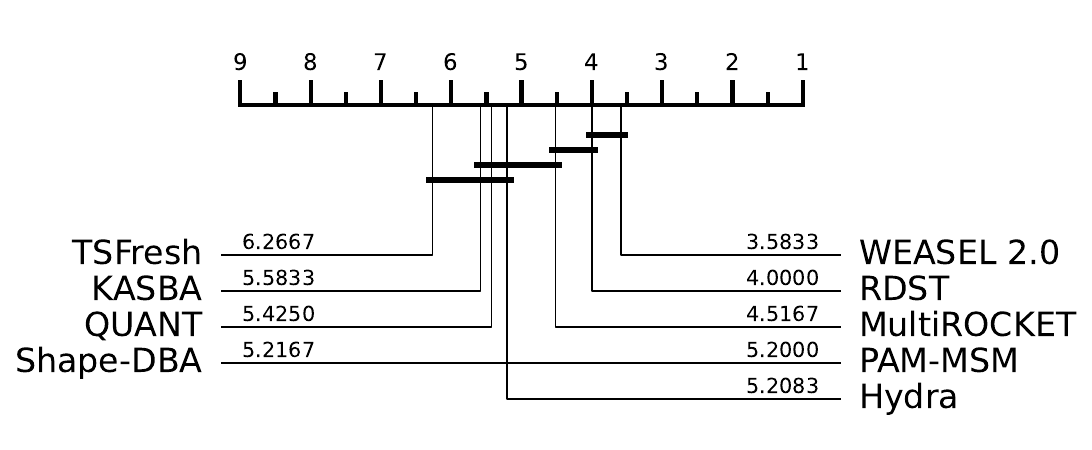}
        \caption{Normalized mutual information}
    \end{subfigure}
    \caption{
        WEASEL 2.0 and five transformation steps of state-of-the-art time series classification algorithms against three state-of-the-art time series clustering algorithms.
        The architectures for these six transformations is the same as the one depicted in Figure~\ref{fig3}, except that the second step is replaced with the corresponding transformation and that the total variance explained ratio has been optimized for each transformation.
        These six algorithms are compared to three state-of-the-art time series clustering algorithms.
        All the algorithms are trained on the default training sets and evaluated on the default test sets.
        The mean ranks are computed over 60 of the 71 evaluation data sets as the results are missing for 11 data sets for some algorithms because of a too high runtime.
    }
    \label{fig11}
\end{figure}

We compare these six new time series clustering algorithms to three existing state-of-the-art time series clustering algorithms.
Figure~\ref{fig11} shows the critical difference diagrams comparing the mean ranks of the algorithms for four clustering metrics.
Our architecture with the transformation steps of WEASEL 2.0, RDST and MultiROCKET is very competitive compared to these state-of-the-art algorithms, with higher mean ranks and significant differences for some metrics.
The results with Hydra, QUANT and TSFresh are less positive, with these algorithms having the highest mean ranks in most cases.
The same conclusions hold when evaluating the algorithms on the 112 data sets (including the 41 development data sets) (see Figure~\ref{figa6}).

\subsubsection{Impact of the maximum feature count}

The transformation step of WEASEL 2.0 extracts at most 30,000 features by default, which is also the value used in CLUES-WEASEL.
We investigate the use of lower values for this hyperparameter, namely 10,000, 5,000, 1,000, 500, and 200.
We perform the same hyperparameter optimization procedure on the total explained variance ratio.
Table~\ref{tab4} presents the results, which are consistent with the ones observed with the default values of 30,000: the highest ranks are obtained with total variance explained ratios of 40\%, 30\%, and 20\% to a lesser degree.
We select the total variance explained ratio corresponding to the lowest mean rank and obtain the following values: 40\% for a maximum feature count of 10,000, and 30\% for maximum feature counts of 10,000, 5,000, 1,000, and 200.

\begin{table}[tbp]
    \caption{
        Results in terms of clustering accuracy for all the total variance explained ratios (rows) and all the values of the maximum number of features (columns) of CLUES-WEASEL investigated across the 41 development data sets.
        The lowest mean rank for each transformation is highlighted in bold.
        }
    \label{tab4}
    \begin{tabular}{lccccc}
        \toprule
        Total variance explained & 10k & 5k & 1k & 500 & 200 \\
        \midrule
        100\% & 7.122 & 6.695 & 7.780 & 8.000 & 7.720 \\
        99\% & 7.683 & 7.085 & 7.537 & 8.098 & 7.463 \\
        95\% & 8.902 & 9.220 & 7.866 & 8.280 & 7.890 \\
        90\% & 9.110 & 8.524 & 8.610 & 7.988 & 7.878 \\
        80\% & 7.622 & 8.402 & 7.683 & 6.963 & 7.268 \\
        70\% & 7.378 & 6.939 & 6.232 & 6.610 & 6.280 \\
        60\% & 6.451 & 6.585 & 6.390 & 6.390 & 6.134 \\
        50\% & 6.000 & 5.707 & 5.659 & 5.415 & 5.988 \\
        40\% & \textbf{5.061} & 5.476 & 5.305 & 5.341 & 5.707 \\
        30\% & 5.110 & \textbf{5.098} & \textbf{5.049} & \textbf{4.500} & \textbf{5.159} \\
        20\% & 5.756 & 5.659 & 5.890 & 6.378 & 6.317 \\
        10\% & 7.183 & 7.561 & 8.195 & 8.390 & 8.415 \\
        5\% & 7.622 & 8.049 & 8.805 & 8.646 & 8.780 \\
        \bottomrule
    \end{tabular}
\end{table}

\begin{figure}
    \begin{subfigure}{0.49\textwidth}
        \includegraphics[width=\textwidth]{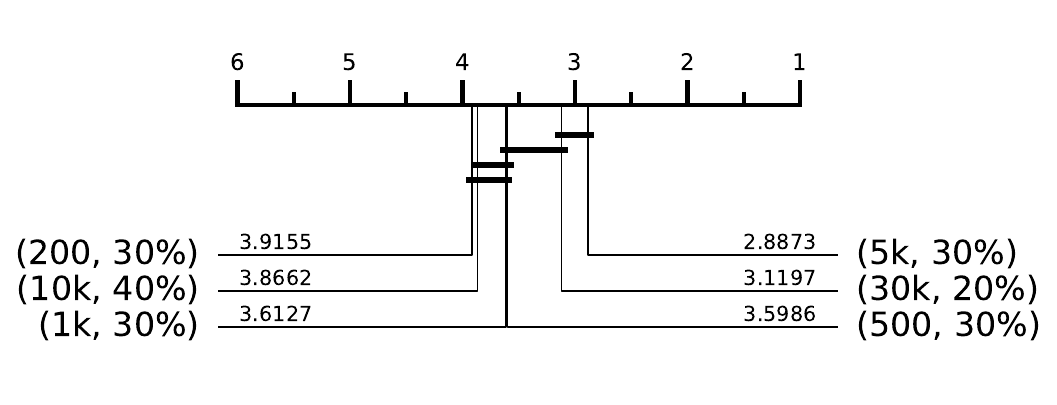}
        \caption{Clustering accuracy}
    \end{subfigure}%
    \hfill
    \begin{subfigure}{0.49\textwidth}
        \includegraphics[width=\textwidth]{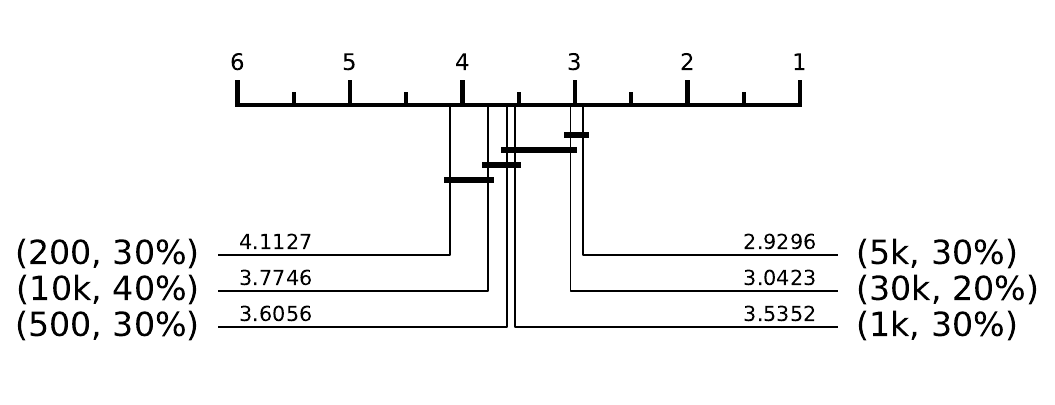}
        \caption{Adjusted Rand index}
    \end{subfigure}%

    \begin{subfigure}{0.49\textwidth}
        \includegraphics[width=\textwidth]{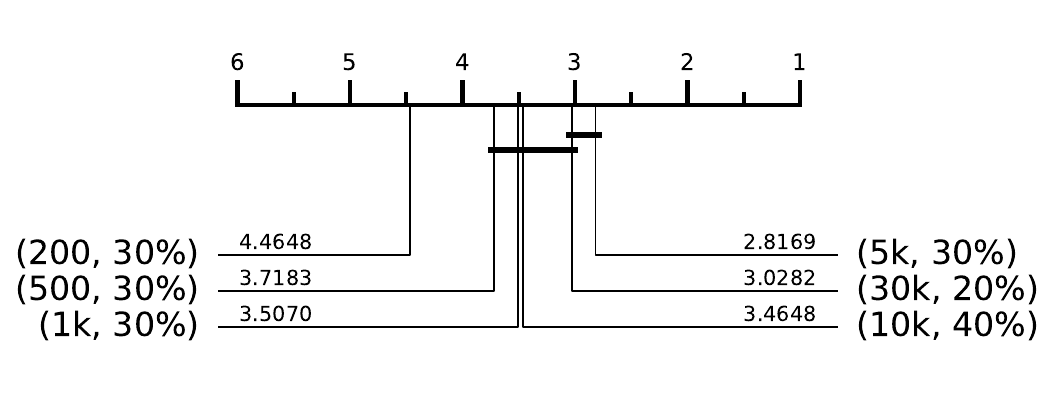}
        \caption{Adjusted mutual information}
    \end{subfigure}%
    \hfill
    \begin{subfigure}{0.49\textwidth}
        \includegraphics[width=\textwidth]{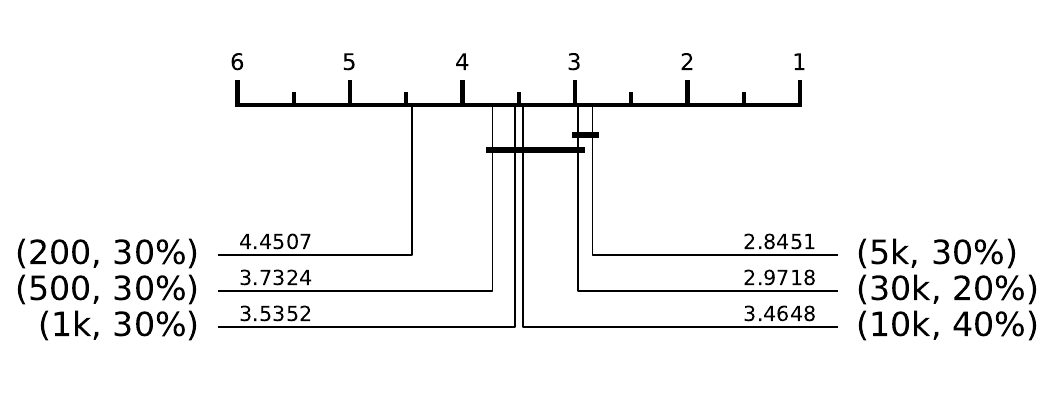}
        \caption{Normalized mutual information}
    \end{subfigure}
    \caption{
        CLUES-WEASEL with different maximum feature counts (first value inside the brackets) and total variance explained ratios (second value inside the brackets).
        All the versions are trained on the default training sets and evaluated on the default test sets.
        The mean ranks are computed over the 71 evaluation data sets.
    }
    \label{fig12}
\end{figure}

\begin{figure}
    \includegraphics[width=\textwidth]{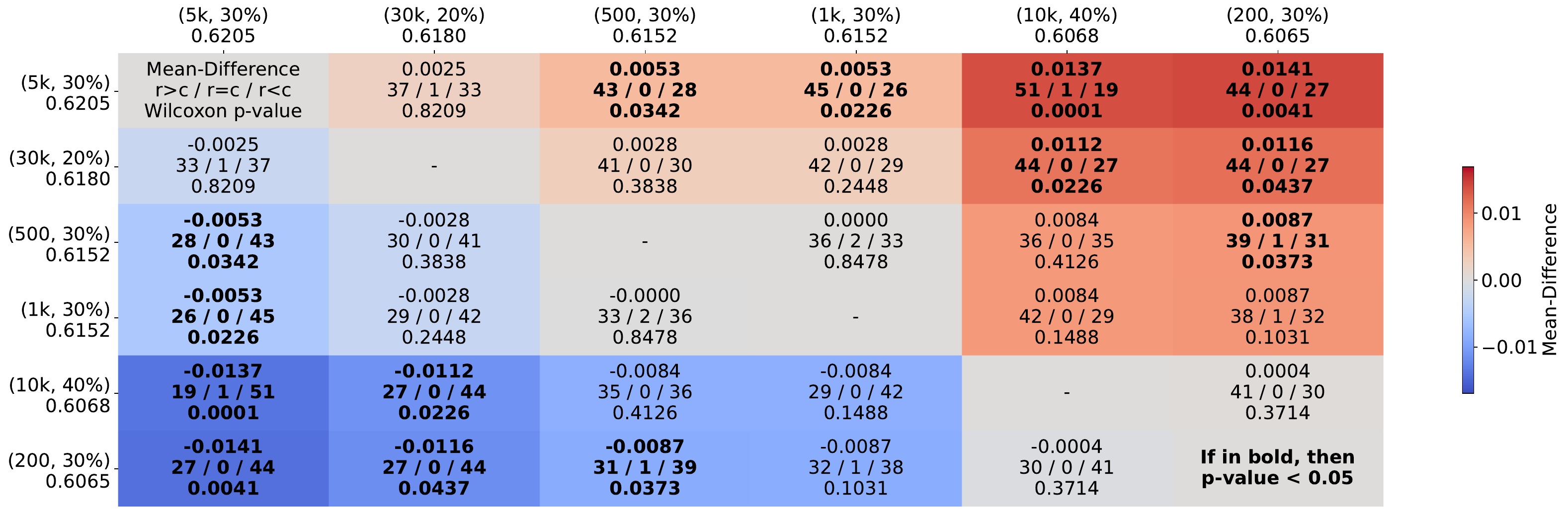}
    \caption{
        Multiple comparison matrix for CLUES-WEASEL with different maximum feature counts (first value inside the brackets) and total variance explained ratios (second value inside the brackets) in terms of clustering accuracy on the 71 evaluation data sets.
        All the versions are trained on the default training sets and evaluated on the default test sets.
        In each cell, the mean difference (top), the numbers of wins/ties/losses (middle), and the $p$-value (unadjusted for multiple comparisons) for a one-sided Wilcoxon sign rank test are highlighted.
        Bold values indicate significant differences (at the confidence level of 0.05).
    }
    \label{fig13}
\end{figure}

Using these optimal values, we evaluate CLUES-WEASEL with different maximum feature counts (and different total variance explained ratios) on the 71 evaluation data sets.
Figure~\ref{fig12} show the critical difference diagrams comparing the mean ranks of these versions for four clustering metrics.
There is clear top 2, with CLUES-WEASEL(5k, 30\%) and CLUES-WEASEL(30k, 20\%) having the lowest and second-lowest mean ranks respectively, although the difference is not significant.
However, CLUES-WEASEL(5k, 30\%) is significantly better than all the other configurations except CLUES-WEASEL(30k, 20\%).
The multiple comparison matrix (see Figure~\ref{fig13}) confirms these results.
The same conclusions hold when evaluating the algorithms on the 112 data sets (including the 41 development data sets) (see Figures~\ref{figa7} and \ref{figa8}).

\section{Exploratory analyses}\label{sec6}

In this section, we perform short exploratory analyses on the reduced extracted features by our architecture with different feature extraction algorithms.
To visualize the results in 2D, we have to potentially perform an additional dimension reduction step if the feature space has a larger dimension.
We use the Uniform Manifold Approximation and Projection (UMAP) algorithm \citep{mcinnesUMAPUniformManifold2020} to project the features in a two-dimensional space.
All the algorithms are trained on the default training sets and the visualized reduced extracted features are computed on the default test sets.

\begin{figure}
    \includegraphics[width=\textwidth]{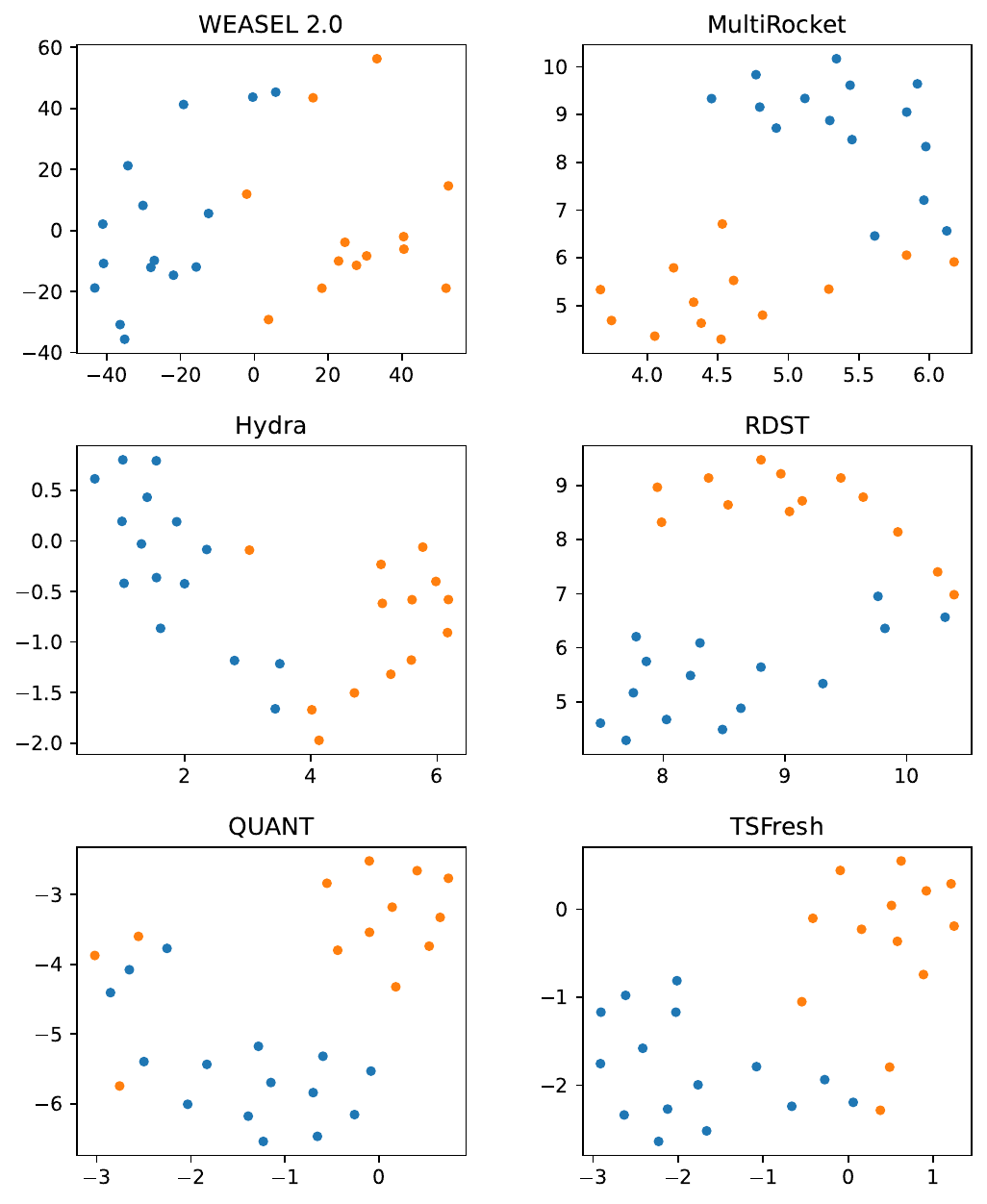}
    \caption{
        Visualization of the reduced extracted features by different algorithms on the Coffee data set.
        All the algorithms are trained on the default train set, and the figure shows the visualization of the reduced extracted features on the test set in a two-dimensional space using UMAP.
        This data set has two classes and the colors highlight the true classes.
        }
    \label{fig14}
\end{figure}

Figure~\ref{fig14} shows the results for the Coffee data set.
This data set consists of 2 classes with 28 training samples and 28 test samples.
It is a very simple data set for which most time series classification algorithms have perfect predictive performance.
The results show that the samples are not really clustered into two distinct groups for any algorithm, but samples from the same classes tend to be closer to each other.

\begin{figure}
    \includegraphics[width=\textwidth]{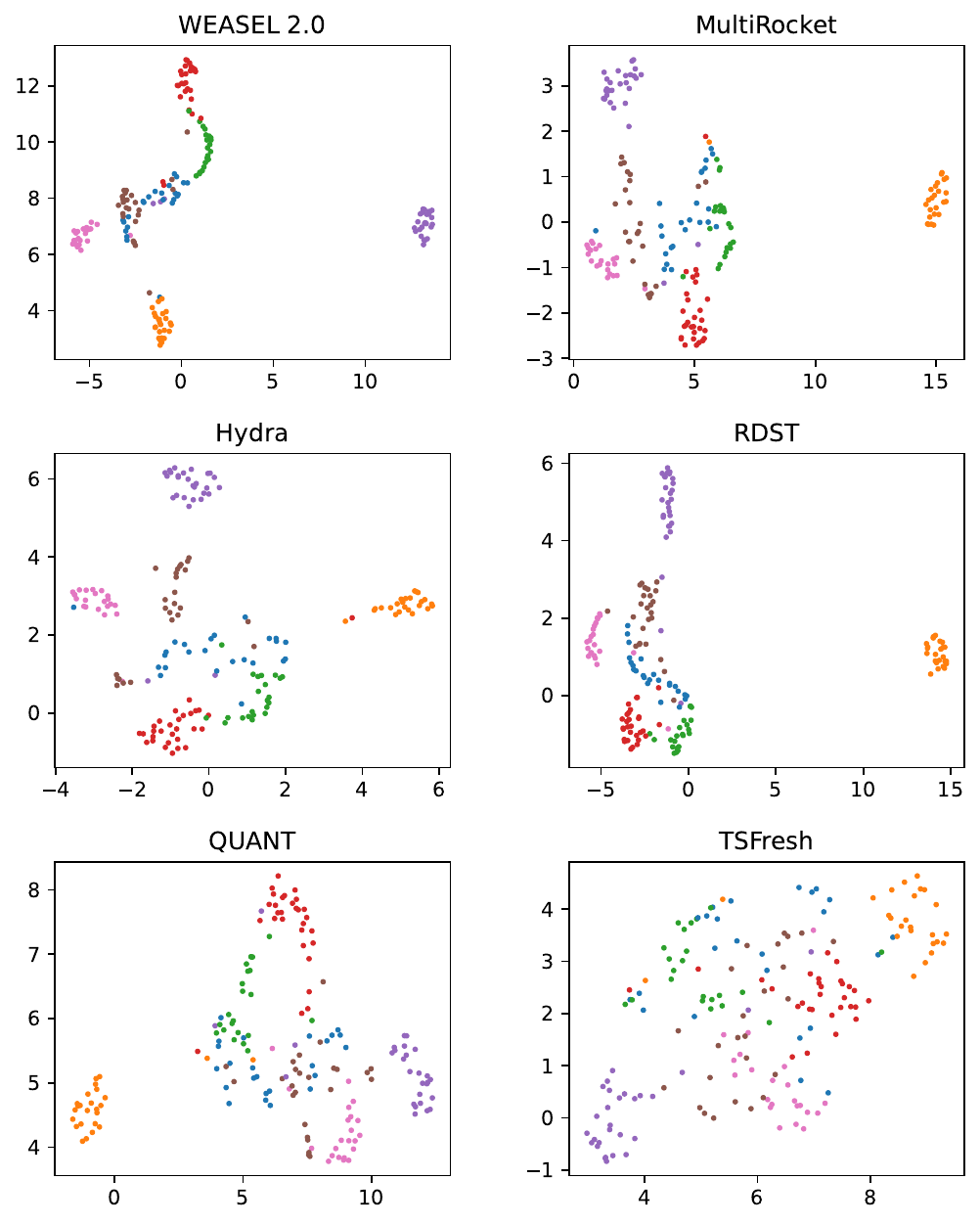}
    \caption{
        Visualization of the reduced extracted features by different algorithms on the Fish data set.
        All the algorithms are trained on the default train set, and the figure shows the visualization of the reduced extracted features on the test set in a two-dimensional space using UMAP.
        This data set has seven classes and the colors highlight the true classes.
        }
    \label{fig15}
\end{figure}

Figure~\ref{fig15} shows the results for the Fish data set.
This data set consists of 7 classes with 175 training samples and 175 test samples.
It is also a simple data set for which most time series classification algorithms have very high predictive performance.
The results show that the samples are clustered into groups that are correlated to the true classes for all the algorithms except TSFresh (and QUANT to a lesser extent).

Additional visualization results for other data sets are available in appendices (see Figures~\ref{figb9}, \ref{figb10} and \ref{figb11}).

\section{Conclusion}\label{sec7}

We have presented CLUES-WEASEL, a novel time series clustering algorithm incorporating feature extraction using the unsupervised version of the transformation step of WEASEL 2.0, dimensionality reduction using principal component analysis, and clustering using $k$-means.
The evaluation of CLUES-WEASEL on the UCR time series classification archive has shown that CLUES-WEASEL is significantly better than any existing time series clustering algorithm, while being much faster than other state-of-the-art algorithms.
We have proved that the architecture of CLUES-WEASEL can be used with other feature extraction algorithms, namely RDST and MultiROCKET, to obtain state-of-the-art clustering performance.

Our work still has several limitations.
CLUES-WEASEL has a non-negligible memory footprint, especially for large data sets, due to the complexity of principal component analysis (linear in terms of number of samples and quadratic in terms of number of features).
This memory footprint might be reduced with the use of fewer extracted features from the transformation step of WEASEL 2.0, as our ablation experiments have shown that it is possible to obtain the similar clustering performance with 5,000 features at most instead of 30,000.
Nonetheless, principal component analysis has the advantage of being an extremely fast dimensionality reduction technique.
We also select the number of components based on the total variance explained ratio.
The current implementation of CLUES-WEASEL uses the PCA implementation of \emph{scikit-learn}: the whole PCA is performed and the number of components is selected afterwards.
We could investigate the use of an iterative version of PCA that is stopped when the total variance explained ratio is above the threshold, but a potential lower runtime is not guaranteed.
CLUES-WEASEL would also benefit from being benchmarked on other data sets, as WEASEL 2.0 was also benchmarked on the UCR time series classification archive.
However, this remark also applies to other time series clustering algorithms.

In conclusion, CLUES-WEASEL is a very powerful time series clustering algorithm.
It is the current state-of-the-art in terms of clustering performance while being really fast, at the cost of relatively high memory footprint for large data sets.
We believe that CLUES-WEASEL is a valuable addition to the list of time series clustering algorithms.

\section{Declarations}

All the data sets used in this study are publicly available from the University of California, Riverside Time Series Archive \citep{dauUCRTimeSeries2019e}.
They can be downloaded in several ways, either manually from the website or automatically using libraries such as the \emph{aeon} \citep{middlehurstAeonPythonToolkit2024} Python package.
The whole source code supporting this study is publicly available on a GitHub repository,\footnote{\url{https://github.com/johannfaouzi/CLUES-WEASEL}} with detailed instructions to reproduce all the experiments.
The results are also directly available, so that other researchers can easily compare their algorithms with ours.
The repository is under the BSD 3-Clause License, making it reusable by other researchers.

No funding was received for conducting this study.
The authors have no relevant financial or non-financial interests to disclose.
The authors have no conflicts of interest to declare that are relevant to the content of this article.
The authors certify that they have no affiliations with or involvement in any organization or entity with any financial interest or non-financial interest in the matter or materials discussed in this manuscript.
The authors have no financial or proprietary interests in any material discussed in this article.

\newpage

\begin{appendices}

\section{Additional results}\label{secA1}

\subsection{Comparisons to other time series clustering algorithms with the same setup}

\subsubsection{With cross-validation}

\begin{figure}[H]
    \begin{subfigure}{0.49\textwidth}
        \includegraphics[width=\textwidth]{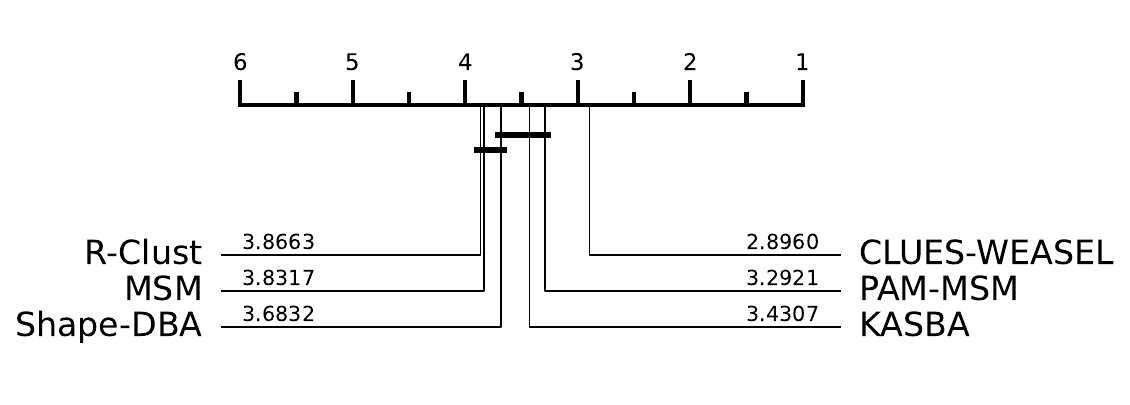}
        \caption{Clustering accuracy}
    \end{subfigure}%
    \hfill
    \begin{subfigure}{0.49\textwidth}
        \includegraphics[width=\textwidth]{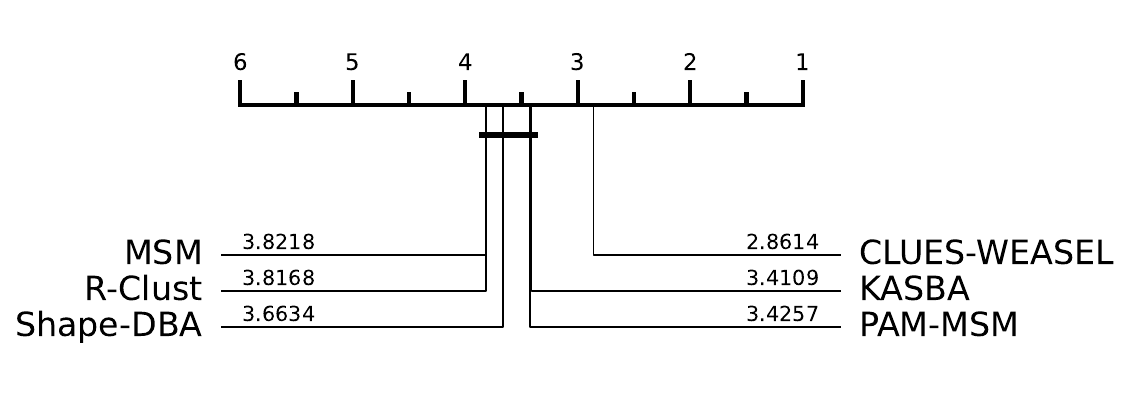}
        \caption{Adjusted Rand index}
    \end{subfigure}%

    \begin{subfigure}{0.49\textwidth}
        \includegraphics[width=\textwidth]{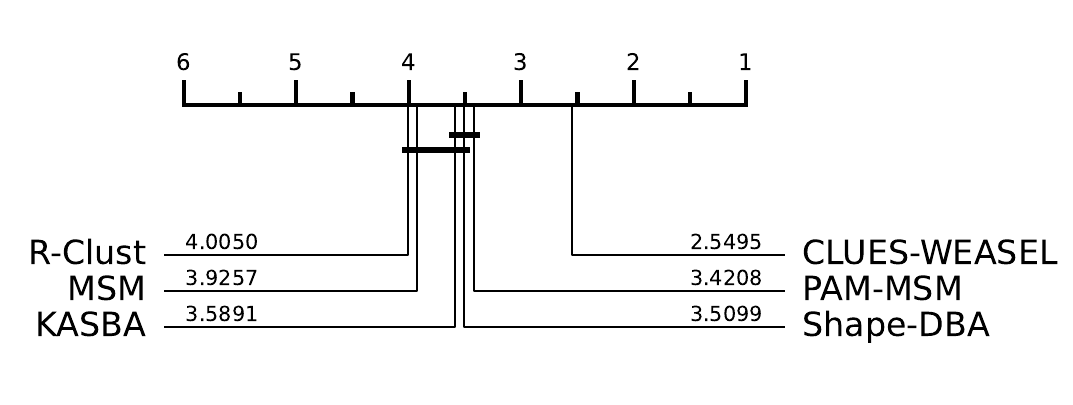}
        \caption{Adjusted mutual information}
    \end{subfigure}%
    \hfill
    \begin{subfigure}{0.49\textwidth}
        \includegraphics[width=\textwidth]{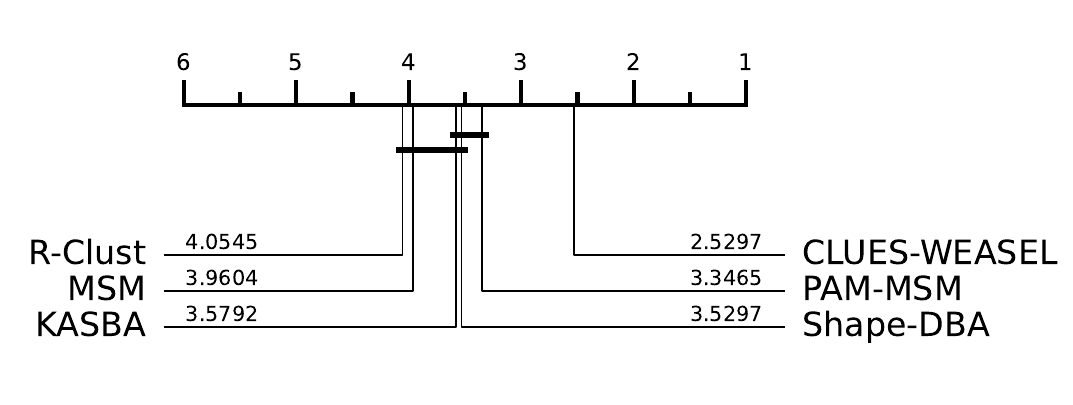}
        \caption{Normalized mutual information}
    \end{subfigure}
    \caption{
        CLUES-WEASEL against five state-of-the-art time series clustering algorithms with cross-validation.
        All the algorithms are trained on the default training sets and evaluated on the default test sets.
        The mean ranks are computed over 101 of the 112 data sets as the results are missing for 11 data sets for some algorithms because of a too high runtime.
    }
    \label{figa1}
\end{figure}

\subsubsection{Without cross-validation}

\begin{figure}[H]
    \begin{subfigure}{0.49\textwidth}
        \includegraphics[width=\textwidth]{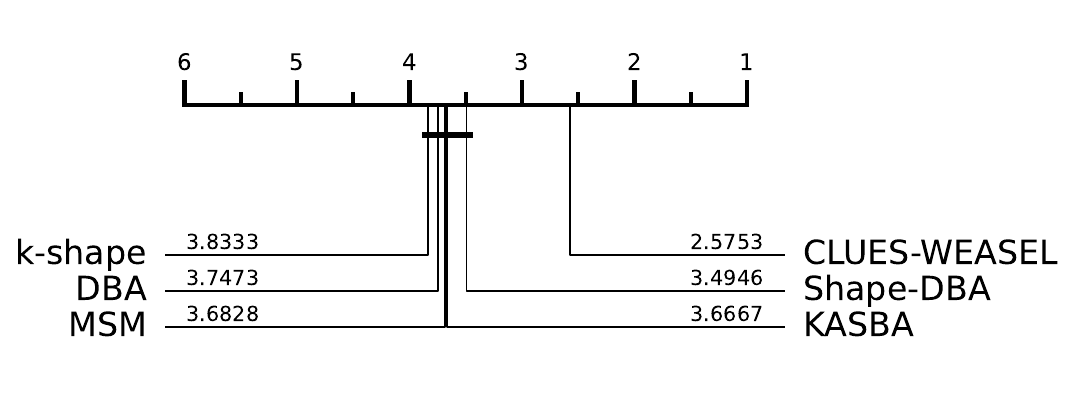}
        \caption{Clustering accuracy}
    \end{subfigure}%
    \hfill
    \begin{subfigure}{0.49\textwidth}
        \includegraphics[width=\textwidth]{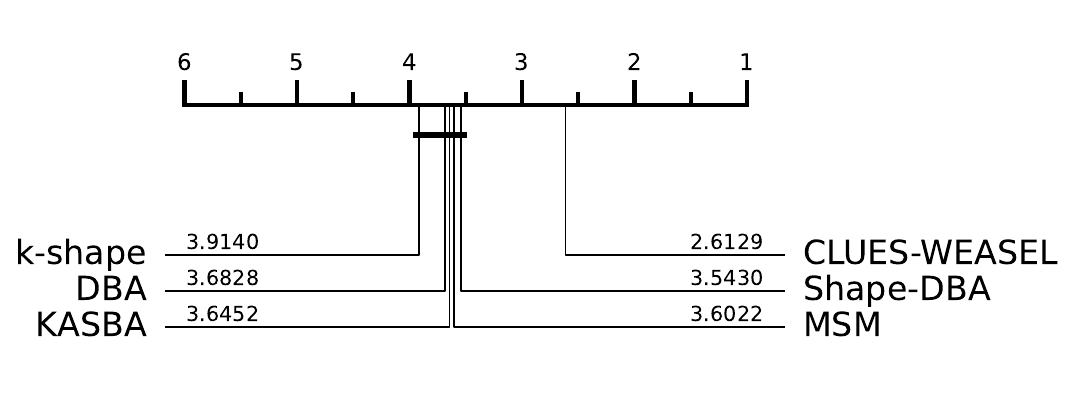}
        \caption{Adjusted Rand index}
    \end{subfigure}%

    \begin{subfigure}{0.49\textwidth}
        \includegraphics[width=\textwidth]{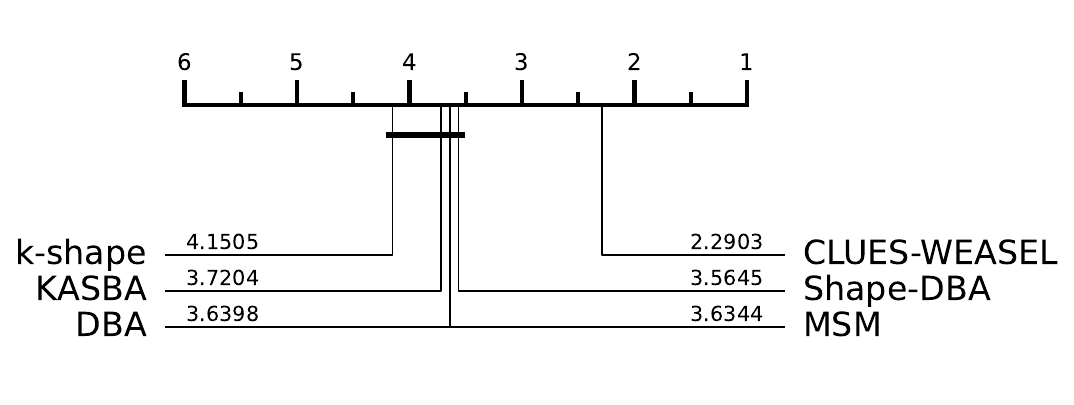}
        \caption{Adjusted mutual information}
    \end{subfigure}%
    \hfill
    \begin{subfigure}{0.49\textwidth}
        \includegraphics[width=\textwidth]{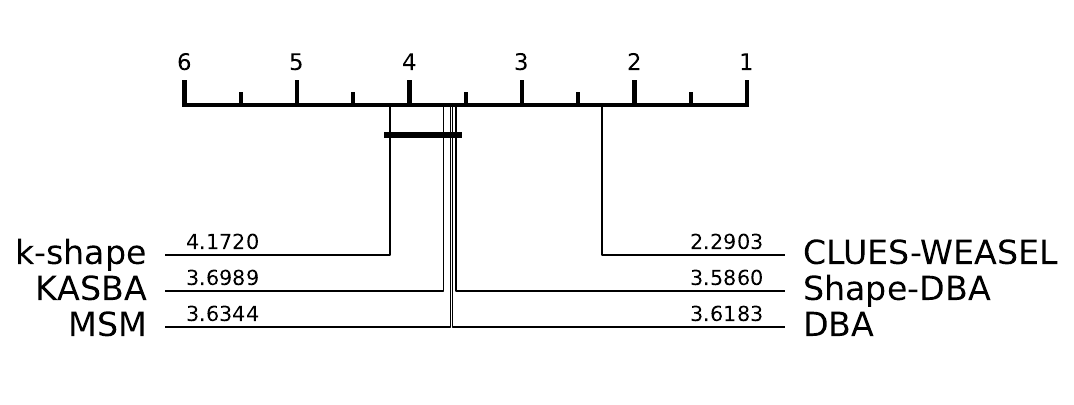}
        \caption{Normalized mutual information}
    \end{subfigure}
    \caption{
        CLUES-WEASEL against five state-of-the-art time series clustering algorithms without cross-validation.
        All the algorithms are trained and evaluated on the merged training and test sets.
        The mean ranks are computed over 93 of the 112 evaluation data sets as the results are missing for 19 data sets for some algorithms because of a too high runtime.
    }
    \label{figa2}
\end{figure}

\subsection{Comparisons to other time series clustering algorithms with different setups}

\subsubsection{Deep learning}

\begin{figure}[H]
    \begin{subfigure}{0.49\textwidth}
        \includegraphics[width=\textwidth]{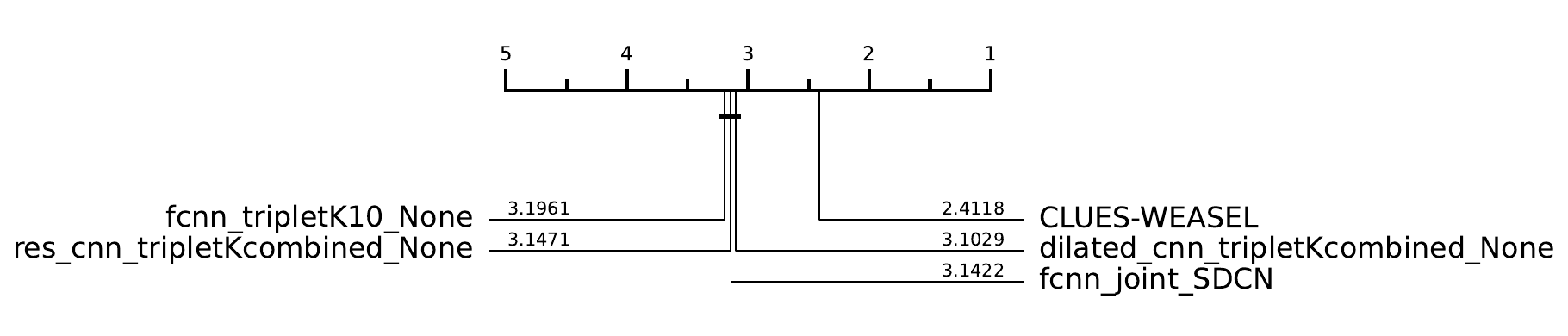}
        \caption{Clustering accuracy - With missing results}
    \end{subfigure}%
    \hfill
    \begin{subfigure}{0.49\textwidth}
        \includegraphics[width=\textwidth]{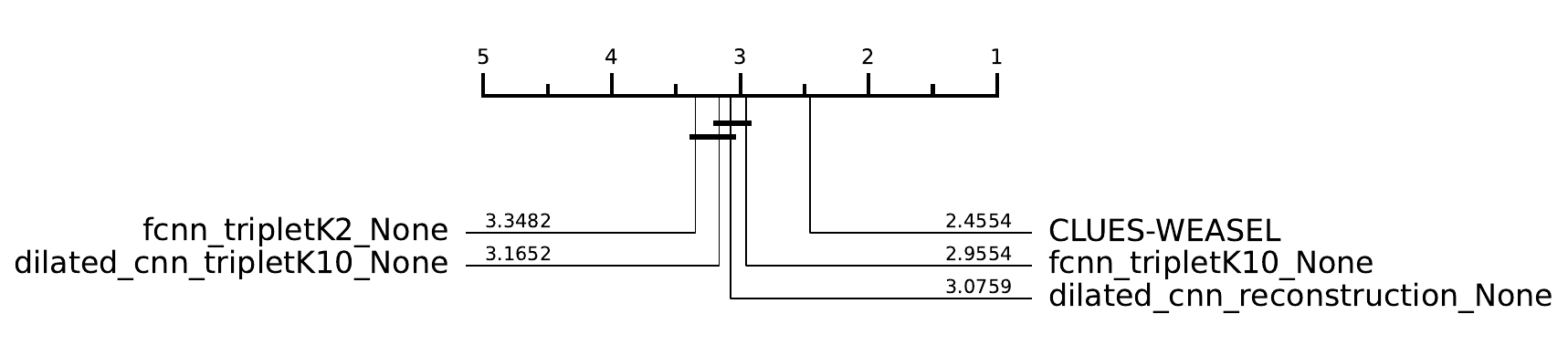}
        \caption{Clustering accuracy - Without missing results}
    \end{subfigure}%

    \begin{subfigure}{0.49\textwidth}
        \includegraphics[width=\textwidth]{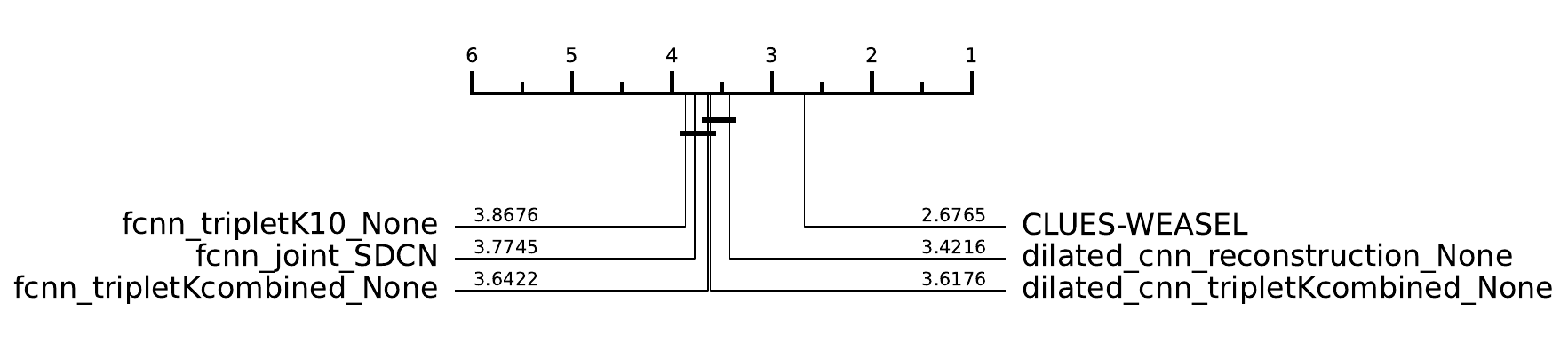}
        \caption{Normalized mutual information - With missing results}
    \end{subfigure}%
    \hfill
    \begin{subfigure}{0.49\textwidth}
        \includegraphics[width=\textwidth]{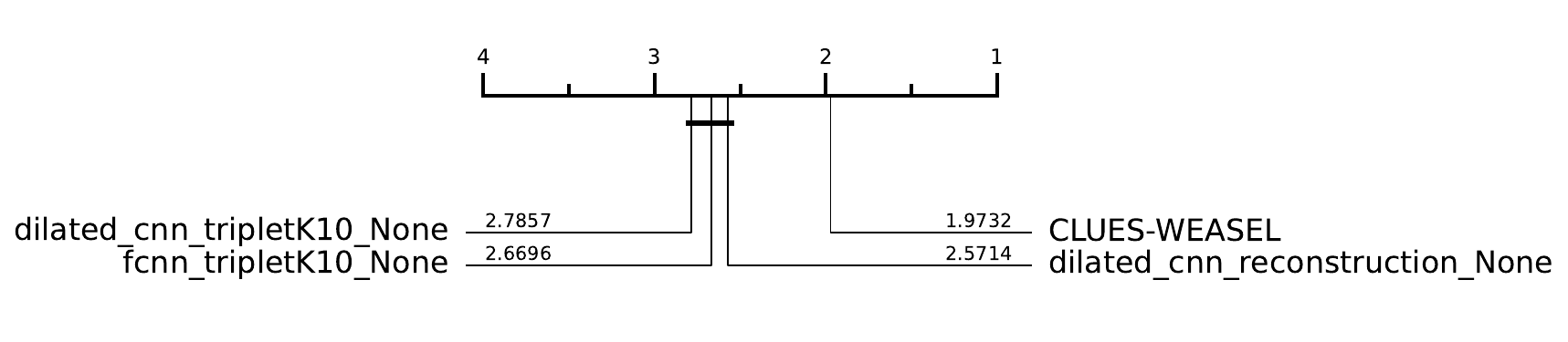}
        \caption{Normalized mutual information - Without missing results}
    \end{subfigure}
    \caption{
        CLUES-WEASEL against state-of-the-art deep learning time series clustering algorithms with cross-validation.
        For each configuration, the best deep learning algorithms have been selected as the union of the top 3 rankings in terms of mean scores and mean ranks.
        All the algorithms are trained on the default training sets and evaluated on the default test sets.
        The mean ranks are computed over all the available data sets.
    }
    \label{figa3}
\end{figure}

\subsection{Ablation experiments}

\subsubsection{Use of other unsupervised transformations}

\begin{figure}[H]
    \begin{subfigure}{0.49\textwidth}
        \includegraphics[width=\textwidth]{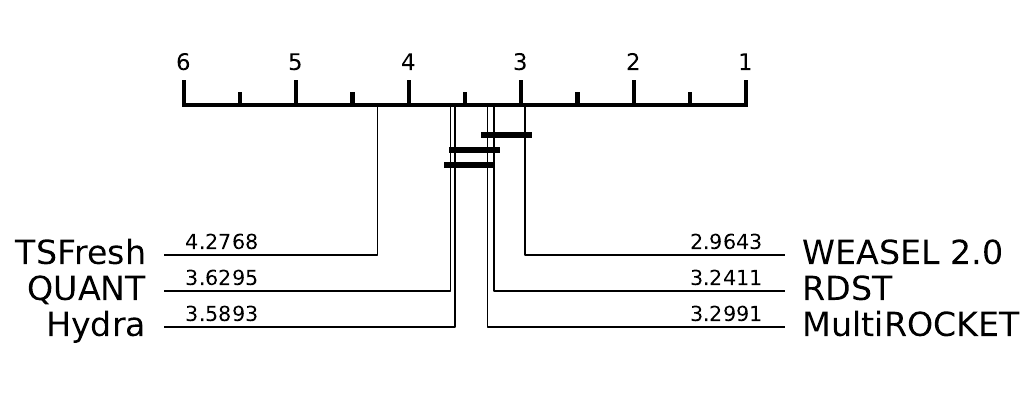}
        \caption{Clustering accuracy}
    \end{subfigure}%
    \hfill
    \begin{subfigure}{0.49\textwidth}
        \includegraphics[width=\textwidth]{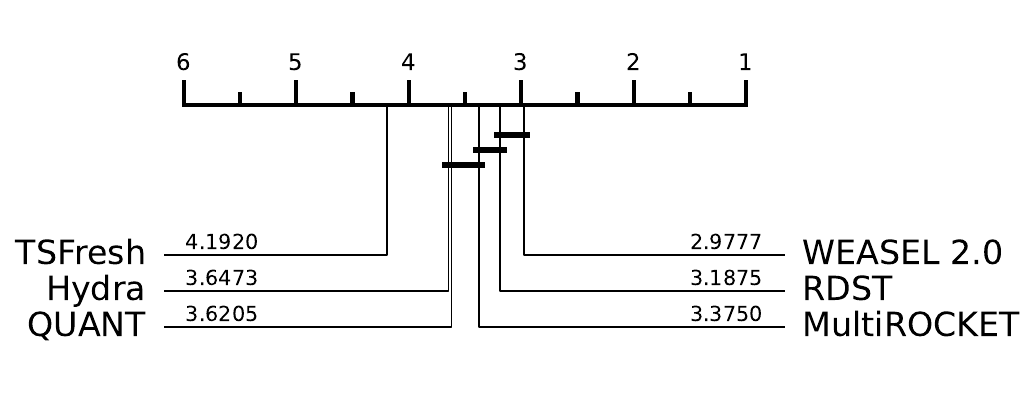}
        \caption{Adjusted Rand index}
    \end{subfigure}%

    \begin{subfigure}{0.49\textwidth}
        \includegraphics[width=\textwidth]{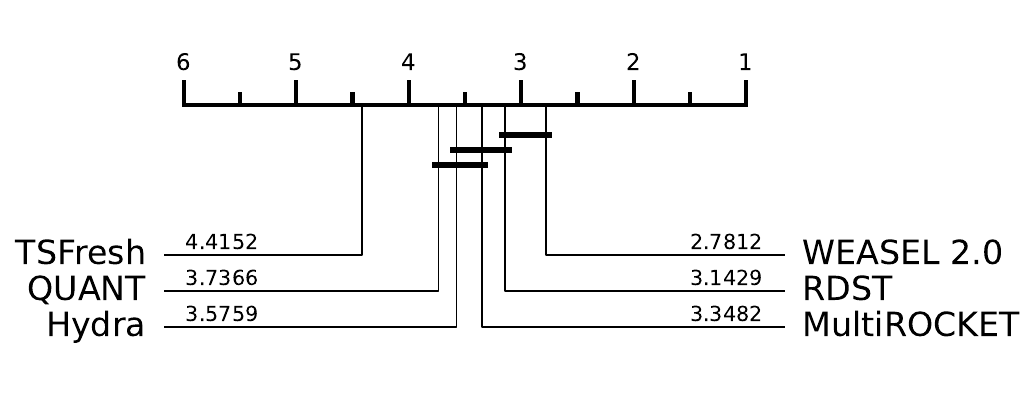}
        \caption{Adjusted mutual information}
    \end{subfigure}%
    \hfill
    \begin{subfigure}{0.49\textwidth}
        \includegraphics[width=\textwidth]{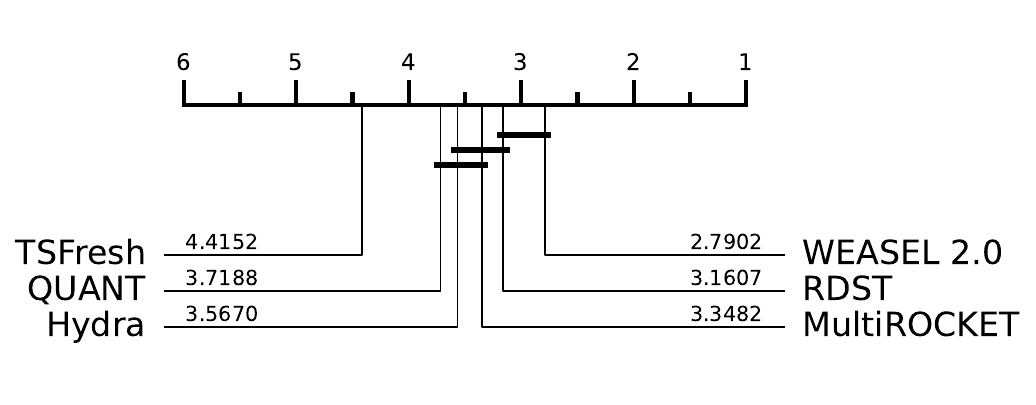}
        \caption{Normalized mutual information}
    \end{subfigure}
    \caption{
        WEASEL 2.0 against five transformation steps of state-of-the-art time series classification algorithms.
        All the algorithms are trained on the default training sets and evaluated on the default test sets.
        The mean ranks are computed over the 112 data sets (the 71 evaluation data sets and the 41 development data sets).
    }
    \label{figa4}
\end{figure}

\begin{figure}[H]
    \includegraphics[width=\textwidth]{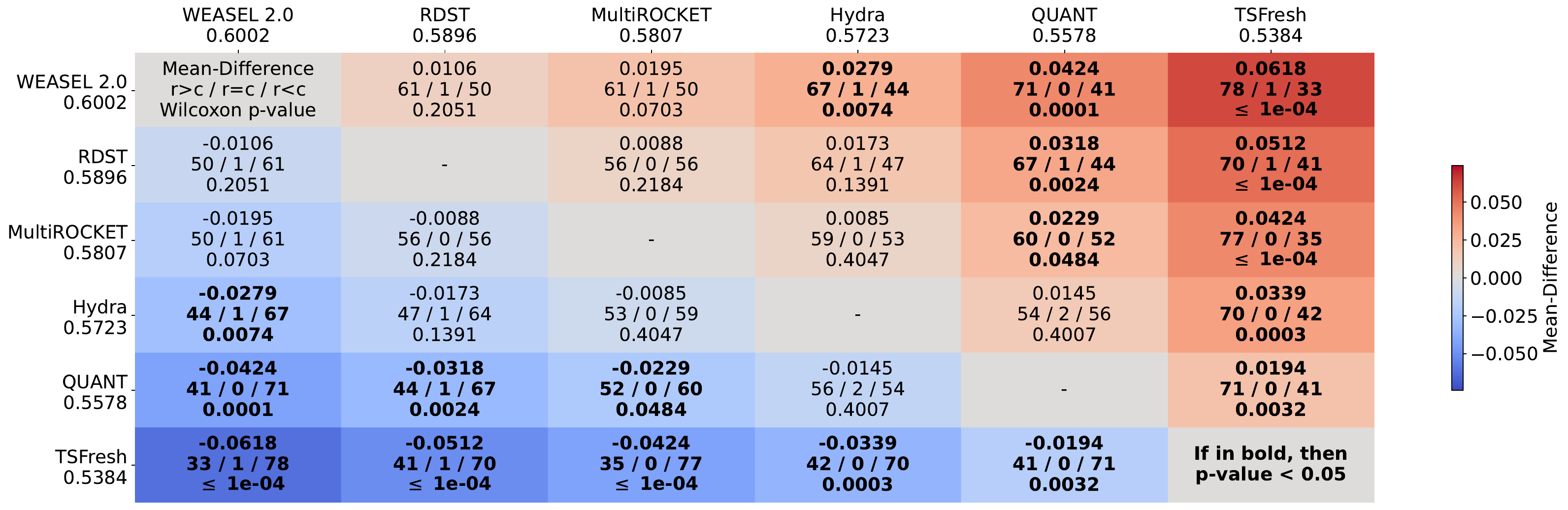}
    \caption{
        Multiple comparison matrix for 6 unsupervised transformation algorithms in terms of clustering accuracy on the 112 data sets (the 71 evaluation data sets and the 41 development data sets).
        All the algorithms are trained on the default training sets and evaluated on the default test sets.
        In each cell, the mean difference (top), the numbers of wins/ties/losses (middle), and the $p$-value (unadjusted for multiple comparisons) for a one-sided Wilcoxon sign rank test are highlighted.
        Bold values indicate significant differences (at the confidence level of 0.05).
    }
    \label{figa5}
\end{figure}

\begin{figure}[H]
    \begin{subfigure}{0.49\textwidth}
        \includegraphics[width=\textwidth]{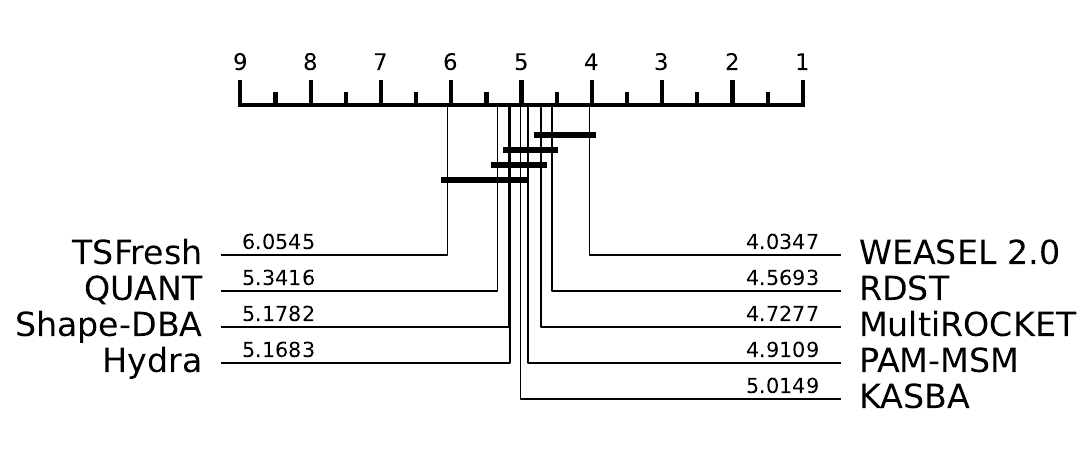}
        \caption{Clustering accuracy}
    \end{subfigure}%
    \hfill
    \begin{subfigure}{0.49\textwidth}
        \includegraphics[width=\textwidth]{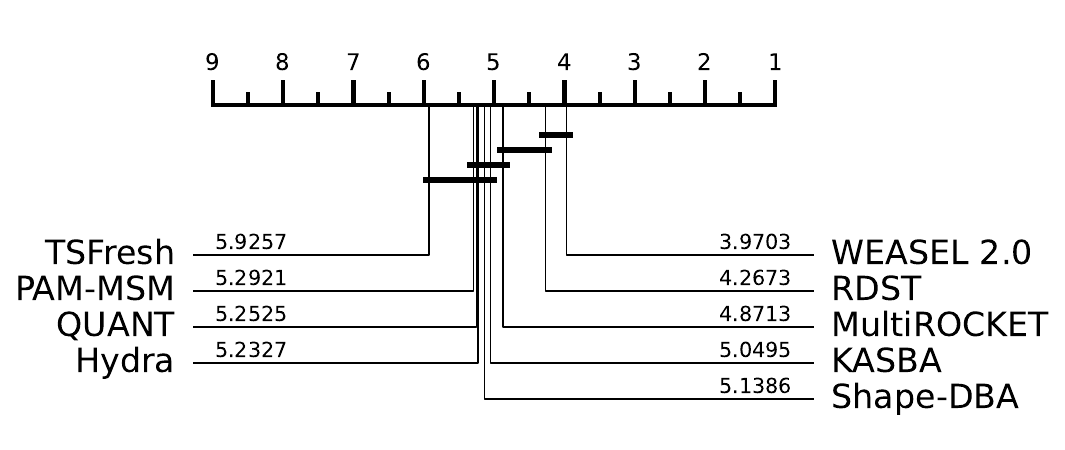}
        \caption{Adjusted Rand index}
    \end{subfigure}%

    \begin{subfigure}{0.49\textwidth}
        \includegraphics[width=\textwidth]{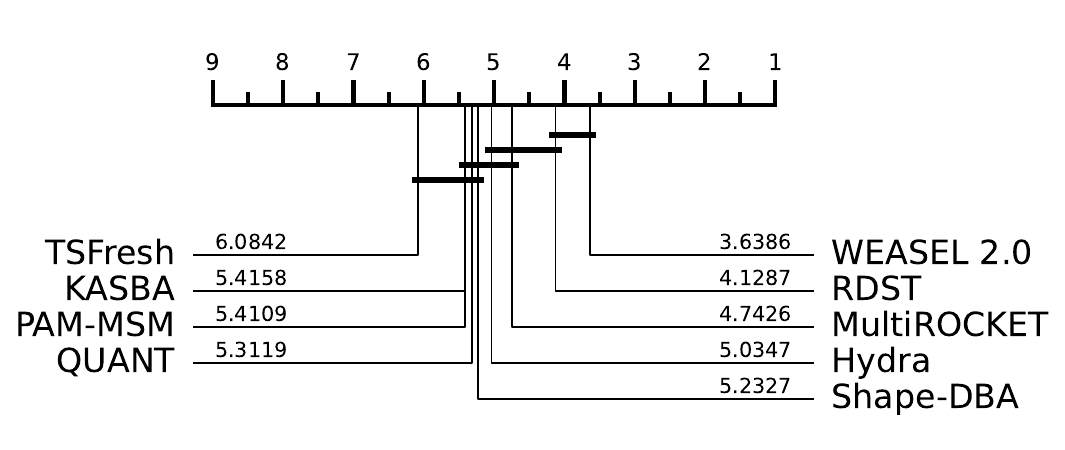}
        \caption{Adjusted mutual information}
    \end{subfigure}%
    \hfill
    \begin{subfigure}{0.49\textwidth}
        \includegraphics[width=\textwidth]{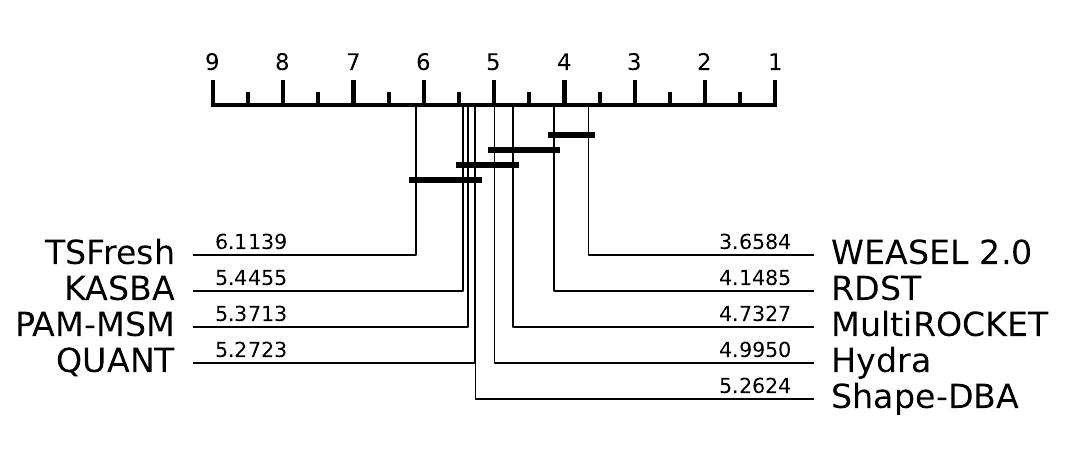}
        \caption{Normalized mutual information}
    \end{subfigure}
    \caption{
        WEASEL 2.0 and five transformation steps of state-of-the-art time series classification algorithms.
        The architectures of these time series clustering algorithms is the same as the one depicted in Figure~\ref{fig3}, except that the second step is replaced with the corresponding transformation and that the total variance explained ratio has been optimized for each transformation.
        These six algorithms are compared to three state-of-the-art time series clustering algorithms.
        All the algorithms are trained on the default training sets and evaluated on the default test sets.
        The mean ranks are computed over 101 of the 112 evaluation data sets as the results are missing for 11 data sets for some algorithms because of a too high runtime.
    }
    \label{figa6}
\end{figure}

\subsubsection{Impact of the maximum feature count}

\begin{figure}[H]
    \begin{subfigure}{0.49\textwidth}
        \includegraphics[width=\textwidth]{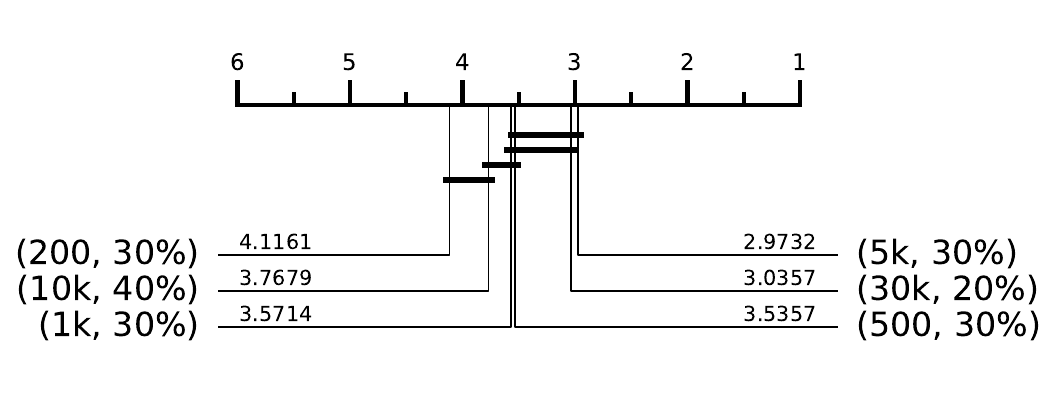}
        \caption{Clustering accuracy}
    \end{subfigure}%
    \hfill
    \begin{subfigure}{0.49\textwidth}
        \includegraphics[width=\textwidth]{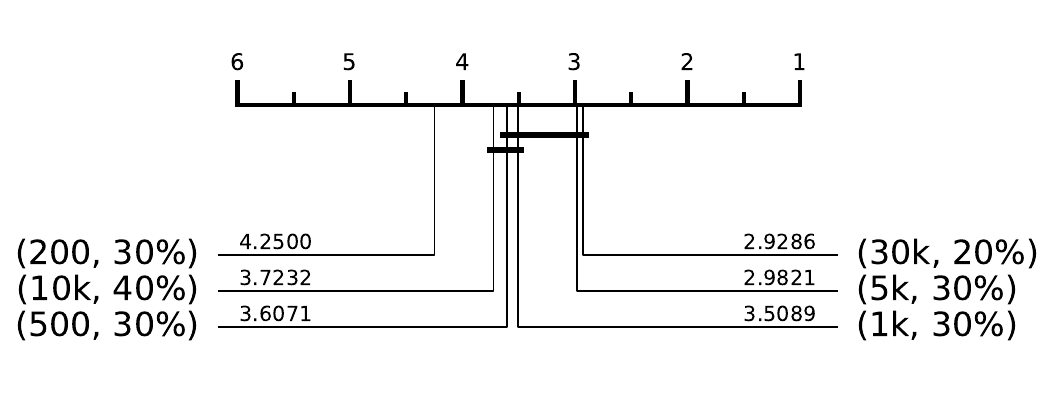}
        \caption{Adjusted Rand index}
    \end{subfigure}%

    \begin{subfigure}{0.49\textwidth}
        \includegraphics[width=\textwidth]{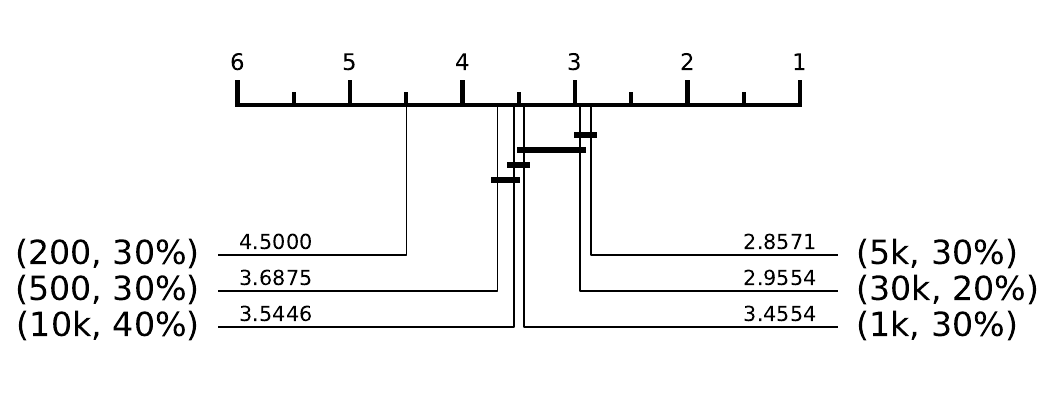}
        \caption{Adjusted mutual information}
    \end{subfigure}%
    \hfill
    \begin{subfigure}{0.49\textwidth}
        \includegraphics[width=\textwidth]{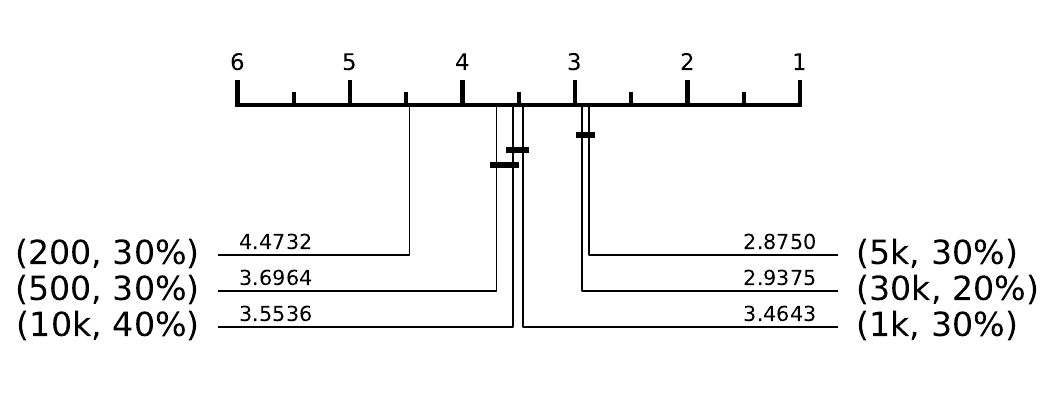}
        \caption{Normalized mutual information}
    \end{subfigure}
    \caption{
        CLUES-WEASEL with different maximum feature counts (first value inside the brackets) and total variance explained ratios (second value inside the brackets).
        All the versions are trained on the default training sets and evaluated on the default test sets.
        The mean ranks are computed over all the 112 data sets.
    }
    \label{figa7}
\end{figure}

\begin{figure}[H]
    \includegraphics[width=\textwidth]{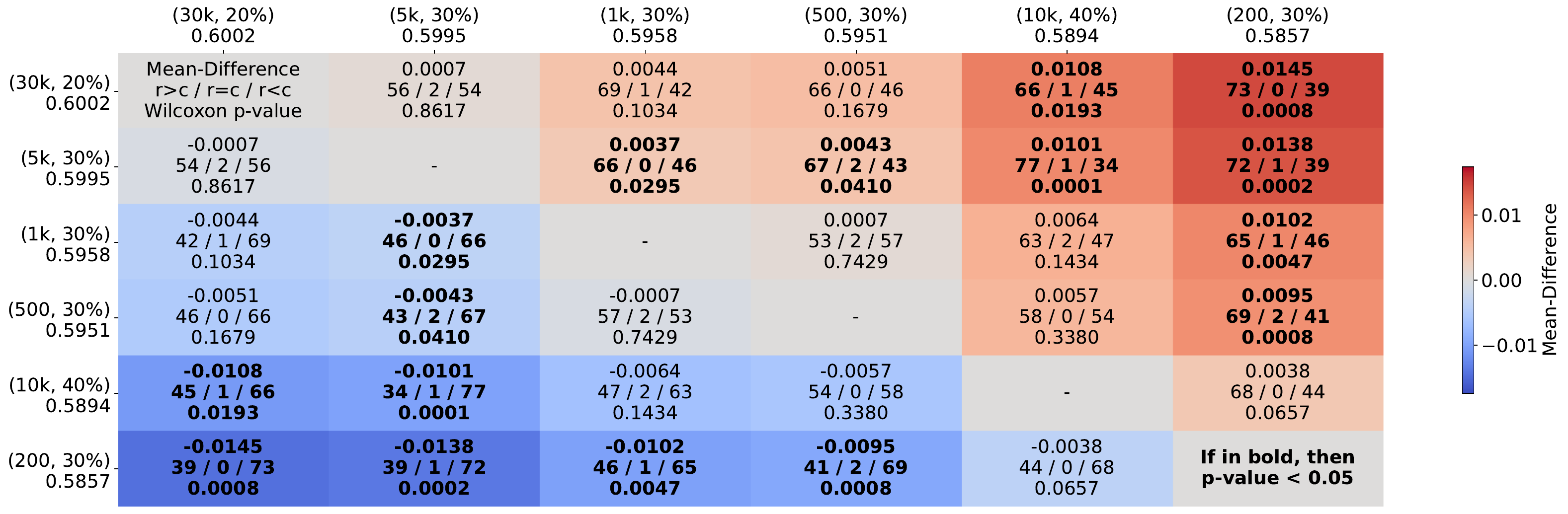}
    \caption{
        Multiple comparison matrix for CLUES-WEASEL with different maximum feature counts (first value inside the brackets) and total variance explained ratios (second value inside the brackets) in terms of clustering accuracy on all the 112 data sets.
        All the versions are trained on the default training sets and evaluated on the default test sets.
        In each cell, the mean difference (top), the numbers of wins/ties/losses (middle), and the $p$-value (unadjusted for multiple comparisons) for a one-sided Wilcoxon sign rank test are highlighted.
        Bold values indicate significant differences (at the confidence level of 0.05).
    }
    \label{figa8}
\end{figure}

\section{Exploratory analyses}

\begin{figure}[H]
    \includegraphics[width=\textwidth]{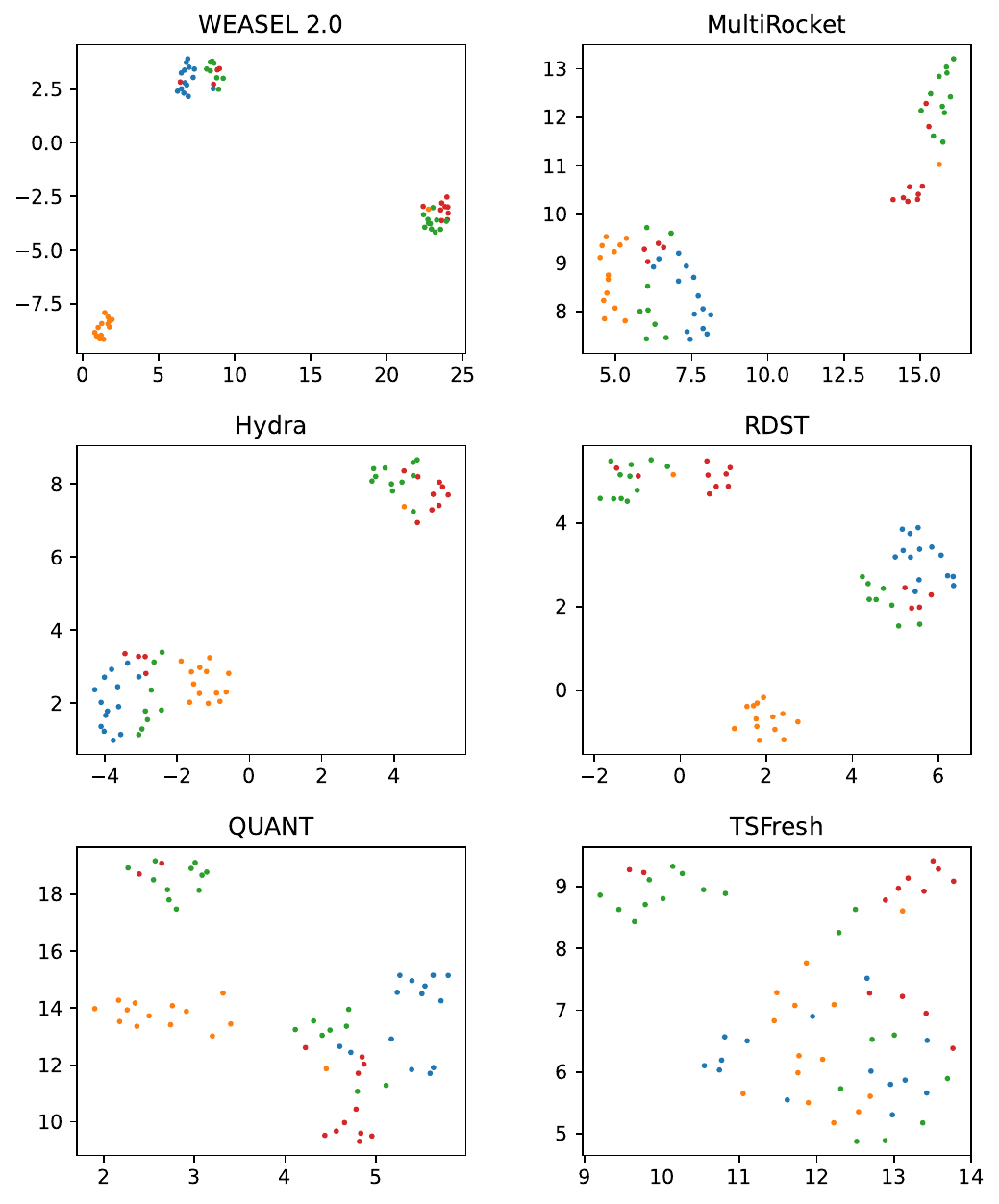}
    \caption{
        Visualization of the reduced extracted features by different algorithms on the Car data set.
        All the algorithms are trained on the default train set, and the figure shows the visualization of the reduced extracted features on the test set in a two-dimensional space using UMAP.
        This data set has four classes and the colors highlight the true classes.
        }
    \label{figb9}
\end{figure}

\begin{figure}[H]
    \includegraphics[width=\textwidth]{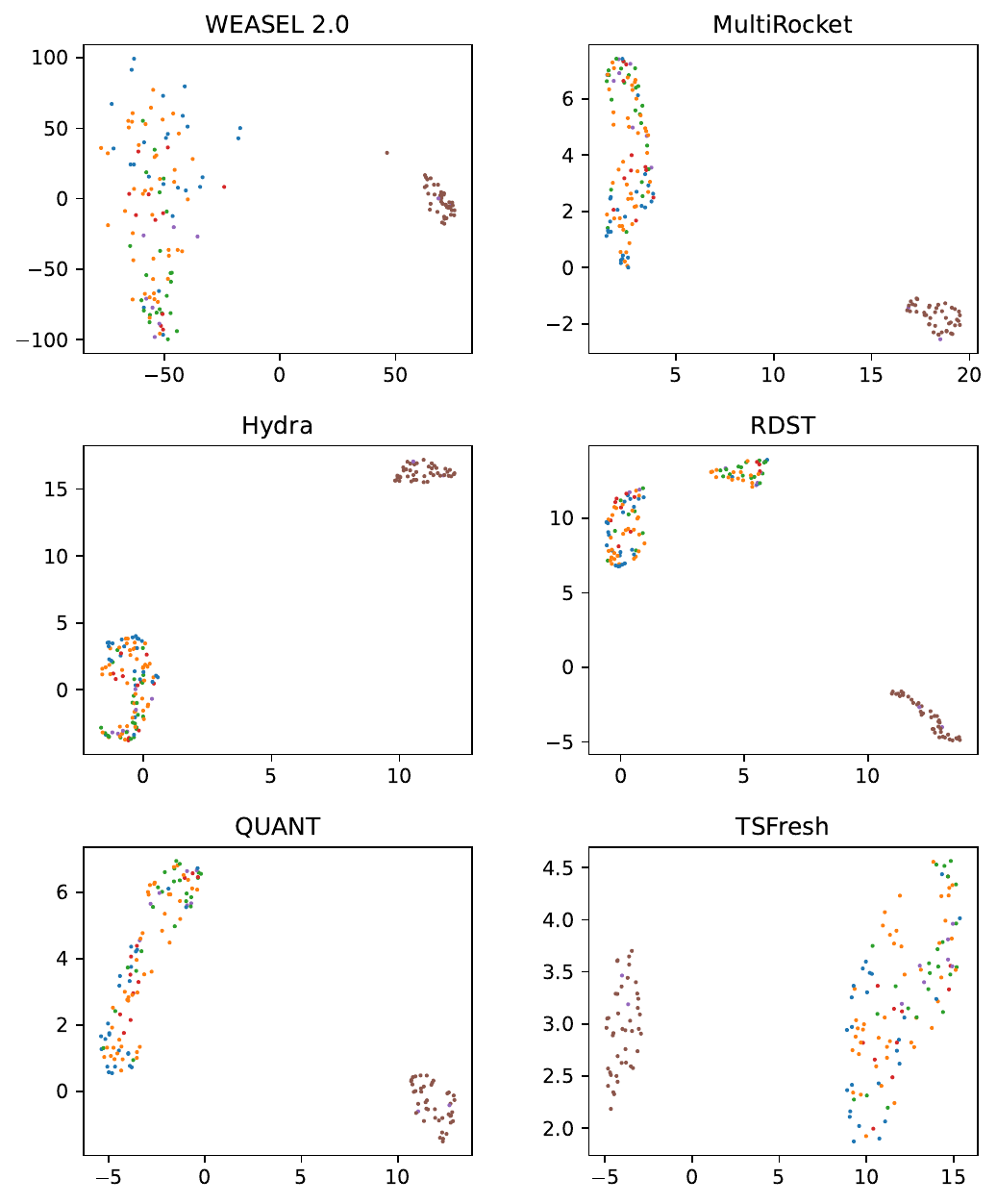}
    \caption{
        Visualization of the reduced extracted features by different algorithms on the MiddlePhalanxTW data set.
        All the algorithms are trained on the default train set, and the figure shows the visualization of the reduced extracted features on the test set in a two-dimensional space using UMAP.
        This data set has six classes and the colors highlight the true classes.
        }
    \label{figb10}
\end{figure}

\begin{figure}[H]
    \includegraphics[width=\textwidth]{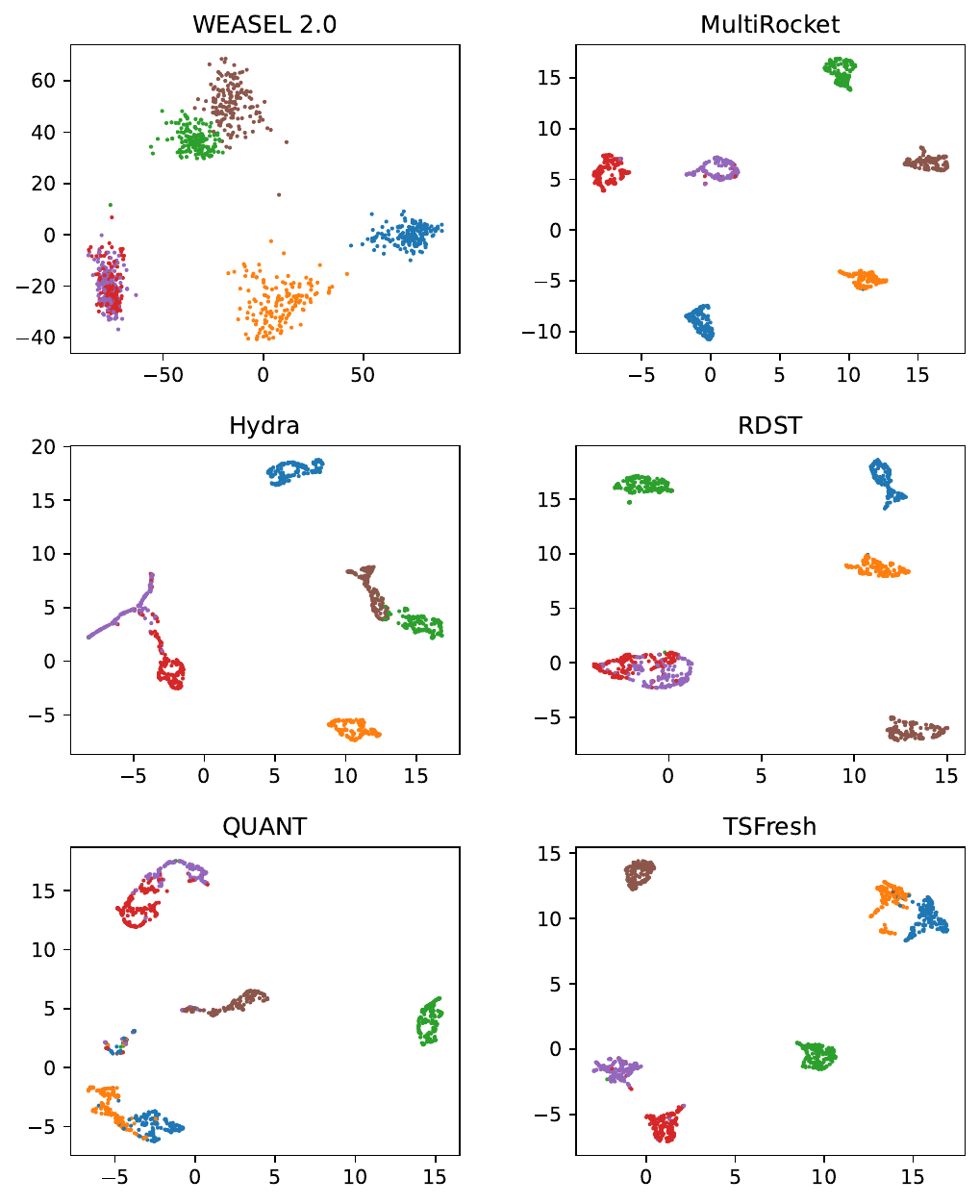}
    \caption{
        Visualization of the reduced extracted features by different algorithms on the Symbols data set.
        All the algorithms are trained on the default train set, and the figure shows the visualization of the reduced extracted features on the test set in a two-dimensional space using UMAP.
        This data set has six classes and the colors highlight the true classes.
        }
    \label{figb11}
\end{figure}




\end{appendices}


\bibliography{sn-bibliography}

\begin{thebibliography}{}
\renewcommand{\doi}[1]{\url{https://doi.org/#1}}
\bibcommenthead

\bibitem [\protect \citeauthoryear {%
Agrawal%
, Kumar%
, Pandey%
\BCBL {}\ \BBA {} Khan%
}{%
Agrawal%
\ \protect \BOthers {.}}{%
{\protect \APACyear {2012}}%
}]{%
agrawal2012application}
\APACinsertmetastar {%
agrawal2012application}%
\begin{APACrefauthors}%
Agrawal, A.%
, Kumar, V.%
, Pandey, A.%
\BCBL {} Khan, I.%
\end{APACrefauthors}%
\unskip\
\newblock
\APACrefYearMonthDay{2012}{}{}.
\newblock
{\BBOQ}\APACrefatitle {An application of time series analysis for weather forecasting} {An application of time series analysis for weather forecasting}.{\BBCQ}
\newblock
\APACjournalVolNumPages{International Journal of Engineering Research and Applications}{2}{2}{974--980,}
\newblock

\newblock

\PrintBackRefs{\CurrentBib}

\bibitem [\protect \citeauthoryear {%
Amei%
, Fu%
\BCBL {}\ \BBA {} Ho%
}{%
Amei%
\ \protect \BOthers {.}}{%
{\protect \APACyear {2012}}%
}]{%
amei2012time}
\APACinsertmetastar {%
amei2012time}%
\begin{APACrefauthors}%
Amei, A.%
, Fu, W.%
\BCBL {} Ho, C\BHBI H.%
\end{APACrefauthors}%
\unskip\
\newblock
\APACrefYearMonthDay{2012}{}{}.
\newblock
{\BBOQ}\APACrefatitle {Time series analysis for predicting the occurrences of large scale earthquakes} {Time series analysis for predicting the occurrences of large scale earthquakes}.{\BBCQ}
\newblock
\APACjournalVolNumPages{International Journal of Applied Science and Technology}{2}{7}{,}
\newblock

\newblock

\PrintBackRefs{\CurrentBib}

\bibitem [\protect \citeauthoryear {%
Arthur%
\ \BBA {} Vassilvitskii%
}{%
Arthur%
\ \BBA {} Vassilvitskii%
}{%
{\protect \APACyear {2007}}%
}]{%
arthurKmeansAdvantagesCareful2007a}
\APACinsertmetastar {%
arthurKmeansAdvantagesCareful2007a}%
\begin{APACrefauthors}%
Arthur, D.%
\BCBT {}\ \BBA {} Vassilvitskii, S.%
\end{APACrefauthors}%
\unskip\
\newblock
\APACrefYearMonthDay{2007}{{\APACmonth{01}}}{}.
\newblock
{\BBOQ}\APACrefatitle {K-Means++: The Advantages of Careful Seeding} {K-means++: The advantages of careful seeding}.{\BBCQ}
\newblock
 \APACrefbtitle {Proceedings of the Eighteenth Annual {{ACM-SIAM}} Symposium on {{Discrete}} Algorithms} {Proceedings of the eighteenth annual {{ACM-SIAM}} symposium on {{Discrete}} algorithms}\ (\BPGS\ 1027--1035).
\newblock
\APACaddressPublisher{USA}{{Society for Industrial and Applied Mathematics}}.
\PrintBackRefs{\CurrentBib}

\bibitem [\protect \citeauthoryear {%
Benavoli%
, Corani%
\BCBL {}\ \BBA {} Mangili%
}{%
Benavoli%
\ \protect \BOthers {.}}{%
{\protect \APACyear {2016}}%
}]{%
benavoliShouldWeReally2016}
\APACinsertmetastar {%
benavoliShouldWeReally2016}%
\begin{APACrefauthors}%
Benavoli, A.%
, Corani, G.%
\BCBL {} Mangili, F.%
\end{APACrefauthors}%
\unskip\
\newblock
\APACrefYearMonthDay{2016}{}{}.
\newblock
{\BBOQ}\APACrefatitle {Should {{We Really Use Post-Hoc Tests Based}} on {{Mean-Ranks}}?} {Should {{We Really Use Post-Hoc Tests Based}} on {{Mean-Ranks}}?}{\BBCQ}
\newblock
\APACjournalVolNumPages{Journal of Machine Learning Research}{17}{5}{1--10,}
\newblock

\newblock

\PrintBackRefs{\CurrentBib}

\bibitem [\protect \citeauthoryear {%
Bentley%
}{%
Bentley%
}{%
{\protect \APACyear {1975}}%
}]{%
bentleyMultidimensionalBinarySearch1975a}
\APACinsertmetastar {%
bentleyMultidimensionalBinarySearch1975a}%
\begin{APACrefauthors}%
Bentley, J.L.%
\end{APACrefauthors}%
\unskip\
\newblock
\APACrefYearMonthDay{1975}{{\APACmonth{09}}}{}.
\newblock
{\BBOQ}\APACrefatitle {Multidimensional Binary Search Trees Used for Associative Searching} {Multidimensional binary search trees used for associative searching}.{\BBCQ}
\newblock
\APACjournalVolNumPages{Commun. ACM}{18}{9}{509--517,}
\newblock
\begin{APACrefDOI} \doi{10.1145/361002.361007} \end{APACrefDOI}
\newblock

\newblock

\PrintBackRefs{\CurrentBib}

\bibitem [\protect \citeauthoryear {%
Berndt%
\ \BBA {} Clifford%
}{%
Berndt%
\ \BBA {} Clifford%
}{%
{\protect \APACyear {1994}}%
}]{%
berndtUsingDynamicTime1994}
\APACinsertmetastar {%
berndtUsingDynamicTime1994}%
\begin{APACrefauthors}%
Berndt, D.J.%
\BCBT {}\ \BBA {} Clifford, J.%
\end{APACrefauthors}%
\unskip\
\newblock
\APACrefYearMonthDay{1994}{{\APACmonth{07}}}{}.
\newblock
{\BBOQ}\APACrefatitle {Using Dynamic Time Warping to Find Patterns in Time Series} {Using dynamic time warping to find patterns in time series}.{\BBCQ}
\newblock
 \APACrefbtitle {Proceedings of the 3rd {{International Conference}} on {{Knowledge Discovery}} and {{Data Mining}}} {Proceedings of the 3rd {{International Conference}} on {{Knowledge Discovery}} and {{Data Mining}}}\ (\BPGS\ 359--370).
\newblock
\APACaddressPublisher{Seattle, WA}{AAAI Press}.
\PrintBackRefs{\CurrentBib}

\bibitem [\protect \citeauthoryear {%
Bostrom%
\ \BBA {} Bagnall%
}{%
Bostrom%
\ \BBA {} Bagnall%
}{%
{\protect \APACyear {2017}}%
}]{%
bostromBinaryShapeletTransform2017}
\APACinsertmetastar {%
bostromBinaryShapeletTransform2017}%
\begin{APACrefauthors}%
Bostrom, A.%
\BCBT {}\ \BBA {} Bagnall, A.%
\end{APACrefauthors}%
\unskip\
\newblock
\APACrefYearMonthDay{2017}{}{}.
\newblock
{\BBOQ}\APACrefatitle {Binary Shapelet Transform for Multiclass Time Series Classification (Extended Version)} {Binary shapelet transform for multiclass time series classification (extended version)}.{\BBCQ}
\newblock
 A.~Hameurlain, J.~K{\"u}ng, R.~Wagner, S.~Madria\BCBL {}\ \BBA {} T.~Hara\ (\BEDS), \APACrefbtitle {Transactions on {{Large-Scale Data-}} and {{Knowledge-Centered Systems XXXII}}} {Transactions on {{Large-Scale Data-}} and {{Knowledge-Centered Systems XXXII}}}\ (\BVOL\ 10420, \BPGS\ 24--46).
\newblock
\APACaddressPublisher{}{Springer}.
\PrintBackRefs{\CurrentBib}

\bibitem [\protect \citeauthoryear {%
Bostrom%
, Bagnall%
\BCBL {}\ \BBA {} Lines%
}{%
Bostrom%
\ \protect \BOthers {.}}{%
{\protect \APACyear {2016}}%
}]{%
bostromEvaluatingImprovementsShapelet2016a}
\APACinsertmetastar {%
bostromEvaluatingImprovementsShapelet2016a}%
\begin{APACrefauthors}%
Bostrom, A.%
, Bagnall, A.%
\BCBL {} Lines, J.%
\end{APACrefauthors}%
\unskip\
\newblock
\APACrefYearMonthDay{2016}{}{}.
\newblock
{\BBOQ}\APACrefatitle {Evaluating {{Improvements}} to the {{Shapelet Transform}}} {Evaluating {{Improvements}} to the {{Shapelet Transform}}}.{\BBCQ}
\newblock
 \APACrefbtitle {Second {{SIGKDD Workshop}} on {{Mining}} and {{Learning}} from {{Time Series}}.} {Second {{SIGKDD Workshop}} on {{Mining}} and {{Learning}} from {{Time Series}}.}
\PrintBackRefs{\CurrentBib}

\bibitem [\protect \citeauthoryear {%
Breiman%
}{%
Breiman%
}{%
{\protect \APACyear {2001}}%
}]{%
breimanRandomForests2001a}
\APACinsertmetastar {%
breimanRandomForests2001a}%
\begin{APACrefauthors}%
Breiman, L.%
\end{APACrefauthors}%
\unskip\
\newblock
\APACrefYearMonthDay{2001}{{\APACmonth{10}}}{}.
\newblock
{\BBOQ}\APACrefatitle {Random {{Forests}}} {Random {{Forests}}}.{\BBCQ}
\newblock
\APACjournalVolNumPages{Machine Learning}{45}{1}{5--32,}
\newblock
\begin{APACrefDOI} \doi{10.1023/A:1010933404324} \end{APACrefDOI}
\newblock

\newblock

\PrintBackRefs{\CurrentBib}

\bibitem [\protect \citeauthoryear {%
Choksi%
, Jain%
\BCBL {}\ \BBA {} Pindoriya%
}{%
Choksi%
\ \protect \BOthers {.}}{%
{\protect \APACyear {2020}}%
}]{%
choksiFeatureBasedClustering2020}
\APACinsertmetastar {%
choksiFeatureBasedClustering2020}%
\begin{APACrefauthors}%
Choksi, K.A.%
, Jain, S.%
\BCBL {} Pindoriya, N.M.%
\end{APACrefauthors}%
\unskip\
\newblock
\APACrefYearMonthDay{2020}{{\APACmonth{06}}}{}.
\newblock
{\BBOQ}\APACrefatitle {Feature Based Clustering Technique for Investigation of Domestic Load Profiles and Probabilistic Variation Assessment: {{Smart}} Meter Dataset} {Feature based clustering technique for investigation of domestic load profiles and probabilistic variation assessment: {{Smart}} meter dataset}.{\BBCQ}
\newblock
\APACjournalVolNumPages{Sustainable Energy, Grids and Networks}{22}{}{100346,}
\newblock
\begin{APACrefDOI} \doi{10.1016/j.segan.2020.100346} \end{APACrefDOI}
\newblock

\newblock

\PrintBackRefs{\CurrentBib}

\bibitem [\protect \citeauthoryear {%
Christ%
, Braun%
, Neuffer%
\BCBL {}\ \BBA {} {Kempa-Liehr}%
}{%
Christ%
\ \protect \BOthers {.}}{%
{\protect \APACyear {2018}}%
}]{%
christTimeSeriesFeatuRe2018a}
\APACinsertmetastar {%
christTimeSeriesFeatuRe2018a}%
\begin{APACrefauthors}%
Christ, M.%
, Braun, N.%
, Neuffer, J.%
\BCBL {} {Kempa-Liehr}, A.W.%
\end{APACrefauthors}%
\unskip\
\newblock
\APACrefYearMonthDay{2018}{{\APACmonth{09}}}{}.
\newblock
{\BBOQ}\APACrefatitle {Time {{Series FeatuRe Extraction}} on Basis of {{Scalable Hypothesis}} Tests (Tsfresh -- {{A Python}} Package)} {Time {{Series FeatuRe Extraction}} on basis of {{Scalable Hypothesis}} tests (tsfresh -- {{A Python}} package)}.{\BBCQ}
\newblock
\APACjournalVolNumPages{Neurocomputing}{307}{}{72--77,}
\newblock
\begin{APACrefDOI} \doi{10.1016/j.neucom.2018.03.067} \end{APACrefDOI}
\newblock

\newblock

\PrintBackRefs{\CurrentBib}

\bibitem [\protect \citeauthoryear {%
Cortes%
\ \BBA {} Vapnik%
}{%
Cortes%
\ \BBA {} Vapnik%
}{%
{\protect \APACyear {1995}}%
}]{%
cortesSupportVectorNetworks1995a}
\APACinsertmetastar {%
cortesSupportVectorNetworks1995a}%
\begin{APACrefauthors}%
Cortes, C.%
\BCBT {}\ \BBA {} Vapnik, V.%
\end{APACrefauthors}%
\unskip\
\newblock
\APACrefYearMonthDay{1995}{{\APACmonth{09}}}{}.
\newblock
{\BBOQ}\APACrefatitle {Support-{{Vector Networks}}} {Support-{{Vector Networks}}}.{\BBCQ}
\newblock
\APACjournalVolNumPages{Machine Learning}{20}{3}{273--297,}
\newblock
\begin{APACrefDOI} \doi{10.1023/A:1022627411411} \end{APACrefDOI}
\newblock

\newblock

\PrintBackRefs{\CurrentBib}

\bibitem [\protect \citeauthoryear {%
Cover%
\ \BBA {} Hart%
}{%
Cover%
\ \BBA {} Hart%
}{%
{\protect \APACyear {1967}}%
}]{%
coverNearestNeighborPattern1967}
\APACinsertmetastar {%
coverNearestNeighborPattern1967}%
\begin{APACrefauthors}%
Cover, T.%
\BCBT {}\ \BBA {} Hart, P.%
\end{APACrefauthors}%
\unskip\
\newblock
\APACrefYearMonthDay{1967}{{\APACmonth{01}}}{}.
\newblock
{\BBOQ}\APACrefatitle {Nearest Neighbor Pattern Classification} {Nearest neighbor pattern classification}.{\BBCQ}
\newblock
\APACjournalVolNumPages{IEEE Transactions on Information Theory}{13}{1}{21--27,}
\newblock
\begin{APACrefDOI} \doi{10.1109/TIT.1967.1053964} \end{APACrefDOI}
\newblock

\newblock

\PrintBackRefs{\CurrentBib}

\bibitem [\protect \citeauthoryear {%
Cuturi%
\ \BBA {} Blondel%
}{%
Cuturi%
\ \BBA {} Blondel%
}{%
{\protect \APACyear {2017}}%
}]{%
cuturiSoftDTWDifferentiableLoss2017a}
\APACinsertmetastar {%
cuturiSoftDTWDifferentiableLoss2017a}%
\begin{APACrefauthors}%
Cuturi, M.%
\BCBT {}\ \BBA {} Blondel, M.%
\end{APACrefauthors}%
\unskip\
\newblock
\APACrefYearMonthDay{2017}{{\APACmonth{08}}}{}.
\newblock
{\BBOQ}\APACrefatitle {Soft-{{DTW}}: A Differentiable Loss Function for Time-Series} {Soft-{{DTW}}: A differentiable loss function for time-series}.{\BBCQ}
\newblock
 \APACrefbtitle {Proceedings of the 34th {{International Conference}} on {{Machine Learning}} - {{Volume}} 70} {Proceedings of the 34th {{International Conference}} on {{Machine Learning}} - {{Volume}} 70}\ (\BPGS\ 894--903).
\newblock
\APACaddressPublisher{Sydney, NSW, Australia}{JMLR.org}.
\PrintBackRefs{\CurrentBib}

\bibitem [\protect \citeauthoryear {%
Dau%
\ \protect \BOthers {.}}{%
Dau%
\ \protect \BOthers {.}}{%
{\protect \APACyear {2019}}%
}]{%
dauUCRTimeSeries2019e}
\APACinsertmetastar {%
dauUCRTimeSeries2019e}%
\begin{APACrefauthors}%
Dau, H.A.%
, Bagnall, A.%
, Kamgar, K.%
, Yeh, C\BHBI C.M.%
, Zhu, Y.%
, Gharghabi, S.%
\BDBL {}Keogh, E.%
\end{APACrefauthors}%
\unskip\
\newblock
\APACrefYearMonthDay{2019}{{\APACmonth{11}}}{}.
\newblock
{\BBOQ}\APACrefatitle {The {{UCR}} Time Series Archive} {The {{UCR}} time series archive}.{\BBCQ}
\newblock
\APACjournalVolNumPages{IEEE/CAA Journal of Automatica Sinica}{6}{6}{1293--1305,}
\newblock
\begin{APACrefDOI} \doi{10.1109/JAS.2019.1911747} \end{APACrefDOI}
\newblock

\newblock

\PrintBackRefs{\CurrentBib}

\bibitem [\protect \citeauthoryear {%
Dempster%
, Petitjean%
\BCBL {}\ \BBA {} Webb%
}{%
Dempster%
\ \protect \BOthers {.}}{%
{\protect \APACyear {2020}}%
}]{%
dempsterROCKETExceptionallyFast2020}
\APACinsertmetastar {%
dempsterROCKETExceptionallyFast2020}%
\begin{APACrefauthors}%
Dempster, A.%
, Petitjean, F.%
\BCBL {} Webb, G.I.%
\end{APACrefauthors}%
\unskip\
\newblock
\APACrefYearMonthDay{2020}{{\APACmonth{09}}}{}.
\newblock
{\BBOQ}\APACrefatitle {{{ROCKET}}: Exceptionally Fast and Accurate Time Series Classification Using Random Convolutional Kernels} {{{ROCKET}}: Exceptionally fast and accurate time series classification using random convolutional kernels}.{\BBCQ}
\newblock
\APACjournalVolNumPages{Data Mining and Knowledge Discovery}{34}{5}{1454--1495,}
\newblock
\begin{APACrefDOI} \doi{10.1007/s10618-020-00701-z} \end{APACrefDOI}
\newblock

\newblock

\PrintBackRefs{\CurrentBib}

\bibitem [\protect \citeauthoryear {%
Dempster%
, Schmidt%
\BCBL {}\ \BBA {} Webb%
}{%
Dempster%
\ \protect \BOthers {.}}{%
{\protect \APACyear {2021}}%
}]{%
dempsterMiniRocketVeryFast2021a}
\APACinsertmetastar {%
dempsterMiniRocketVeryFast2021a}%
\begin{APACrefauthors}%
Dempster, A.%
, Schmidt, D.F.%
\BCBL {} Webb, G.I.%
\end{APACrefauthors}%
\unskip\
\newblock
\APACrefYearMonthDay{2021}{{\APACmonth{08}}}{}.
\newblock
{\BBOQ}\APACrefatitle {{{MiniRocket}}: {{A Very Fast}} ({{Almost}}) {{Deterministic Transform}} for {{Time Series Classification}}} {{{MiniRocket}}: {{A Very Fast}} ({{Almost}}) {{Deterministic Transform}} for {{Time Series Classification}}}.{\BBCQ}
\newblock
 \APACrefbtitle {Proceedings of the 27th {{ACM SIGKDD Conference}} on {{Knowledge Discovery}} \& {{Data Mining}}} {Proceedings of the 27th {{ACM SIGKDD Conference}} on {{Knowledge Discovery}} \& {{Data Mining}}}\ (\BPGS\ 248--257).
\newblock
\APACaddressPublisher{New York, NY, USA}{Association for Computing Machinery}.
\PrintBackRefs{\CurrentBib}

\bibitem [\protect \citeauthoryear {%
Dempster%
, Schmidt%
\BCBL {}\ \BBA {} Webb%
}{%
Dempster%
\ \protect \BOthers {.}}{%
{\protect \APACyear {2023}}%
}]{%
dempsterHydraCompetingConvolutional2023}
\APACinsertmetastar {%
dempsterHydraCompetingConvolutional2023}%
\begin{APACrefauthors}%
Dempster, A.%
, Schmidt, D.F.%
\BCBL {} Webb, G.I.%
\end{APACrefauthors}%
\unskip\
\newblock
\APACrefYearMonthDay{2023}{{\APACmonth{09}}}{}.
\newblock
{\BBOQ}\APACrefatitle {Hydra: Competing Convolutional Kernels for Fast and Accurate Time Series Classification} {Hydra: Competing convolutional kernels for fast and accurate time series classification}.{\BBCQ}
\newblock
\APACjournalVolNumPages{Data Mining and Knowledge Discovery}{37}{5}{1779--1805,}
\newblock
\begin{APACrefDOI} \doi{10.1007/s10618-023-00939-3} \end{APACrefDOI}
\newblock

\newblock

\PrintBackRefs{\CurrentBib}

\bibitem [\protect \citeauthoryear {%
Dempster%
, Schmidt%
\BCBL {}\ \BBA {} Webb%
}{%
Dempster%
\ \protect \BOthers {.}}{%
{\protect \APACyear {2024}}%
}]{%
dempsterQuantMinimalistInterval2024}
\APACinsertmetastar {%
dempsterQuantMinimalistInterval2024}%
\begin{APACrefauthors}%
Dempster, A.%
, Schmidt, D.F.%
\BCBL {} Webb, G.I.%
\end{APACrefauthors}%
\unskip\
\newblock
\APACrefYearMonthDay{2024}{{\APACmonth{07}}}{}.
\newblock
{\BBOQ}\APACrefatitle {Quant: A Minimalist Interval Method for Time Series Classification} {Quant: A minimalist interval method for time series classification}.{\BBCQ}
\newblock
\APACjournalVolNumPages{Data Mining and Knowledge Discovery}{38}{4}{2377--2402,}
\newblock
\begin{APACrefDOI} \doi{10.1007/s10618-024-01036-9} \end{APACrefDOI}
\newblock

\newblock

\PrintBackRefs{\CurrentBib}

\bibitem [\protect \citeauthoryear {%
Dem{\v s}ar%
}{%
Dem{\v s}ar%
}{%
{\protect \APACyear {2006}}%
}]{%
demsarStatisticalComparisonsClassifiers2006a}
\APACinsertmetastar {%
demsarStatisticalComparisonsClassifiers2006a}%
\begin{APACrefauthors}%
Dem{\v s}ar, J.%
\end{APACrefauthors}%
\unskip\
\newblock
\APACrefYearMonthDay{2006}{}{}.
\newblock
{\BBOQ}\APACrefatitle {Statistical {{Comparisons}} of {{Classifiers}} over {{Multiple Data Sets}}} {Statistical {{Comparisons}} of {{Classifiers}} over {{Multiple Data Sets}}}.{\BBCQ}
\newblock
\APACjournalVolNumPages{Journal of Machine Learning Research}{7}{1}{1--30,}
\newblock

\newblock

\PrintBackRefs{\CurrentBib}

\bibitem [\protect \citeauthoryear {%
Deng%
, Runger%
, Tuv%
\BCBL {}\ \BBA {} Vladimir%
}{%
Deng%
\ \protect \BOthers {.}}{%
{\protect \APACyear {2013}}%
}]{%
dengTimeSeriesForest2013}
\APACinsertmetastar {%
dengTimeSeriesForest2013}%
\begin{APACrefauthors}%
Deng, H.%
, Runger, G.%
, Tuv, E.%
\BCBL {} Vladimir, M.%
\end{APACrefauthors}%
\unskip\
\newblock
\APACrefYearMonthDay{2013}{{\APACmonth{08}}}{}.
\newblock
{\BBOQ}\APACrefatitle {A Time Series Forest for Classification and Feature Extraction} {A time series forest for classification and feature extraction}.{\BBCQ}
\newblock
\APACjournalVolNumPages{Information Sciences}{239}{}{142--153,}
\newblock
\begin{APACrefDOI} \doi{10.1016/j.ins.2013.02.030} \end{APACrefDOI}
\newblock

\newblock

\PrintBackRefs{\CurrentBib}

\bibitem [\protect \citeauthoryear {%
Ester%
, Kriegel%
, Sander%
\BCBL {}\ \BBA {} Xu%
}{%
Ester%
\ \protect \BOthers {.}}{%
{\protect \APACyear {1996}}%
}]{%
esterDensitybasedAlgorithmDiscovering1996}
\APACinsertmetastar {%
esterDensitybasedAlgorithmDiscovering1996}%
\begin{APACrefauthors}%
Ester, M.%
, Kriegel, H\BHBI P.%
, Sander, J.%
\BCBL {} Xu, X.%
\end{APACrefauthors}%
\unskip\
\newblock
\APACrefYearMonthDay{1996}{{\APACmonth{08}}}{}.
\newblock
{\BBOQ}\APACrefatitle {A Density-Based Algorithm for Discovering Clusters in Large Spatial Databases with Noise} {A density-based algorithm for discovering clusters in large spatial databases with noise}.{\BBCQ}
\newblock
 \APACrefbtitle {Proceedings of the {{Second International Conference}} on {{Knowledge Discovery}} and {{Data Mining}}} {Proceedings of the {{Second International Conference}} on {{Knowledge Discovery}} and {{Data Mining}}}\ (\BPGS\ 226--231).
\newblock
\APACaddressPublisher{Portland, Oregon}{AAAI Press}.
\PrintBackRefs{\CurrentBib}

\bibitem [\protect \citeauthoryear {%
Fern%
\ \BBA {} Brodley%
}{%
Fern%
\ \BBA {} Brodley%
}{%
{\protect \APACyear {2004}}%
}]{%
fernSolvingClusterEnsemble2004}
\APACinsertmetastar {%
fernSolvingClusterEnsemble2004}%
\begin{APACrefauthors}%
Fern, X.Z.%
\BCBT {}\ \BBA {} Brodley, C.E.%
\end{APACrefauthors}%
\unskip\
\newblock
\APACrefYearMonthDay{2004}{{\APACmonth{07}}}{}.
\newblock
{\BBOQ}\APACrefatitle {Solving Cluster Ensemble Problems by Bipartite Graph Partitioning} {Solving cluster ensemble problems by bipartite graph partitioning}.{\BBCQ}
\newblock
 \APACrefbtitle {Proceedings of the Twenty-First International Conference on {{Machine}} Learning} {Proceedings of the twenty-first international conference on {{Machine}} learning}\ (\BPG~36).
\newblock
\APACaddressPublisher{New York, NY, USA}{Association for Computing Machinery}.
\PrintBackRefs{\CurrentBib}

\bibitem [\protect \citeauthoryear {%
Fix%
\ \BBA {} Hodges%
}{%
Fix%
\ \BBA {} Hodges%
}{%
{\protect \APACyear {1989}}%
}]{%
fixDiscriminatoryAnalysisNonparametric1989}
\APACinsertmetastar {%
fixDiscriminatoryAnalysisNonparametric1989}%
\begin{APACrefauthors}%
Fix, E.%
\BCBT {}\ \BBA {} Hodges, J.L.%
\end{APACrefauthors}%
\unskip\
\newblock
\APACrefYearMonthDay{1989}{}{}.
\newblock
{\BBOQ}\APACrefatitle {Discriminatory {{Analysis}}. {{Nonparametric Discrimination}}: {{Consistency Properties}}} {Discriminatory {{Analysis}}. {{Nonparametric Discrimination}}: {{Consistency Properties}}}.{\BBCQ}
\newblock
\APACjournalVolNumPages{International Statistical Review / Revue Internationale de Statistique}{57}{3}{238--247,}
\newblock
\begin{APACrefDOI} \doi{10.2307/1403797} \end{APACrefDOI}
\newblock
{\href{https://arxiv.org/abs/1403797}{{1403797}}}
\newblock

\PrintBackRefs{\CurrentBib}

\bibitem [\protect \citeauthoryear {%
Forgy%
}{%
Forgy%
}{%
{\protect \APACyear {1965}}%
}]{%
forgy1965cluster}
\APACinsertmetastar {%
forgy1965cluster}%
\begin{APACrefauthors}%
Forgy, E.W.%
\end{APACrefauthors}%
\unskip\
\newblock
\APACrefYearMonthDay{1965}{}{}.
\newblock
{\BBOQ}\APACrefatitle {Cluster analysis of multivariate data: efficiency versus interpretability of classifications} {Cluster analysis of multivariate data: efficiency versus interpretability of classifications}.{\BBCQ}
\newblock
\APACjournalVolNumPages{biometrics}{21}{}{768--769,}
\newblock

\newblock

\PrintBackRefs{\CurrentBib}

\bibitem [\protect \citeauthoryear {%
Freund%
\ \BBA {} Schapire%
}{%
Freund%
\ \BBA {} Schapire%
}{%
{\protect \APACyear {1996}}%
}]{%
freundExperimentsNewBoosting1996}
\APACinsertmetastar {%
freundExperimentsNewBoosting1996}%
\begin{APACrefauthors}%
Freund, Y.%
\BCBT {}\ \BBA {} Schapire, R.E.%
\end{APACrefauthors}%
\unskip\
\newblock
\APACrefYearMonthDay{1996}{{\APACmonth{07}}}{}.
\newblock
{\BBOQ}\APACrefatitle {Experiments with a New Boosting Algorithm} {Experiments with a new boosting algorithm}.{\BBCQ}
\newblock
 \APACrefbtitle {Proceedings of the {{Thirteenth International Conference}} on {{International Conference}} on {{Machine Learning}}} {Proceedings of the {{Thirteenth International Conference}} on {{International Conference}} on {{Machine Learning}}}\ (\BPGS\ 148--156).
\newblock
\APACaddressPublisher{San Francisco, CA, USA}{Morgan Kaufmann Publishers Inc.}
\PrintBackRefs{\CurrentBib}

\bibitem [\protect \citeauthoryear {%
Garc{\'i}a%
\ \BBA {} Herrera%
}{%
Garc{\'i}a%
\ \BBA {} Herrera%
}{%
{\protect \APACyear {2008}}%
}]{%
garciaExtensionStatisticalComparisons2008}
\APACinsertmetastar {%
garciaExtensionStatisticalComparisons2008}%
\begin{APACrefauthors}%
Garc{\'i}a, S.%
\BCBT {}\ \BBA {} Herrera, F.%
\end{APACrefauthors}%
\unskip\
\newblock
\APACrefYearMonthDay{2008}{}{}.
\newblock
{\BBOQ}\APACrefatitle {An {{Extension}} on ``{{Statistical Comparisons}} of {{Classifiers}} over {{Multiple Data Sets}}'' for All {{Pairwise Comparisons}}} {An {{Extension}} on ``{{Statistical Comparisons}} of {{Classifiers}} over {{Multiple Data Sets}}'' for all {{Pairwise Comparisons}}}.{\BBCQ}
\newblock
\APACjournalVolNumPages{Journal of Machine Learning Research}{9}{89}{2677--2694,}
\newblock

\newblock

\PrintBackRefs{\CurrentBib}

\bibitem [\protect \citeauthoryear {%
Geurts%
, Ernst%
\BCBL {}\ \BBA {} Wehenkel%
}{%
Geurts%
\ \protect \BOthers {.}}{%
{\protect \APACyear {2006}}%
}]{%
geurtsExtremelyRandomizedTrees2006a}
\APACinsertmetastar {%
geurtsExtremelyRandomizedTrees2006a}%
\begin{APACrefauthors}%
Geurts, P.%
, Ernst, D.%
\BCBL {} Wehenkel, L.%
\end{APACrefauthors}%
\unskip\
\newblock
\APACrefYearMonthDay{2006}{{\APACmonth{04}}}{}.
\newblock
{\BBOQ}\APACrefatitle {Extremely Randomized Trees} {Extremely randomized trees}.{\BBCQ}
\newblock
\APACjournalVolNumPages{Machine Learning}{63}{1}{3--42,}
\newblock
\begin{APACrefDOI} \doi{10.1007/s10994-006-6226-1} \end{APACrefDOI}
\newblock

\newblock

\PrintBackRefs{\CurrentBib}

\bibitem [\protect \citeauthoryear {%
Gewers%
\ \protect \BOthers {.}}{%
Gewers%
\ \protect \BOthers {.}}{%
{\protect \APACyear {2021}}%
}]{%
gewersPrincipalComponentAnalysis2021}
\APACinsertmetastar {%
gewersPrincipalComponentAnalysis2021}%
\begin{APACrefauthors}%
Gewers, F.L.%
, Ferreira, G.R.%
, Arruda, H.F.D.%
, Silva, F.N.%
, Comin, C.H.%
, Amancio, D.R.%
\BCBL {} Costa, L.D.F.%
\end{APACrefauthors}%
\unskip\
\newblock
\APACrefYearMonthDay{2021}{{\APACmonth{05}}}{}.
\newblock
{\BBOQ}\APACrefatitle {Principal {{Component Analysis}}: {{A Natural Approach}} to {{Data Exploration}}} {Principal {{Component Analysis}}: {{A Natural Approach}} to {{Data Exploration}}}.{\BBCQ}
\newblock
\APACjournalVolNumPages{ACM Comput. Surv.}{54}{4}{70:1--70:34,}
\newblock
\begin{APACrefDOI} \doi{10.1145/3447755} \end{APACrefDOI}
\newblock

\newblock

\PrintBackRefs{\CurrentBib}

\bibitem [\protect \citeauthoryear {%
Guillaume%
, Vrain%
\BCBL {}\ \BBA {} Elloumi%
}{%
Guillaume%
\ \protect \BOthers {.}}{%
{\protect \APACyear {2022}}%
}]{%
guillaumeRandomDilatedShapelet2022}
\APACinsertmetastar {%
guillaumeRandomDilatedShapelet2022}%
\begin{APACrefauthors}%
Guillaume, A.%
, Vrain, C.%
\BCBL {} Elloumi, W.%
\end{APACrefauthors}%
\unskip\
\newblock
\APACrefYearMonthDay{2022}{{\APACmonth{06}}}{}.
\newblock
{\BBOQ}\APACrefatitle {Random {{Dilated Shapelet Transform}}: {{A New Approach}} for~{{Time Series Shapelets}}} {Random {{Dilated Shapelet Transform}}: {{A New Approach}} for~{{Time Series Shapelets}}}.{\BBCQ}
\newblock
 \APACrefbtitle {Pattern {{Recognition}} and {{Artificial Intelligence}}: {{Third International Conference}}, {{ICPRAI}} 2022, {{Paris}}, {{France}}, {{June}} 1--3, 2022, {{Proceedings}}, {{Part I}}} {Pattern {{Recognition}} and {{Artificial Intelligence}}: {{Third International Conference}}, {{ICPRAI}} 2022, {{Paris}}, {{France}}, {{June}} 1--3, 2022, {{Proceedings}}, {{Part I}}}\ (\BPGS\ 653--664).
\newblock
\APACaddressPublisher{Berlin, Heidelberg}{Springer-Verlag}.
\PrintBackRefs{\CurrentBib}

\bibitem [\protect \citeauthoryear {%
Harris%
\ \protect \BOthers {.}}{%
Harris%
\ \protect \BOthers {.}}{%
{\protect \APACyear {2020}}%
}]{%
harrisArrayProgrammingNumPy2020}
\APACinsertmetastar {%
harrisArrayProgrammingNumPy2020}%
\begin{APACrefauthors}%
Harris, C.R.%
, Millman, K.J.%
, {van der Walt}, S.J.%
, Gommers, R.%
, Virtanen, P.%
, Cournapeau, D.%
\BDBL {}Oliphant, T.E.%
\end{APACrefauthors}%
\unskip\
\newblock
\APACrefYearMonthDay{2020}{{\APACmonth{09}}}{}.
\newblock
{\BBOQ}\APACrefatitle {Array Programming with {{NumPy}}} {Array programming with {{NumPy}}}.{\BBCQ}
\newblock
\APACjournalVolNumPages{Nature}{585}{7825}{357--362,}
\newblock
\begin{APACrefDOI} \doi{10.1038/s41586-020-2649-2} \end{APACrefDOI}
\newblock

\newblock

\PrintBackRefs{\CurrentBib}

\bibitem [\protect \citeauthoryear {%
Herrmann%
\ \BBA {} Webb%
}{%
Herrmann%
\ \BBA {} Webb%
}{%
{\protect \APACyear {2023}}%
}]{%
herrmannAmercingIntuitiveEffective2023}
\APACinsertmetastar {%
herrmannAmercingIntuitiveEffective2023}%
\begin{APACrefauthors}%
Herrmann, M.%
\BCBT {}\ \BBA {} Webb, G.I.%
\end{APACrefauthors}%
\unskip\
\newblock
\APACrefYearMonthDay{2023}{{\APACmonth{05}}}{}.
\newblock
{\BBOQ}\APACrefatitle {Amercing: {{An}} Intuitive and Effective Constraint for Dynamic Time Warping} {Amercing: {{An}} intuitive and effective constraint for dynamic time warping}.{\BBCQ}
\newblock
\APACjournalVolNumPages{Pattern Recognition}{137}{}{109333,}
\newblock
\begin{APACrefDOI} \doi{10.1016/j.patcog.2023.109333} \end{APACrefDOI}
\newblock

\newblock

\PrintBackRefs{\CurrentBib}

\bibitem [\protect \citeauthoryear {%
Hills%
, Lines%
, Baranauskas%
, Mapp%
\BCBL {}\ \BBA {} Bagnall%
}{%
Hills%
\ \protect \BOthers {.}}{%
{\protect \APACyear {2014}}%
}]{%
hillsClassificationTimeSeries2014a}
\APACinsertmetastar {%
hillsClassificationTimeSeries2014a}%
\begin{APACrefauthors}%
Hills, J.%
, Lines, J.%
, Baranauskas, E.%
, Mapp, J.%
\BCBL {} Bagnall, A.%
\end{APACrefauthors}%
\unskip\
\newblock
\APACrefYearMonthDay{2014}{{\APACmonth{07}}}{}.
\newblock
{\BBOQ}\APACrefatitle {Classification of Time Series by Shapelet Transformation} {Classification of time series by shapelet transformation}.{\BBCQ}
\newblock
\APACjournalVolNumPages{Data Min. Knowl. Discov.}{28}{4}{851--881,}
\newblock
\begin{APACrefDOI} \doi{10.1007/s10618-013-0322-1} \end{APACrefDOI}
\newblock

\newblock

\PrintBackRefs{\CurrentBib}

\bibitem [\protect \citeauthoryear {%
Hoerl%
\ \BBA {} Kennard%
}{%
Hoerl%
\ \BBA {} Kennard%
}{%
{\protect \APACyear {1970}}%
{\protect \APACexlab {{\protect \BCnt {1}}}}}]{%
hoerlRidgeRegressionApplications1970}
\APACinsertmetastar {%
hoerlRidgeRegressionApplications1970}%
\begin{APACrefauthors}%
Hoerl, A.E.%
\BCBT {}\ \BBA {} Kennard, R.W.%
\end{APACrefauthors}%
\unskip\
\newblock
\APACrefYearMonthDay{1970{\protect \BCnt {1}}}{}{}.
\newblock
{\BBOQ}\APACrefatitle {Ridge {{Regression}}: {{Applications}} to {{Nonorthogonal Problems}}} {Ridge {{Regression}}: {{Applications}} to {{Nonorthogonal Problems}}}.{\BBCQ}
\newblock
\APACjournalVolNumPages{Technometrics}{12}{1}{69--82,}
\newblock
\begin{APACrefDOI} \doi{10.2307/1267352} \end{APACrefDOI}
\newblock
{\href{https://arxiv.org/abs/1267352}{{1267352}}}
\newblock

\PrintBackRefs{\CurrentBib}

\bibitem [\protect \citeauthoryear {%
Hoerl%
\ \BBA {} Kennard%
}{%
Hoerl%
\ \BBA {} Kennard%
}{%
{\protect \APACyear {1970}}%
{\protect \APACexlab {{\protect \BCnt {2}}}}}]{%
hoerlRidgeRegressionBiased1970}
\APACinsertmetastar {%
hoerlRidgeRegressionBiased1970}%
\begin{APACrefauthors}%
Hoerl, A.E.%
\BCBT {}\ \BBA {} Kennard, R.W.%
\end{APACrefauthors}%
\unskip\
\newblock
\APACrefYearMonthDay{1970{\protect \BCnt {2}}}{}{}.
\newblock
{\BBOQ}\APACrefatitle {Ridge {{Regression}}: {{Biased Estimation}} for {{Nonorthogonal Problems}}} {Ridge {{Regression}}: {{Biased Estimation}} for {{Nonorthogonal Problems}}}.{\BBCQ}
\newblock
\APACjournalVolNumPages{Technometrics}{12}{1}{55--67,}
\newblock
\begin{APACrefDOI} \doi{10.2307/1267351} \end{APACrefDOI}
\newblock
{\href{https://arxiv.org/abs/1267351}{{1267351}}}
\newblock

\PrintBackRefs{\CurrentBib}

\bibitem [\protect \citeauthoryear {%
Holder%
\ \BBA {} Bagnall%
}{%
Holder%
\ \BBA {} Bagnall%
}{%
{\protect \APACyear {2024}}%
}]{%
holderRockKASBABlazingly2024}
\APACinsertmetastar {%
holderRockKASBABlazingly2024}%
\begin{APACrefauthors}%
Holder, C.%
\BCBT {}\ \BBA {} Bagnall, A.%
\end{APACrefauthors}%
\unskip\
\newblock
\APACrefYearMonthDay{2024}{{\APACmonth{11}}}{}.
\newblock
\APACrefbtitle {Rock the {{KASBA}}: {{Blazingly Fast}} and {{Accurate Time Series Clustering}}} {Rock the {{KASBA}}: {{Blazingly Fast}} and {{Accurate Time Series Clustering}}}\ (\BNUM\ arXiv:2411.17838).
\newblock
\APACaddressPublisher{}{arXiv}.
\PrintBackRefs{\CurrentBib}

\bibitem [\protect \citeauthoryear {%
Holder%
, {Guijo-Rubio}%
\BCBL {}\ \BBA {} Bagnall%
}{%
Holder%
\ \protect \BOthers {.}}{%
{\protect \APACyear {2023}}%
{\protect \APACexlab {{\protect \BCnt {1}}}}}]{%
holderBarycentreAveragingMoveSplitMerge2023}
\APACinsertmetastar {%
holderBarycentreAveragingMoveSplitMerge2023}%
\begin{APACrefauthors}%
Holder, C.%
, {Guijo-Rubio}, D.%
\BCBL {} Bagnall, A.%
\end{APACrefauthors}%
\unskip\
\newblock
\APACrefYearMonthDay{2023{\protect \BCnt {1}}}{{\APACmonth{11}}}{}.
\newblock
{\BBOQ}\APACrefatitle {Barycentre {{Averaging}} for the {{Move-Split-Merge Time Series Distance Measure}}} {Barycentre {{Averaging}} for the {{Move-Split-Merge Time Series Distance Measure}}}.{\BBCQ}
\newblock
 \APACrefbtitle {15th {{International Conference}} on {{Knowledge Discovery}} and {{Information Retrieval}}} {15th {{International Conference}} on {{Knowledge Discovery}} and {{Information Retrieval}}}\ (\BPGS\ 51--62).
\PrintBackRefs{\CurrentBib}

\bibitem [\protect \citeauthoryear {%
Holder%
, {Guijo-Rubio}%
\BCBL {}\ \BBA {} Bagnall%
}{%
Holder%
\ \protect \BOthers {.}}{%
{\protect \APACyear {2023}}%
{\protect \APACexlab {{\protect \BCnt {2}}}}}]{%
holderClusteringTimeSeries2023}
\APACinsertmetastar {%
holderClusteringTimeSeries2023}%
\begin{APACrefauthors}%
Holder, C.%
, {Guijo-Rubio}, D.%
\BCBL {} Bagnall, A.%
\end{APACrefauthors}%
\unskip\
\newblock
\APACrefYearMonthDay{2023{\protect \BCnt {2}}}{}{}.
\newblock
{\BBOQ}\APACrefatitle {Clustering {{Time Series}} with~K-{{Medoids Based Algorithms}}} {Clustering {{Time Series}} with~k-{{Medoids Based Algorithms}}}.{\BBCQ}
\newblock
 G.~Ifrim\ \BOthers {.}\ (\BEDS), \APACrefbtitle {Advanced {{Analytics}} and {{Learning}} on {{Temporal Data}}} {Advanced {{Analytics}} and {{Learning}} on {{Temporal Data}}}\ (\BPGS\ 39--55).
\newblock
\APACaddressPublisher{Cham}{Springer Nature Switzerland}.
\PrintBackRefs{\CurrentBib}

\bibitem [\protect \citeauthoryear {%
Holder%
, Middlehurst%
\BCBL {}\ \BBA {} Bagnall%
}{%
Holder%
\ \protect \BOthers {.}}{%
{\protect \APACyear {2024}}%
}]{%
holderReviewEvaluationElastic2024}
\APACinsertmetastar {%
holderReviewEvaluationElastic2024}%
\begin{APACrefauthors}%
Holder, C.%
, Middlehurst, M.%
\BCBL {} Bagnall, A.%
\end{APACrefauthors}%
\unskip\
\newblock
\APACrefYearMonthDay{2024}{{\APACmonth{02}}}{}.
\newblock
{\BBOQ}\APACrefatitle {A Review and Evaluation of Elastic Distance Functions for Time Series Clustering} {A review and evaluation of elastic distance functions for time series clustering}.{\BBCQ}
\newblock
\APACjournalVolNumPages{Knowledge and Information Systems}{66}{2}{765--809,}
\newblock
\begin{APACrefDOI} \doi{10.1007/s10115-023-01952-0} \end{APACrefDOI}
\newblock

\newblock

\PrintBackRefs{\CurrentBib}

\bibitem [\protect \citeauthoryear {%
Hunter%
}{%
Hunter%
}{%
{\protect \APACyear {2007}}%
}]{%
hunterMatplotlib2DGraphics2007}
\APACinsertmetastar {%
hunterMatplotlib2DGraphics2007}%
\begin{APACrefauthors}%
Hunter, J.D.%
\end{APACrefauthors}%
\unskip\
\newblock
\APACrefYearMonthDay{2007}{{\APACmonth{05}}}{}.
\newblock
{\BBOQ}\APACrefatitle {Matplotlib: {{A 2D Graphics Environment}}} {Matplotlib: {{A 2D Graphics Environment}}}.{\BBCQ}
\newblock
\APACjournalVolNumPages{Computing in Science Engineering}{9}{3}{90--95,}
\newblock
\begin{APACrefDOI} \doi{10.1109/MCSE.2007.55} \end{APACrefDOI}
\newblock

\newblock

\PrintBackRefs{\CurrentBib}

\bibitem [\protect \citeauthoryear {%
{Ismail-Fawaz}%
, Dempster%
\BCBL {}\ \protect \BOthers {.}}{%
{Ismail-Fawaz}%
, Dempster%
\BCBL {}\ \protect \BOthers {.}}{%
{\protect \APACyear {2023}}%
}]{%
ismail-fawazApproachMultipleComparison2023}
\APACinsertmetastar {%
ismail-fawazApproachMultipleComparison2023}%
\begin{APACrefauthors}%
{Ismail-Fawaz}, A.%
, Dempster, A.%
, Tan, C.W.%
, Herrmann, M.%
, Miller, L.%
, Schmidt, D.F.%
\BDBL {}Webb, G.I.%
\end{APACrefauthors}%
\unskip\
\newblock
\APACrefYearMonthDay{2023}{{\APACmonth{05}}}{}.
\newblock
\APACrefbtitle {An {{Approach}} to {{Multiple Comparison Benchmark Evaluations}} That Is {{Stable Under Manipulation}} of the {{Comparate Set}}} {An {{Approach}} to {{Multiple Comparison Benchmark Evaluations}} that is {{Stable Under Manipulation}} of the {{Comparate Set}}}\ (\BNUM\ arXiv:2305.11921).
\newblock
\APACaddressPublisher{}{arXiv}.
\PrintBackRefs{\CurrentBib}

\bibitem [\protect \citeauthoryear {%
{Ismail-Fawaz}%
, Ismail~Fawaz%
\BCBL {}\ \protect \BOthers {.}}{%
{Ismail-Fawaz}%
, Ismail~Fawaz%
\BCBL {}\ \protect \BOthers {.}}{%
{\protect \APACyear {2023}}%
}]{%
ismail-fawazShapeDBAGeneratingEffective2023}
\APACinsertmetastar {%
ismail-fawazShapeDBAGeneratingEffective2023}%
\begin{APACrefauthors}%
{Ismail-Fawaz}, A.%
, Ismail~Fawaz, H.%
, Petitjean, F.%
, Devanne, M.%
, Weber, J.%
, Berretti, S.%
\BDBL {}Forestier, G.%
\end{APACrefauthors}%
\unskip\
\newblock
\APACrefYearMonthDay{2023}{}{}.
\newblock
{\BBOQ}\APACrefatitle {{{ShapeDBA}}: {{Generating Effective Time Series Prototypes Using ShapeDTW Barycenter Averaging}}} {{{ShapeDBA}}: {{Generating Effective Time Series Prototypes Using ShapeDTW Barycenter Averaging}}}.{\BBCQ}
\newblock
 G.~Ifrim\ \BOthers {.}\ (\BEDS), \APACrefbtitle {Advanced {{Analytics}} and {{Learning}} on {{Temporal Data}}} {Advanced {{Analytics}} and {{Learning}} on {{Temporal Data}}}\ (\BPGS\ 127--142).
\newblock
\APACaddressPublisher{Cham}{Springer Nature Switzerland}.
\PrintBackRefs{\CurrentBib}

\bibitem [\protect \citeauthoryear {%
Itakura%
}{%
Itakura%
}{%
{\protect \APACyear {1975}}%
}]{%
itakuraMinimumPredictionResidual1975}
\APACinsertmetastar {%
itakuraMinimumPredictionResidual1975}%
\begin{APACrefauthors}%
Itakura, F.%
\end{APACrefauthors}%
\unskip\
\newblock
\APACrefYearMonthDay{1975}{{\APACmonth{02}}}{}.
\newblock
{\BBOQ}\APACrefatitle {Minimum Prediction Residual Principle Applied to Speech Recognition} {Minimum prediction residual principle applied to speech recognition}.{\BBCQ}
\newblock
\APACjournalVolNumPages{IEEE Transactions on Acoustics, Speech, and Signal Processing}{23}{1}{67--72,}
\newblock
\begin{APACrefDOI} \doi{10.1109/TASSP.1975.1162641} \end{APACrefDOI}
\newblock

\newblock

\PrintBackRefs{\CurrentBib}

\bibitem [\protect \citeauthoryear {%
Jain%
\ \BBA {} Dubes%
}{%
Jain%
\ \BBA {} Dubes%
}{%
{\protect \APACyear {1988}}%
}]{%
jainAlgorithmsClusteringData1988}
\APACinsertmetastar {%
jainAlgorithmsClusteringData1988}%
\begin{APACrefauthors}%
Jain, A.K.%
\BCBT {}\ \BBA {} Dubes, R.C.%
\end{APACrefauthors}%
\unskip\
\newblock
\APACrefYear{1988}.
\newblock
\APACrefbtitle {Algorithms for Clustering Data} {Algorithms for clustering data}.
\newblock
\APACaddressPublisher{USA}{Prentice-Hall, Inc.}
\PrintBackRefs{\CurrentBib}

\bibitem [\protect \citeauthoryear {%
Javed%
, Lee%
\BCBL {}\ \BBA {} Rizzo%
}{%
Javed%
\ \protect \BOthers {.}}{%
{\protect \APACyear {2020}}%
}]{%
javedBenchmarkStudyTime2020}
\APACinsertmetastar {%
javedBenchmarkStudyTime2020}%
\begin{APACrefauthors}%
Javed, A.%
, Lee, B.S.%
\BCBL {} Rizzo, D.M.%
\end{APACrefauthors}%
\unskip\
\newblock
\APACrefYearMonthDay{2020}{{\APACmonth{09}}}{}.
\newblock
{\BBOQ}\APACrefatitle {A Benchmark Study on Time Series Clustering} {A benchmark study on time series clustering}.{\BBCQ}
\newblock
\APACjournalVolNumPages{Machine Learning with Applications}{1}{}{100001,}
\newblock
\begin{APACrefDOI} \doi{10.1016/j.mlwa.2020.100001} \end{APACrefDOI}
\newblock

\newblock

\PrintBackRefs{\CurrentBib}

\bibitem [\protect \citeauthoryear {%
Jeong%
, Jeong%
\BCBL {}\ \BBA {} Omitaomu%
}{%
Jeong%
\ \protect \BOthers {.}}{%
{\protect \APACyear {2011}}%
}]{%
jeongWeightedDynamicTime2011}
\APACinsertmetastar {%
jeongWeightedDynamicTime2011}%
\begin{APACrefauthors}%
Jeong, Y\BHBI S.%
, Jeong, M.K.%
\BCBL {} Omitaomu, O.A.%
\end{APACrefauthors}%
\unskip\
\newblock
\APACrefYearMonthDay{2011}{{\APACmonth{09}}}{}.
\newblock
{\BBOQ}\APACrefatitle {Weighted Dynamic Time Warping for Time Series Classification} {Weighted dynamic time warping for time series classification}.{\BBCQ}
\newblock
\APACjournalVolNumPages{Pattern Recognition}{44}{9}{2231--2240,}
\newblock
\begin{APACrefDOI} \doi{10.1016/j.patcog.2010.09.022} \end{APACrefDOI}
\newblock

\newblock

\PrintBackRefs{\CurrentBib}

\bibitem [\protect \citeauthoryear {%
Jorge%
\ \BBA {} Rub{\'e}n%
}{%
Jorge%
\ \BBA {} Rub{\'e}n%
}{%
{\protect \APACyear {2024}}%
}]{%
jorgeTimeSeriesClustering2024}
\APACinsertmetastar {%
jorgeTimeSeriesClustering2024}%
\begin{APACrefauthors}%
Jorge, M\BHBI B.%
\BCBT {}\ \BBA {} Rub{\'e}n, C.%
\end{APACrefauthors}%
\unskip\
\newblock
\APACrefYearMonthDay{2024}{{\APACmonth{07}}}{}.
\newblock
{\BBOQ}\APACrefatitle {Time Series Clustering with Random Convolutional Kernels} {Time series clustering with random convolutional kernels}.{\BBCQ}
\newblock
\APACjournalVolNumPages{Data Mining and Knowledge Discovery}{38}{4}{1862--1888,}
\newblock
\begin{APACrefDOI} \doi{10.1007/s10618-024-01018-x} \end{APACrefDOI}
\newblock

\newblock

\PrintBackRefs{\CurrentBib}

\bibitem [\protect \citeauthoryear {%
Keogh%
, Lonardi%
\BCBL {}\ \BBA {} Ratanamahatana%
}{%
Keogh%
\ \protect \BOthers {.}}{%
{\protect \APACyear {2004}}%
}]{%
keoghParameterfreeDataMining2004}
\APACinsertmetastar {%
keoghParameterfreeDataMining2004}%
\begin{APACrefauthors}%
Keogh, E.%
, Lonardi, S.%
\BCBL {} Ratanamahatana, C.A.%
\end{APACrefauthors}%
\unskip\
\newblock
\APACrefYearMonthDay{2004}{{\APACmonth{08}}}{}.
\newblock
{\BBOQ}\APACrefatitle {Towards Parameter-Free Data Mining} {Towards parameter-free data mining}.{\BBCQ}
\newblock
 \APACrefbtitle {Proceedings of the Tenth {{ACM SIGKDD}} International Conference on {{Knowledge}} Discovery and Data Mining} {Proceedings of the tenth {{ACM SIGKDD}} international conference on {{Knowledge}} discovery and data mining}\ (\BPGS\ 206--215).
\newblock
\APACaddressPublisher{Seattle WA USA}{ACM}.
\PrintBackRefs{\CurrentBib}

\bibitem [\protect \citeauthoryear {%
Lafabregue%
, Weber%
, Gan{\c c}arski%
\BCBL {}\ \BBA {} Forestier%
}{%
Lafabregue%
\ \protect \BOthers {.}}{%
{\protect \APACyear {2022}}%
}]{%
lafabregueEndtoendDeepRepresentation2022}
\APACinsertmetastar {%
lafabregueEndtoendDeepRepresentation2022}%
\begin{APACrefauthors}%
Lafabregue, B.%
, Weber, J.%
, Gan{\c c}arski, P.%
\BCBL {} Forestier, G.%
\end{APACrefauthors}%
\unskip\
\newblock
\APACrefYearMonthDay{2022}{{\APACmonth{01}}}{}.
\newblock
{\BBOQ}\APACrefatitle {End-to-End Deep Representation Learning for Time Series Clustering: A Comparative Study} {End-to-end deep representation learning for time series clustering: A comparative study}.{\BBCQ}
\newblock
\APACjournalVolNumPages{Data Min. Knowl. Discov.}{36}{1}{29--81,}
\newblock
\begin{APACrefDOI} \doi{10.1007/s10618-021-00796-y} \end{APACrefDOI}
\newblock

\newblock

\PrintBackRefs{\CurrentBib}

\bibitem [\protect \citeauthoryear {%
Leonard~Kaufman%
}{%
Leonard~Kaufman%
}{%
{\protect \APACyear {1990}}%
}]{%
leonard1990partitioning}
\APACinsertmetastar {%
leonard1990partitioning}%
\begin{APACrefauthors}%
Leonard~Kaufman, P.%
\end{APACrefauthors}%
\unskip\
\newblock
\APACrefYearMonthDay{1990}{}{}.
\newblock
\APACrefbtitle {Partitioning around medoids (program PAM), chapter 2.} {Partitioning around medoids (program pam), chapter 2.}
\newblock
\APACaddressPublisher{}{Wiley}.
\PrintBackRefs{\CurrentBib}

\bibitem [\protect \citeauthoryear {%
Levy%
\ \BBA {} Lindenbaum%
}{%
Levy%
\ \BBA {} Lindenbaum%
}{%
{\protect \APACyear {1998}}%
}]{%
levySequentialKarhunenLoeveBasis1998}
\APACinsertmetastar {%
levySequentialKarhunenLoeveBasis1998}%
\begin{APACrefauthors}%
Levy, A.%
\BCBT {}\ \BBA {} Lindenbaum, M.%
\end{APACrefauthors}%
\unskip\
\newblock
\APACrefYearMonthDay{1998}{{\APACmonth{10}}}{}.
\newblock
{\BBOQ}\APACrefatitle {Sequential {{Karhunen-Loeve}} Basis Extraction and Its Application to Images} {Sequential {{Karhunen-Loeve}} basis extraction and its application to images}.{\BBCQ}
\newblock
 \APACrefbtitle {Proceedings 1998 {{International Conference}} on {{Image Processing}}. {{ICIP98}} ({{Cat}}. {{No}}.{{98CB36269}})} {Proceedings 1998 {{International Conference}} on {{Image Processing}}. {{ICIP98}} ({{Cat}}. {{No}}.{{98CB36269}})}\ (\BVOL~2, \BPG~456-460 vol.2).
\PrintBackRefs{\CurrentBib}

\bibitem [\protect \citeauthoryear {%
A.~Li%
\ \protect \BOthers {.}}{%
A.~Li%
\ \protect \BOthers {.}}{%
{\protect \APACyear {2023}}%
}]{%
liAngClustAngleFeatureBased2023}
\APACinsertmetastar {%
liAngClustAngleFeatureBased2023}%
\begin{APACrefauthors}%
Li, A.%
, Xiong, S.%
, Li, J.%
, Mallik, S.%
, Liu, Y.%
, Fei, R.%
\BDBL {}Liu, G.%
\end{APACrefauthors}%
\unskip\
\newblock
\APACrefYearMonthDay{2023}{{\APACmonth{03}}}{}.
\newblock
{\BBOQ}\APACrefatitle {{{AngClust}}: {{Angle Feature-Based Clustering}} for {{Short Time Series Gene Expression Profiles}}} {{{AngClust}}: {{Angle Feature-Based Clustering}} for {{Short Time Series Gene Expression Profiles}}}.{\BBCQ}
\newblock
\APACjournalVolNumPages{IEEE/ACM Transactions on Computational Biology and Bioinformatics}{20}{2}{1574--1580,}
\newblock
\begin{APACrefDOI} \doi{10.1109/TCBB.2022.3192306} \end{APACrefDOI}
\newblock

\newblock

\PrintBackRefs{\CurrentBib}

\bibitem [\protect \citeauthoryear {%
X.~Li%
, Lin%
\BCBL {}\ \BBA {} Zhao%
}{%
X.~Li%
\ \protect \BOthers {.}}{%
{\protect \APACyear {2019}}%
}]{%
ijcai2019p406}
\APACinsertmetastar {%
ijcai2019p406}%
\begin{APACrefauthors}%
Li, X.%
, Lin, J.%
\BCBL {} Zhao, L.%
\end{APACrefauthors}%
\unskip\
\newblock
\APACrefYearMonthDay{2019}{{\APACmonth{07}}}{}.
\newblock
{\BBOQ}\APACrefatitle {Linear Time Complexity Time Series Clustering with Symbolic Pattern Forest} {Linear time complexity time series clustering with symbolic pattern forest}.{\BBCQ}
\newblock
 \APACrefbtitle {Proceedings of the Twenty-Eighth International Joint Conference on Artificial Intelligence, {{IJCAI-19}}} {Proceedings of the twenty-eighth international joint conference on artificial intelligence, {{IJCAI-19}}}\ (\BPGS\ 2930--2936).
\newblock
\APACaddressPublisher{}{International Joint Conferences on Artificial Intelligence Organization}.
\PrintBackRefs{\CurrentBib}

\bibitem [\protect \citeauthoryear {%
X.~Li%
, Xi%
\BCBL {}\ \BBA {} Lin%
}{%
X.~Li%
\ \protect \BOthers {.}}{%
{\protect \APACyear {2024}}%
}]{%
liRandomnetClusteringTime2024}
\APACinsertmetastar {%
liRandomnetClusteringTime2024}%
\begin{APACrefauthors}%
Li, X.%
, Xi, W.%
\BCBL {} Lin, J.%
\end{APACrefauthors}%
\unskip\
\newblock
\APACrefYearMonthDay{2024}{{\APACmonth{11}}}{}.
\newblock
{\BBOQ}\APACrefatitle {Randomnet: Clustering Time Series Using Untrained Deep Neural Networks} {Randomnet: Clustering time series using untrained deep neural networks}.{\BBCQ}
\newblock
\APACjournalVolNumPages{Data Mining and Knowledge Discovery}{38}{6}{3473--3502,}
\newblock
\begin{APACrefDOI} \doi{10.1007/s10618-024-01048-5} \end{APACrefDOI}
\newblock

\newblock

\PrintBackRefs{\CurrentBib}

\bibitem [\protect \citeauthoryear {%
Lin%
, Keogh%
, Wei%
\BCBL {}\ \BBA {} Lonardi%
}{%
Lin%
\ \protect \BOthers {.}}{%
{\protect \APACyear {2007}}%
}]{%
linExperiencingSAXNovel2007a}
\APACinsertmetastar {%
linExperiencingSAXNovel2007a}%
\begin{APACrefauthors}%
Lin, J.%
, Keogh, E.%
, Wei, L.%
\BCBL {} Lonardi, S.%
\end{APACrefauthors}%
\unskip\
\newblock
\APACrefYearMonthDay{2007}{{\APACmonth{10}}}{}.
\newblock
{\BBOQ}\APACrefatitle {Experiencing {{SAX}}: A Novel Symbolic Representation of Time Series} {Experiencing {{SAX}}: A novel symbolic representation of time series}.{\BBCQ}
\newblock
\APACjournalVolNumPages{Data Mining and Knowledge Discovery}{15}{2}{107--144,}
\newblock
\begin{APACrefDOI} \doi{10.1007/s10618-007-0064-z} \end{APACrefDOI}
\newblock

\newblock

\PrintBackRefs{\CurrentBib}

\bibitem [\protect \citeauthoryear {%
Lloyd%
}{%
Lloyd%
}{%
{\protect \APACyear {1982}}%
}]{%
lloydLeastSquaresQuantization1982a}
\APACinsertmetastar {%
lloydLeastSquaresQuantization1982a}%
\begin{APACrefauthors}%
Lloyd, S.%
\end{APACrefauthors}%
\unskip\
\newblock
\APACrefYearMonthDay{1982}{{\APACmonth{03}}}{}.
\newblock
{\BBOQ}\APACrefatitle {Least Squares Quantization in {{PCM}}} {Least squares quantization in {{PCM}}}.{\BBCQ}
\newblock
\APACjournalVolNumPages{IEEE Transactions on Information Theory}{28}{2}{129--137,}
\newblock
\begin{APACrefDOI} \doi{10.1109/TIT.1982.1056489} \end{APACrefDOI}
\newblock

\newblock

\PrintBackRefs{\CurrentBib}

\bibitem [\protect \citeauthoryear {%
Lubba%
\ \protect \BOthers {.}}{%
Lubba%
\ \protect \BOthers {.}}{%
{\protect \APACyear {2019}}%
}]{%
lubbaCatch22CAnonicalTimeseries2019}
\APACinsertmetastar {%
lubbaCatch22CAnonicalTimeseries2019}%
\begin{APACrefauthors}%
Lubba, C.H.%
, Sethi, S.S.%
, Knaute, P.%
, Schultz, S.R.%
, Fulcher, B.D.%
\BCBL {} Jones, N.S.%
\end{APACrefauthors}%
\unskip\
\newblock
\APACrefYearMonthDay{2019}{{\APACmonth{11}}}{}.
\newblock
{\BBOQ}\APACrefatitle {Catch22: {{CAnonical Time-series CHaracteristics}}} {Catch22: {{CAnonical Time-series CHaracteristics}}}.{\BBCQ}
\newblock
\APACjournalVolNumPages{Data Mining and Knowledge Discovery}{33}{6}{1821--1852,}
\newblock
\begin{APACrefDOI} \doi{10.1007/s10618-019-00647-x} \end{APACrefDOI}
\newblock

\newblock

\PrintBackRefs{\CurrentBib}

\bibitem [\protect \citeauthoryear {%
MacQueen%
}{%
MacQueen%
}{%
{\protect \APACyear {1967}}%
}]{%
macqueenMethodsClassificationAnalysis1967}
\APACinsertmetastar {%
macqueenMethodsClassificationAnalysis1967}%
\begin{APACrefauthors}%
MacQueen, J.%
\end{APACrefauthors}%
\unskip\
\newblock
\APACrefYearMonthDay{1967}{{\APACmonth{01}}}{}.
\newblock
{\BBOQ}\APACrefatitle {Some Methods for Classification and Analysis of Multivariate Observations} {Some methods for classification and analysis of multivariate observations}.{\BBCQ}
\newblock
 \APACrefbtitle {Proceedings of the {{Fifth Berkeley Symposium}} on {{Mathematical Statistics}} and {{Probability}}, {{Volume}} 1: {{Statistics}}} {Proceedings of the {{Fifth Berkeley Symposium}} on {{Mathematical Statistics}} and {{Probability}}, {{Volume}} 1: {{Statistics}}}\ (\BVOL~5.1, \BPGS\ 281--298).
\newblock
\APACaddressPublisher{}{University of California Press}.
\PrintBackRefs{\CurrentBib}

\bibitem [\protect \citeauthoryear {%
McInnes%
, Healy%
\BCBL {}\ \BBA {} Melville%
}{%
McInnes%
\ \protect \BOthers {.}}{%
{\protect \APACyear {2020}}%
}]{%
mcinnesUMAPUniformManifold2020}
\APACinsertmetastar {%
mcinnesUMAPUniformManifold2020}%
\begin{APACrefauthors}%
McInnes, L.%
, Healy, J.%
\BCBL {} Melville, J.%
\end{APACrefauthors}%
\unskip\
\newblock
\APACrefYearMonthDay{2020}{{\APACmonth{09}}}{}.
\newblock
\APACrefbtitle {{{UMAP}}: {{Uniform Manifold Approximation}} and {{Projection}} for {{Dimension Reduction}}} {{{UMAP}}: {{Uniform Manifold Approximation}} and {{Projection}} for {{Dimension Reduction}}}\ (\BNUM\ arXiv:1802.03426).
\newblock
\APACaddressPublisher{}{arXiv}.
\PrintBackRefs{\CurrentBib}

\bibitem [\protect \citeauthoryear {%
McInnes%
, Healy%
, Saul%
\BCBL {}\ \BBA {} Gro{\ss}berger%
}{%
McInnes%
\ \protect \BOthers {.}}{%
{\protect \APACyear {2018}}%
}]{%
mcinnesUMAPUniformManifold2018}
\APACinsertmetastar {%
mcinnesUMAPUniformManifold2018}%
\begin{APACrefauthors}%
McInnes, L.%
, Healy, J.%
, Saul, N.%
\BCBL {} Gro{\ss}berger, L.%
\end{APACrefauthors}%
\unskip\
\newblock
\APACrefYearMonthDay{2018}{{\APACmonth{09}}}{}.
\newblock
{\BBOQ}\APACrefatitle {{{UMAP}}: {{Uniform Manifold Approximation}} and {{Projection}}} {{{UMAP}}: {{Uniform Manifold Approximation}} and {{Projection}}}.{\BBCQ}
\newblock
\APACjournalVolNumPages{Journal of Open Source Software}{3}{29}{861,}
\newblock
\begin{APACrefDOI} \doi{10.21105/joss.00861} \end{APACrefDOI}
\newblock

\newblock

\PrintBackRefs{\CurrentBib}

\bibitem [\protect \citeauthoryear {%
Middlehurst%
, {Ismail-Fawaz}%
\BCBL {}\ \protect \BOthers {.}}{%
Middlehurst%
, {Ismail-Fawaz}%
\BCBL {}\ \protect \BOthers {.}}{%
{\protect \APACyear {2024}}%
}]{%
middlehurstAeonPythonToolkit2024}
\APACinsertmetastar {%
middlehurstAeonPythonToolkit2024}%
\begin{APACrefauthors}%
Middlehurst, M.%
, {Ismail-Fawaz}, A.%
, Guillaume, A.%
, Holder, C.%
, {Guijo-Rubio}, D.%
, Bulatova, G.%
\BDBL {}Bagnall, A.%
\end{APACrefauthors}%
\unskip\
\newblock
\APACrefYearMonthDay{2024}{}{}.
\newblock
{\BBOQ}\APACrefatitle {Aeon: A {{Python Toolkit}} for {{Learning}} from {{Time Series}}} {Aeon: A {{Python Toolkit}} for {{Learning}} from {{Time Series}}}.{\BBCQ}
\newblock
\APACjournalVolNumPages{Journal of Machine Learning Research}{25}{289}{1--10,}
\newblock

\newblock

\PrintBackRefs{\CurrentBib}

\bibitem [\protect \citeauthoryear {%
Middlehurst%
, Large%
\BCBL {}\ \BBA {} Bagnall%
}{%
Middlehurst%
\ \protect \BOthers {.}}{%
{\protect \APACyear {2020}}%
}]{%
middlehurstCanonicalIntervalForest2020}
\APACinsertmetastar {%
middlehurstCanonicalIntervalForest2020}%
\begin{APACrefauthors}%
Middlehurst, M.%
, Large, J.%
\BCBL {} Bagnall, A.%
\end{APACrefauthors}%
\unskip\
\newblock
\APACrefYearMonthDay{2020}{{\APACmonth{12}}}{}.
\newblock
{\BBOQ}\APACrefatitle {The {{Canonical Interval Forest}} ({{CIF}}) {{Classifier}} for {{Time Series Classification}}} {The {{Canonical Interval Forest}} ({{CIF}}) {{Classifier}} for {{Time Series Classification}}}.{\BBCQ}
\newblock
 \APACrefbtitle {2020 {{IEEE International Conference}} on {{Big Data}} ({{Big Data}})} {2020 {{IEEE International Conference}} on {{Big Data}} ({{Big Data}})}\ (\BPGS\ 188--195).
\PrintBackRefs{\CurrentBib}

\bibitem [\protect \citeauthoryear {%
Middlehurst%
\ \protect \BOthers {.}}{%
Middlehurst%
\ \protect \BOthers {.}}{%
{\protect \APACyear {2021}}%
}]{%
middlehurstHIVECOTE20New2021a}
\APACinsertmetastar {%
middlehurstHIVECOTE20New2021a}%
\begin{APACrefauthors}%
Middlehurst, M.%
, Large, J.%
, Flynn, M.%
, Lines, J.%
, Bostrom, A.%
\BCBL {} Bagnall, A.%
\end{APACrefauthors}%
\unskip\
\newblock
\APACrefYearMonthDay{2021}{{\APACmonth{12}}}{}.
\newblock
{\BBOQ}\APACrefatitle {{{HIVE-COTE}} 2.0: A New Meta Ensemble for Time Series Classification} {{{HIVE-COTE}} 2.0: A new meta ensemble for time series classification}.{\BBCQ}
\newblock
\APACjournalVolNumPages{Machine Learning}{110}{11}{3211--3243,}
\newblock
\begin{APACrefDOI} \doi{10.1007/s10994-021-06057-9} \end{APACrefDOI}
\newblock

\newblock

\PrintBackRefs{\CurrentBib}

\bibitem [\protect \citeauthoryear {%
Middlehurst%
, Sch{\"a}fer%
\BCBL {}\ \BBA {} Bagnall%
}{%
Middlehurst%
, Sch{\"a}fer%
\BCBL {}\ \BBA {} Bagnall%
}{%
{\protect \APACyear {2024}}%
}]{%
middlehurstBakeReduxReview2024}
\APACinsertmetastar {%
middlehurstBakeReduxReview2024}%
\begin{APACrefauthors}%
Middlehurst, M.%
, Sch{\"a}fer, P.%
\BCBL {} Bagnall, A.%
\end{APACrefauthors}%
\unskip\
\newblock
\APACrefYearMonthDay{2024}{{\APACmonth{07}}}{}.
\newblock
{\BBOQ}\APACrefatitle {Bake off Redux: A Review and Experimental Evaluation of Recent Time Series Classification Algorithms} {Bake off redux: A review and experimental evaluation of recent time series classification algorithms}.{\BBCQ}
\newblock
\APACjournalVolNumPages{Data Mining and Knowledge Discovery}{38}{4}{1958--2031,}
\newblock
\begin{APACrefDOI} \doi{10.1007/s10618-024-01022-1} \end{APACrefDOI}
\newblock

\newblock

\PrintBackRefs{\CurrentBib}

\bibitem [\protect \citeauthoryear {%
Murtagh%
\ \BBA {} Contreras%
}{%
Murtagh%
\ \BBA {} Contreras%
}{%
{\protect \APACyear {2012}}%
}]{%
murtaghAlgorithmsHierarchicalClustering2012}
\APACinsertmetastar {%
murtaghAlgorithmsHierarchicalClustering2012}%
\begin{APACrefauthors}%
Murtagh, F.%
\BCBT {}\ \BBA {} Contreras, P.%
\end{APACrefauthors}%
\unskip\
\newblock
\APACrefYearMonthDay{2012}{}{}.
\newblock
{\BBOQ}\APACrefatitle {Algorithms for Hierarchical Clustering: An Overview} {Algorithms for hierarchical clustering: An overview}.{\BBCQ}
\newblock
\APACjournalVolNumPages{WIREs Data Mining and Knowledge Discovery}{2}{1}{86--97,}
\newblock
\begin{APACrefDOI} \doi{10.1002/widm.53} \end{APACrefDOI}
\newblock

\newblock

\PrintBackRefs{\CurrentBib}

\bibitem [\protect \citeauthoryear {%
Omohundro%
}{%
Omohundro%
}{%
{\protect \APACyear {1989}}%
}]{%
omohundro1989five}
\APACinsertmetastar {%
omohundro1989five}%
\begin{APACrefauthors}%
Omohundro, S.M.%
\end{APACrefauthors}%
\unskip\
\newblock
\APACrefYearMonthDay{1989}{}{}.
\newblock
\APACrefbtitle {Five balltree construction algorithms} {Five balltree construction algorithms}\ \APACbVolEdTR{}{\BTR{}}.
\newblock
\APACaddressInstitution{947 Center Street, Suite 600, Berkeley, California 94704}{International Computer Science Institute}.
\PrintBackRefs{\CurrentBib}

\bibitem [\protect \citeauthoryear {%
Oyewole%
\ \BBA {} Thopil%
}{%
Oyewole%
\ \BBA {} Thopil%
}{%
{\protect \APACyear {2023}}%
}]{%
oyewole2023data}
\APACinsertmetastar {%
oyewole2023data}%
\begin{APACrefauthors}%
Oyewole, G.J.%
\BCBT {}\ \BBA {} Thopil, G.A.%
\end{APACrefauthors}%
\unskip\
\newblock
\APACrefYearMonthDay{2023}{}{}.
\newblock
{\BBOQ}\APACrefatitle {Data clustering: application and trends} {Data clustering: application and trends}.{\BBCQ}
\newblock
\APACjournalVolNumPages{Artificial intelligence review}{56}{7}{6439--6475,}
\newblock

\newblock

\PrintBackRefs{\CurrentBib}

\bibitem [\protect \citeauthoryear {%
Paparrizos%
\ \BBA {} Gravano%
}{%
Paparrizos%
\ \BBA {} Gravano%
}{%
{\protect \APACyear {2016}}%
}]{%
paparrizosKShapeEfficientAccurate2016}
\APACinsertmetastar {%
paparrizosKShapeEfficientAccurate2016}%
\begin{APACrefauthors}%
Paparrizos, J.%
\BCBT {}\ \BBA {} Gravano, L.%
\end{APACrefauthors}%
\unskip\
\newblock
\APACrefYearMonthDay{2016}{{\APACmonth{06}}}{}.
\newblock
{\BBOQ}\APACrefatitle {K-{{Shape}}: {{Efficient}} and {{Accurate Clustering}} of {{Time Series}}} {K-{{Shape}}: {{Efficient}} and {{Accurate Clustering}} of {{Time Series}}}.{\BBCQ}
\newblock
\APACjournalVolNumPages{SIGMOD Rec.}{45}{1}{69--76,}
\newblock
\begin{APACrefDOI} \doi{10.1145/2949741.2949758} \end{APACrefDOI}
\newblock

\newblock

\PrintBackRefs{\CurrentBib}

\bibitem [\protect \citeauthoryear {%
Paparrizos%
, Yang%
\BCBL {}\ \BBA {} Li%
}{%
Paparrizos%
\ \protect \BOthers {.}}{%
{\protect \APACyear {2024}}%
}]{%
paparrizosBridgingGapDecade2024}
\APACinsertmetastar {%
paparrizosBridgingGapDecade2024}%
\begin{APACrefauthors}%
Paparrizos, J.%
, Yang, F.%
\BCBL {} Li, H.%
\end{APACrefauthors}%
\unskip\
\newblock
\APACrefYearMonthDay{2024}{{\APACmonth{12}}}{}.
\newblock
\APACrefbtitle {Bridging the {{Gap}}: {{A Decade Review}} of {{Time-Series Clustering Methods}}} {Bridging the {{Gap}}: {{A Decade Review}} of {{Time-Series Clustering Methods}}}\ (\BNUM\ arXiv:2412.20582).
\newblock
\APACaddressPublisher{}{arXiv}.
\PrintBackRefs{\CurrentBib}

\bibitem [\protect \citeauthoryear {%
Pedregosa%
\ \protect \BOthers {.}}{%
Pedregosa%
\ \protect \BOthers {.}}{%
{\protect \APACyear {2011}}%
}]{%
pedregosaScikitlearnMachineLearning2011}
\APACinsertmetastar {%
pedregosaScikitlearnMachineLearning2011}%
\begin{APACrefauthors}%
Pedregosa, F.%
, Varoquaux, G.%
, Gramfort, A.%
, Michel, V.%
, Thirion, B.%
, Grisel, O.%
\BDBL {}Duchesnay, {\'E}.%
\end{APACrefauthors}%
\unskip\
\newblock
\APACrefYearMonthDay{2011}{{\APACmonth{10}}}{}.
\newblock
{\BBOQ}\APACrefatitle {Scikit-Learn: {{Machine Learning}} in {{Python}}} {Scikit-learn: {{Machine Learning}} in {{Python}}}.{\BBCQ}
\newblock
\APACjournalVolNumPages{Journal of Machine Learning Research}{12}{}{2825-2830,}
\newblock

\newblock

\PrintBackRefs{\CurrentBib}

\bibitem [\protect \citeauthoryear {%
Perron%
}{%
Perron%
}{%
{\protect \APACyear {1989}}%
}]{%
perron1989great}
\APACinsertmetastar {%
perron1989great}%
\begin{APACrefauthors}%
Perron, P.%
\end{APACrefauthors}%
\unskip\
\newblock
\APACrefYearMonthDay{1989}{}{}.
\newblock
{\BBOQ}\APACrefatitle {The great crash, the oil price shock, and the unit root hypothesis} {The great crash, the oil price shock, and the unit root hypothesis}.{\BBCQ}
\newblock
\APACjournalVolNumPages{Econometrica: journal of the Econometric Society}{}{}{1361--1401,}
\newblock

\newblock

\PrintBackRefs{\CurrentBib}

\bibitem [\protect \citeauthoryear {%
Petitjean%
, Ketterlin%
\BCBL {}\ \BBA {} Gan{\c c}arski%
}{%
Petitjean%
\ \protect \BOthers {.}}{%
{\protect \APACyear {2011}}%
}]{%
petitjeanGlobalAveragingMethod2011a}
\APACinsertmetastar {%
petitjeanGlobalAveragingMethod2011a}%
\begin{APACrefauthors}%
Petitjean, F.%
, Ketterlin, A.%
\BCBL {} Gan{\c c}arski, P.%
\end{APACrefauthors}%
\unskip\
\newblock
\APACrefYearMonthDay{2011}{{\APACmonth{03}}}{}.
\newblock
{\BBOQ}\APACrefatitle {A Global Averaging Method for Dynamic Time Warping, with Applications to Clustering} {A global averaging method for dynamic time warping, with applications to clustering}.{\BBCQ}
\newblock
\APACjournalVolNumPages{Pattern Recognition}{44}{3}{678--693,}
\newblock
\begin{APACrefDOI} \doi{10.1016/j.patcog.2010.09.013} \end{APACrefDOI}
\newblock

\newblock

\PrintBackRefs{\CurrentBib}

\bibitem [\protect \citeauthoryear {%
Rakthanmanon%
\ \protect \BOthers {.}}{%
Rakthanmanon%
\ \protect \BOthers {.}}{%
{\protect \APACyear {2013}}%
}]{%
rakthanmanonAddressingBigData2013}
\APACinsertmetastar {%
rakthanmanonAddressingBigData2013}%
\begin{APACrefauthors}%
Rakthanmanon, T.%
, Campana, B.%
, Mueen, A.%
, Batista, G.%
, Westover, B.%
, Zhu, Q.%
\BDBL {}Keogh, E.%
\end{APACrefauthors}%
\unskip\
\newblock
\APACrefYearMonthDay{2013}{{\APACmonth{09}}}{}.
\newblock
{\BBOQ}\APACrefatitle {Addressing {{Big Data Time Series}}: {{Mining Trillions}} of {{Time Series Subsequences Under Dynamic Time Warping}}} {Addressing {{Big Data Time Series}}: {{Mining Trillions}} of {{Time Series Subsequences Under Dynamic Time Warping}}}.{\BBCQ}
\newblock
\APACjournalVolNumPages{ACM Trans. Knowl. Discov. Data}{7}{3}{10:1--10:31,}
\newblock
\begin{APACrefDOI} \doi{10.1145/2500489} \end{APACrefDOI}
\newblock

\newblock

\PrintBackRefs{\CurrentBib}

\bibitem [\protect \citeauthoryear {%
R{\"a}s{\"a}nen%
\ \BBA {} Kolehmainen%
}{%
R{\"a}s{\"a}nen%
\ \BBA {} Kolehmainen%
}{%
{\protect \APACyear {2009}}%
}]{%
rasanenFeatureBasedClusteringElectricity2009}
\APACinsertmetastar {%
rasanenFeatureBasedClusteringElectricity2009}%
\begin{APACrefauthors}%
R{\"a}s{\"a}nen, T.%
\BCBT {}\ \BBA {} Kolehmainen, M.%
\end{APACrefauthors}%
\unskip\
\newblock
\APACrefYearMonthDay{2009}{}{}.
\newblock
{\BBOQ}\APACrefatitle {Feature-{{Based Clustering}} for {{Electricity Use Time Series Data}}} {Feature-{{Based Clustering}} for {{Electricity Use Time Series Data}}}.{\BBCQ}
\newblock
 M.~Kolehmainen, P.~Toivanen\BCBL {}\ \BBA {} B.~Beliczynski\ (\BEDS), \APACrefbtitle {Adaptive and {{Natural Computing Algorithms}}} {Adaptive and {{Natural Computing Algorithms}}}\ (\BPGS\ 401--412).
\newblock
\APACaddressPublisher{Berlin, Heidelberg}{Springer}.
\PrintBackRefs{\CurrentBib}

\bibitem [\protect \citeauthoryear {%
Rokach%
\ \BBA {} Maimon%
}{%
Rokach%
\ \BBA {} Maimon%
}{%
{\protect \APACyear {2005}}%
}]{%
rokach2005clustering}
\APACinsertmetastar {%
rokach2005clustering}%
\begin{APACrefauthors}%
Rokach, L.%
\BCBT {}\ \BBA {} Maimon, O.%
\end{APACrefauthors}%
\unskip\
\newblock
\APACrefYearMonthDay{2005}{}{}.
\newblock
{\BBOQ}\APACrefatitle {Clustering methods} {Clustering methods}.{\BBCQ}
\newblock
 \APACrefbtitle {Data mining and knowledge discovery handbook} {Data mining and knowledge discovery handbook}\ (\BPGS\ 321--352).
\newblock
\APACaddressPublisher{}{Springer}.
\PrintBackRefs{\CurrentBib}

\bibitem [\protect \citeauthoryear {%
Ross%
, Lim%
, Lin%
\BCBL {}\ \BBA {} Yang%
}{%
Ross%
\ \protect \BOthers {.}}{%
{\protect \APACyear {2008}}%
}]{%
rossIncrementalLearningRobust2008}
\APACinsertmetastar {%
rossIncrementalLearningRobust2008}%
\begin{APACrefauthors}%
Ross, D.A.%
, Lim, J.%
, Lin, R\BHBI S.%
\BCBL {} Yang, M\BHBI H.%
\end{APACrefauthors}%
\unskip\
\newblock
\APACrefYearMonthDay{2008}{{\APACmonth{05}}}{}.
\newblock
{\BBOQ}\APACrefatitle {Incremental {{Learning}} for {{Robust Visual Tracking}}} {Incremental {{Learning}} for {{Robust Visual Tracking}}}.{\BBCQ}
\newblock
\APACjournalVolNumPages{International Journal of Computer Vision}{77}{1}{125--141,}
\newblock
\begin{APACrefDOI} \doi{10.1007/s11263-007-0075-7} \end{APACrefDOI}
\newblock

\newblock

\PrintBackRefs{\CurrentBib}

\bibitem [\protect \citeauthoryear {%
Ruta%
, Sawada%
, McKeough%
, Behrisch%
\BCBL {}\ \BBA {} Beyer%
}{%
Ruta%
\ \protect \BOthers {.}}{%
{\protect \APACyear {2019}}%
}]{%
rutaSAXNavigatorTime2019}
\APACinsertmetastar {%
rutaSAXNavigatorTime2019}%
\begin{APACrefauthors}%
Ruta, N.%
, Sawada, N.%
, McKeough, K.%
, Behrisch, M.%
\BCBL {} Beyer, J.%
\end{APACrefauthors}%
\unskip\
\newblock
\APACrefYearMonthDay{2019}{{\APACmonth{10}}}{}.
\newblock
{\BBOQ}\APACrefatitle {{{SAX Navigator}}: {{Time Series Exploration}} through {{Hierarchical Clustering}}} {{{SAX Navigator}}: {{Time Series Exploration}} through {{Hierarchical Clustering}}}.{\BBCQ}
\newblock
 \APACrefbtitle {2019 {{IEEE Visualization Conference}} ({{VIS}})} {2019 {{IEEE Visualization Conference}} ({{VIS}})}\ (\BPGS\ 236--240).
\PrintBackRefs{\CurrentBib}

\bibitem [\protect \citeauthoryear {%
Sakoe%
\ \BBA {} Chiba%
}{%
Sakoe%
\ \BBA {} Chiba%
}{%
{\protect \APACyear {1978}}%
}]{%
sakoeDynamicProgrammingAlgorithm1978}
\APACinsertmetastar {%
sakoeDynamicProgrammingAlgorithm1978}%
\begin{APACrefauthors}%
Sakoe, H.%
\BCBT {}\ \BBA {} Chiba, S.%
\end{APACrefauthors}%
\unskip\
\newblock
\APACrefYearMonthDay{1978}{{\APACmonth{02}}}{}.
\newblock
{\BBOQ}\APACrefatitle {Dynamic Programming Algorithm Optimization for Spoken Word Recognition} {Dynamic programming algorithm optimization for spoken word recognition}.{\BBCQ}
\newblock
\APACjournalVolNumPages{IEEE Transactions on Acoustics, Speech, and Signal Processing}{26}{1}{43--49,}
\newblock
\begin{APACrefDOI} \doi{10.1109/TASSP.1978.1163055} \end{APACrefDOI}
\newblock

\newblock

\PrintBackRefs{\CurrentBib}

\bibitem [\protect \citeauthoryear {%
Sch{\"a}fer%
}{%
Sch{\"a}fer%
}{%
{\protect \APACyear {2015}}%
{\protect \APACexlab {{\protect \BCnt {1}}}}}]{%
schaferBagOfSFASymbolsVectorSpace2015}
\APACinsertmetastar {%
schaferBagOfSFASymbolsVectorSpace2015}%
\begin{APACrefauthors}%
Sch{\"a}fer, P.%
\end{APACrefauthors}%
\unskip\
\newblock
\APACrefYearMonthDay{2015{\protect \BCnt {1}}}{{\APACmonth{05}}}{}.
\newblock
{\BBOQ}\APACrefatitle {Bag-{{Of-SFA-Symbols}} in {{Vector Space}} ({{BOSS VS}})} {Bag-{{Of-SFA-Symbols}} in {{Vector Space}} ({{BOSS VS}})}.{\BBCQ}.
\PrintBackRefs{\CurrentBib}

\bibitem [\protect \citeauthoryear {%
Sch{\"a}fer%
}{%
Sch{\"a}fer%
}{%
{\protect \APACyear {2015}}%
{\protect \APACexlab {{\protect \BCnt {2}}}}}]{%
schaferBOSSConcernedTime2015}
\APACinsertmetastar {%
schaferBOSSConcernedTime2015}%
\begin{APACrefauthors}%
Sch{\"a}fer, P.%
\end{APACrefauthors}%
\unskip\
\newblock
\APACrefYearMonthDay{2015{\protect \BCnt {2}}}{{\APACmonth{11}}}{}.
\newblock
{\BBOQ}\APACrefatitle {The {{BOSS}} Is Concerned with Time Series Classification in the Presence of Noise} {The {{BOSS}} is concerned with time series classification in the presence of noise}.{\BBCQ}
\newblock
\APACjournalVolNumPages{Data Mining and Knowledge Discovery}{29}{6}{1505--1530,}
\newblock
\begin{APACrefDOI} \doi{10.1007/s10618-014-0377-7} \end{APACrefDOI}
\newblock

\newblock

\PrintBackRefs{\CurrentBib}

\bibitem [\protect \citeauthoryear {%
Sch{\"a}fer%
\ \BBA {} H{\"o}gqvist%
}{%
Sch{\"a}fer%
\ \BBA {} H{\"o}gqvist%
}{%
{\protect \APACyear {2012}}%
}]{%
schaferSFASymbolicFourier2012}
\APACinsertmetastar {%
schaferSFASymbolicFourier2012}%
\begin{APACrefauthors}%
Sch{\"a}fer, P.%
\BCBT {}\ \BBA {} H{\"o}gqvist, M.%
\end{APACrefauthors}%
\unskip\
\newblock
\APACrefYearMonthDay{2012}{}{}.
\newblock
{\BBOQ}\APACrefatitle {{{SFA}}: A Symbolic Fourier Approximation and Index for Similarity Search in High Dimensional Datasets} {{{SFA}}: A symbolic fourier approximation and index for similarity search in high dimensional datasets}.{\BBCQ}
\newblock
 \APACrefbtitle {Proceedings of the 15th {{International Conference}} on {{Extending Database Technology}} - {{EDBT}} '12} {Proceedings of the 15th {{International Conference}} on {{Extending Database Technology}} - {{EDBT}} '12}\ (\BPG~516).
\newblock
\APACaddressPublisher{Berlin, Germany}{ACM Press}.
\PrintBackRefs{\CurrentBib}

\bibitem [\protect \citeauthoryear {%
Sch{\"a}fer%
\ \BBA {} Leser%
}{%
Sch{\"a}fer%
\ \BBA {} Leser%
}{%
{\protect \APACyear {2017}}%
}]{%
schaferFastAccurateTime2017a}
\APACinsertmetastar {%
schaferFastAccurateTime2017a}%
\begin{APACrefauthors}%
Sch{\"a}fer, P.%
\BCBT {}\ \BBA {} Leser, U.%
\end{APACrefauthors}%
\unskip\
\newblock
\APACrefYearMonthDay{2017}{{\APACmonth{11}}}{}.
\newblock
{\BBOQ}\APACrefatitle {Fast and {{Accurate Time Series Classification}} with {{WEASEL}}} {Fast and {{Accurate Time Series Classification}} with {{WEASEL}}}.{\BBCQ}
\newblock
 \APACrefbtitle {Proceedings of the 2017 {{ACM}} on {{Conference}} on {{Information}} and {{Knowledge Management}}} {Proceedings of the 2017 {{ACM}} on {{Conference}} on {{Information}} and {{Knowledge Management}}}\ (\BPGS\ 637--646).
\newblock
\APACaddressPublisher{New York, NY, USA}{Association for Computing Machinery}.
\PrintBackRefs{\CurrentBib}

\bibitem [\protect \citeauthoryear {%
Sch{\"a}fer%
\ \BBA {} Leser%
}{%
Sch{\"a}fer%
\ \BBA {} Leser%
}{%
{\protect \APACyear {2023}}%
}]{%
schaferWEASEL20Random2023}
\APACinsertmetastar {%
schaferWEASEL20Random2023}%
\begin{APACrefauthors}%
Sch{\"a}fer, P.%
\BCBT {}\ \BBA {} Leser, U.%
\end{APACrefauthors}%
\unskip\
\newblock
\APACrefYearMonthDay{2023}{{\APACmonth{12}}}{}.
\newblock
{\BBOQ}\APACrefatitle {{{WEASEL}} 2.0: A Random Dilated Dictionary Transform for Fast, Accurate and Memory Constrained Time Series Classification} {{{WEASEL}} 2.0: A random dilated dictionary transform for fast, accurate and memory constrained time series classification}.{\BBCQ}
\newblock
\APACjournalVolNumPages{Machine Learning}{112}{12}{4763--4788,}
\newblock
\begin{APACrefDOI} \doi{10.1007/s10994-023-06395-w} \end{APACrefDOI}
\newblock

\newblock

\PrintBackRefs{\CurrentBib}

\bibitem [\protect \citeauthoryear {%
Senin%
\ \BBA {} Malinchik%
}{%
Senin%
\ \BBA {} Malinchik%
}{%
{\protect \APACyear {2013}}%
}]{%
seninSAXVSMInterpretableTime2013}
\APACinsertmetastar {%
seninSAXVSMInterpretableTime2013}%
\begin{APACrefauthors}%
Senin, P.%
\BCBT {}\ \BBA {} Malinchik, S.%
\end{APACrefauthors}%
\unskip\
\newblock
\APACrefYearMonthDay{2013}{{\APACmonth{12}}}{}.
\newblock
{\BBOQ}\APACrefatitle {{{SAX-VSM}}: {{Interpretable Time Series Classification Using SAX}} and {{Vector Space Model}}} {{{SAX-VSM}}: {{Interpretable Time Series Classification Using SAX}} and {{Vector Space Model}}}.{\BBCQ}
\newblock
 \APACrefbtitle {2013 {{IEEE}} 13th {{International Conference}} on {{Data Mining}}} {2013 {{IEEE}} 13th {{International Conference}} on {{Data Mining}}}\ (\BPGS\ 1175--1180).
\PrintBackRefs{\CurrentBib}

\bibitem [\protect \citeauthoryear {%
Stefan%
, Athitsos%
\BCBL {}\ \BBA {} Das%
}{%
Stefan%
\ \protect \BOthers {.}}{%
{\protect \APACyear {2013}}%
}]{%
stefanMoveSplitMergeMetricTime2013}
\APACinsertmetastar {%
stefanMoveSplitMergeMetricTime2013}%
\begin{APACrefauthors}%
Stefan, A.%
, Athitsos, V.%
\BCBL {} Das, G.%
\end{APACrefauthors}%
\unskip\
\newblock
\APACrefYearMonthDay{2013}{{\APACmonth{06}}}{}.
\newblock
{\BBOQ}\APACrefatitle {The {{Move-Split-Merge Metric}} for {{Time Series}}} {The {{Move-Split-Merge Metric}} for {{Time Series}}}.{\BBCQ}
\newblock
\APACjournalVolNumPages{IEEE Transactions on Knowledge and Data Engineering}{25}{6}{1425--1438,}
\newblock
\begin{APACrefDOI} \doi{10.1109/TKDE.2012.88} \end{APACrefDOI}
\newblock

\newblock

\PrintBackRefs{\CurrentBib}

\bibitem [\protect \citeauthoryear {%
Tan%
, Dempster%
, Bergmeir%
\BCBL {}\ \BBA {} Webb%
}{%
Tan%
\ \protect \BOthers {.}}{%
{\protect \APACyear {2022}}%
}]{%
tanMultiRocketMultiplePooling2022}
\APACinsertmetastar {%
tanMultiRocketMultiplePooling2022}%
\begin{APACrefauthors}%
Tan, C.W.%
, Dempster, A.%
, Bergmeir, C.%
\BCBL {} Webb, G.I.%
\end{APACrefauthors}%
\unskip\
\newblock
\APACrefYearMonthDay{2022}{{\APACmonth{09}}}{}.
\newblock
{\BBOQ}\APACrefatitle {{{MultiRocket}}: Multiple Pooling Operators and Transformations for Fast and Effective Time Series Classification} {{{MultiRocket}}: Multiple pooling operators and transformations for fast and effective time series classification}.{\BBCQ}
\newblock
\APACjournalVolNumPages{Data Mining and Knowledge Discovery}{36}{5}{1623--1646,}
\newblock
\begin{APACrefDOI} \doi{10.1007/s10618-022-00844-1} \end{APACrefDOI}
\newblock

\newblock

\PrintBackRefs{\CurrentBib}

\bibitem [\protect \citeauthoryear {%
Tiano%
, Bonifati%
\BCBL {}\ \BBA {} Ng%
}{%
Tiano%
\ \protect \BOthers {.}}{%
{\protect \APACyear {2021}}%
}]{%
tianoFeatTSFeaturebasedTime2021}
\APACinsertmetastar {%
tianoFeatTSFeaturebasedTime2021}%
\begin{APACrefauthors}%
Tiano, D.%
, Bonifati, A.%
\BCBL {} Ng, R.%
\end{APACrefauthors}%
\unskip\
\newblock
\APACrefYearMonthDay{2021}{{\APACmonth{06}}}{}.
\newblock
{\BBOQ}\APACrefatitle {{{FeatTS}}: {{Feature-based Time Series Clustering}}} {{{FeatTS}}: {{Feature-based Time Series Clustering}}}.{\BBCQ}
\newblock
 \APACrefbtitle {Proceedings of the 2021 {{International Conference}} on {{Management}} of {{Data}}} {Proceedings of the 2021 {{International Conference}} on {{Management}} of {{Data}}}\ (\BPGS\ 2784--2788).
\newblock
\APACaddressPublisher{New York, NY, USA}{Association for Computing Machinery}.
\PrintBackRefs{\CurrentBib}

\bibitem [\protect \citeauthoryear {%
van~der Maaten%
\ \BBA {} Hinton%
}{%
van~der Maaten%
\ \BBA {} Hinton%
}{%
{\protect \APACyear {2008}}%
}]{%
maatenVisualizingDataUsing2008}
\APACinsertmetastar {%
maatenVisualizingDataUsing2008}%
\begin{APACrefauthors}%
van~der Maaten, L.%
\BCBT {}\ \BBA {} Hinton, G.%
\end{APACrefauthors}%
\unskip\
\newblock
\APACrefYearMonthDay{2008}{}{}.
\newblock
{\BBOQ}\APACrefatitle {Visualizing {{Data}} Using T-{{SNE}}} {Visualizing {{Data}} using t-{{SNE}}}.{\BBCQ}
\newblock
\APACjournalVolNumPages{Journal of Machine Learning Research}{9}{86}{2579--2605,}
\newblock

\newblock

\PrintBackRefs{\CurrentBib}

\bibitem [\protect \citeauthoryear {%
{von Luxburg}%
}{%
{von Luxburg}%
}{%
{\protect \APACyear {2007}}%
}]{%
vonluxburgTutorialSpectralClustering2007}
\APACinsertmetastar {%
vonluxburgTutorialSpectralClustering2007}%
\begin{APACrefauthors}%
{von Luxburg}, U.%
\end{APACrefauthors}%
\unskip\
\newblock
\APACrefYearMonthDay{2007}{{\APACmonth{12}}}{}.
\newblock
{\BBOQ}\APACrefatitle {A Tutorial on Spectral Clustering} {A tutorial on spectral clustering}.{\BBCQ}
\newblock
\APACjournalVolNumPages{Statistics and Computing}{17}{4}{395--416,}
\newblock
\begin{APACrefDOI} \doi{10.1007/s11222-007-9033-z} \end{APACrefDOI}
\newblock

\newblock

\PrintBackRefs{\CurrentBib}

\bibitem [\protect \citeauthoryear {%
Waskom%
}{%
Waskom%
}{%
{\protect \APACyear {2021}}%
}]{%
waskomSeabornStatisticalData2021}
\APACinsertmetastar {%
waskomSeabornStatisticalData2021}%
\begin{APACrefauthors}%
Waskom, M.L.%
\end{APACrefauthors}%
\unskip\
\newblock
\APACrefYearMonthDay{2021}{{\APACmonth{04}}}{}.
\newblock
{\BBOQ}\APACrefatitle {Seaborn: Statistical Data Visualization} {Seaborn: Statistical data visualization}.{\BBCQ}
\newblock
\APACjournalVolNumPages{Journal of Open Source Software}{6}{60}{3021,}
\newblock
\begin{APACrefDOI} \doi{10.21105/joss.03021} \end{APACrefDOI}
\newblock

\newblock

\PrintBackRefs{\CurrentBib}

\bibitem [\protect \citeauthoryear {%
{W}es {M}c{K}inney%
}{%
{W}es {M}c{K}inney%
}{%
{\protect \APACyear {2010}}%
}]{%
mckinney-proc-scipy-2010}
\APACinsertmetastar {%
mckinney-proc-scipy-2010}%
\begin{APACrefauthors}%
{W}es {M}c{K}inney%
\end{APACrefauthors}%
\unskip\
\newblock
\APACrefYearMonthDay{2010}{}{}.
\newblock
{\BBOQ}\APACrefatitle {{D}ata {S}tructures for {S}tatistical {C}omputing in {P}ython} {{D}ata {S}tructures for {S}tatistical {C}omputing in {P}ython}.{\BBCQ}
\newblock
 {S}t\'efan van~der {W}alt\ \BBA {} {J}arrod {M}illman\ (\BEDS), \APACrefbtitle {{P}roceedings of the 9th {P}ython in {S}cience {C}onference} {{P}roceedings of the 9th {P}ython in {S}cience {C}onference}\ (\BPG~56 - 61).
\PrintBackRefs{\CurrentBib}

\bibitem [\protect \citeauthoryear {%
Yang%
\ \BBA {} Leskovec%
}{%
Yang%
\ \BBA {} Leskovec%
}{%
{\protect \APACyear {2011}}%
}]{%
yangPatternsTemporalVariation2011}
\APACinsertmetastar {%
yangPatternsTemporalVariation2011}%
\begin{APACrefauthors}%
Yang, J.%
\BCBT {}\ \BBA {} Leskovec, J.%
\end{APACrefauthors}%
\unskip\
\newblock
\APACrefYearMonthDay{2011}{{\APACmonth{02}}}{}.
\newblock
{\BBOQ}\APACrefatitle {Patterns of Temporal Variation in Online Media} {Patterns of temporal variation in online media}.{\BBCQ}
\newblock
 \APACrefbtitle {Proceedings of the Fourth {{ACM}} International Conference on {{Web}} Search and Data Mining} {Proceedings of the fourth {{ACM}} international conference on {{Web}} search and data mining}\ (\BPGS\ 177--186).
\newblock
\APACaddressPublisher{New York, NY, USA}{Association for Computing Machinery}.
\PrintBackRefs{\CurrentBib}

\bibitem [\protect \citeauthoryear {%
Ye%
\ \BBA {} Keogh%
}{%
Ye%
\ \BBA {} Keogh%
}{%
{\protect \APACyear {2011}}%
}]{%
yeTimeSeriesShapelets2011}
\APACinsertmetastar {%
yeTimeSeriesShapelets2011}%
\begin{APACrefauthors}%
Ye, L.%
\BCBT {}\ \BBA {} Keogh, E.%
\end{APACrefauthors}%
\unskip\
\newblock
\APACrefYearMonthDay{2011}{{\APACmonth{01}}}{}.
\newblock
{\BBOQ}\APACrefatitle {Time Series Shapelets: A Novel Technique That Allows Accurate, Interpretable and Fast Classification} {Time series shapelets: A novel technique that allows accurate, interpretable and fast classification}.{\BBCQ}
\newblock
\APACjournalVolNumPages{Data Mining and Knowledge Discovery}{22}{1}{149--182,}
\newblock
\begin{APACrefDOI} \doi{10.1007/s10618-010-0179-5} \end{APACrefDOI}
\newblock

\newblock

\PrintBackRefs{\CurrentBib}

\bibitem [\protect \citeauthoryear {%
Zakaria%
, Mueen%
\BCBL {}\ \BBA {} Keogh%
}{%
Zakaria%
\ \protect \BOthers {.}}{%
{\protect \APACyear {2012}}%
}]{%
zakariaClusteringTimeSeries2012}
\APACinsertmetastar {%
zakariaClusteringTimeSeries2012}%
\begin{APACrefauthors}%
Zakaria, J.%
, Mueen, A.%
\BCBL {} Keogh, E.%
\end{APACrefauthors}%
\unskip\
\newblock
\APACrefYearMonthDay{2012}{{\APACmonth{12}}}{}.
\newblock
{\BBOQ}\APACrefatitle {Clustering {{Time Series Using Unsupervised-Shapelets}}} {Clustering {{Time Series Using Unsupervised-Shapelets}}}.{\BBCQ}
\newblock
 \APACrefbtitle {2012 {{IEEE}} 12th {{International Conference}} on {{Data Mining}}} {2012 {{IEEE}} 12th {{International Conference}} on {{Data Mining}}}\ (\BPGS\ 785--794).
\PrintBackRefs{\CurrentBib}

\bibitem [\protect \citeauthoryear {%
Zhang%
, Wu%
, Yang%
, Tian%
\BCBL {}\ \BBA {} Zhang%
}{%
Zhang%
\ \protect \BOthers {.}}{%
{\protect \APACyear {2016}}%
}]{%
zhangUnsupervisedFeatureLearning2016}
\APACinsertmetastar {%
zhangUnsupervisedFeatureLearning2016}%
\begin{APACrefauthors}%
Zhang, Q.%
, Wu, J.%
, Yang, H.%
, Tian, Y.%
\BCBL {} Zhang, C.%
\end{APACrefauthors}%
\unskip\
\newblock
\APACrefYearMonthDay{2016}{{\APACmonth{07}}}{}.
\newblock
{\BBOQ}\APACrefatitle {Unsupervised Feature Learning from Time Series} {Unsupervised feature learning from time series}.{\BBCQ}
\newblock
 \APACrefbtitle {Proceedings of the {{Twenty-Fifth International Joint Conference}} on {{Artificial Intelligence}}} {Proceedings of the {{Twenty-Fifth International Joint Conference}} on {{Artificial Intelligence}}}\ (\BPGS\ 2322--2328).
\newblock
\APACaddressPublisher{New York, New York, USA}{AAAI Press}.
\PrintBackRefs{\CurrentBib}

\bibitem [\protect \citeauthoryear {%
Zhao%
\ \BBA {} Itti%
}{%
Zhao%
\ \BBA {} Itti%
}{%
{\protect \APACyear {2018}}%
}]{%
zhaoShapeDTWShapeDynamic2018}
\APACinsertmetastar {%
zhaoShapeDTWShapeDynamic2018}%
\begin{APACrefauthors}%
Zhao, J.%
\BCBT {}\ \BBA {} Itti, L.%
\end{APACrefauthors}%
\unskip\
\newblock
\APACrefYearMonthDay{2018}{{\APACmonth{02}}}{}.
\newblock
{\BBOQ}\APACrefatitle {{{shapeDTW}}: {{Shape Dynamic Time Warping}}} {{{shapeDTW}}: {{Shape Dynamic Time Warping}}}.{\BBCQ}
\newblock
\APACjournalVolNumPages{Pattern Recognition}{74}{}{171--184,}
\newblock
\begin{APACrefDOI} \doi{10.1016/j.patcog.2017.09.020} \end{APACrefDOI}
\newblock

\newblock

\PrintBackRefs{\CurrentBib}

\end{thebibliography}

\end{document}